\documentclass[12pt,twoside,phd]{dms}

\usepackage[utf8]{inputenc} 
\usepackage[T1]{fontenc}    %

\usepackage{lmodern}

\anglais

\def\sloppy{%
  \tolerance 500
  \emergencystretch 3em%
  \hfuzz .5pt
  \vfuzz\hfuzz}
\newcommand{\distances}{\mathrm{T}}

\newcommand{\tree}{\mathbf{T}}

\newcommand{\cummax}{\mathrm{cummax}}
\newcommand{\cell}{\mathbf{c}}
\newcommand{\cp}[0]{\overrightarrow{\pi}}
\newcommand{\rcp}[0]{\overleftarrow{\pi}}
\newcommand{\vg}[0]{v}
\newcommand{\hg}[0]{h}
\newcommand{\cg}[0]{c}

\newcommand{\option}{\mathbf{o}}
\newcommand{\controller}{\mathbf{c}}
\newcommand{\dem}{\mathbf{d}}
\newcommand{\dems}{\mathbf{D}}
\newcommand{\task}{\tau}
\newcommand{\tasks}{\mathbf{T}}
\newcommand{\action}{\mathbf{a}}
\newcommand{\prob}{\mathbf{p}}
\newcommand{\hid}{\mathbf{h}}
\newcommand{\x}{\mathbf{x}}

\newcommand{\e}{\mathbf{e}}
\newcommand{\h}{\mathbf{h}}
\newcommand{\mask}{m}
\newcommand{\head}{\mathbf{h}^{\mathrm{H}}}
\newcommand{\dep}{\mathbf{h}^{\mathrm{D}}}
\newcommand{\query}{\mathbf{q}}
\newcommand{\key}{\mathbf{k}}
\newcommand{\gate}{\mathbf{g}}
\newcommand{\val}{\mathbf{v}}
\newcommand{\weight}{\mathbf{W}}
\newcommand{\out}{\mathbf{o}}
\newcommand{\bias}{\mathbf{b}}
\newcommand{\R}{\mathbb{R}}

\usepackage{graphicx,amssymb,icomma}
\usepackage{amsmath}
\usepackage{amsthm}
\usepackage{algorithm}
\usepackage{algpseudocode}
\usepackage{multirow}
\usepackage{booktabs}  
\usepackage{natbib}
\usepackage{diagbox}
\usepackage{forest}
\forestset{
    nice empty nodes/.style={
        for tree={
            s sep=0.1em, 
            l sep=0.33em,
            inner ysep=0.4em, 
            inner xsep=0.05em,
            l=0,
            calign=midpoint,
            fit=tight,
            where n children=0{
               tier=word,
               minimum height=1.25em,
            }{},
            where n children=2{
               l-=1em,
            }{},
            parent anchor=south,
            child anchor=north,
            delay={if content={}{
                    inner sep=0pt,
                    edge path={\noexpand\path [\forestoption{edge}] 
                    			(!u.parent anchor) 
                               -- (.south)\forestoption{edge label};}
                }{}}
        },
    },
}

\usepackage[labelfont=bf, width=\linewidth]{caption}
\usepackage{subcaption}

\usepackage{hyperref}  
\hypersetup{colorlinks=true,allcolors=black}
\usepackage{hypcap}   
\usepackage{bookmark} 

\newtheorem{cor}{\corollaryname}[section]
\newtheorem{deff}[cor]{\definitionname}

\theoremstyle{definition}

\numberwithin{equation}{section}

\usepackage[onehalfspacing]{setspace}

\begin{document}

\version{1}



\title{Syntactic Inductive Biases
for Deep Learning Methods}

\author{Yikang Shen}

\copyrightyear{2021}

\department{Département d'informatique et de recherche opérationnelle}

\date{\today} 

\sujet{Computer Science}

\president{Sarath Chandar}

\directeur{Aaron Courville}


\membrejury{Yoshua Bengio}

\examinateur{Alexander Rush}   



\repdoyen{François Lareau} 


\maketitle

\maketitle


\francais
\chapter*{Résumé}

Le débat entre connexionnisme et symbolisme est l'une des forces majeures qui animent le développement de l'Intelligence Artificielle.
L'apprentissage profond et la linguistique théorique sont les domaines d'études les plus représentatifs pour les deux écoles respectivement.
Alors que la méthode d'apprentissage profond a fait des percées impressionnantes et est devenue la principale raison de la récente prospérité de l'IA pour l'industrie et les universités, la linguistique et le symbolisme occupent quelque domaines importantes, notamment l'interprétabilité et la fiabilité.

Dans cette thèse, nous essayons de construire une connexion entre les deux écoles en introduisant des biais inductifs linguistiques pour les modèles d'apprentissage profond.
Nous proposons deux familles de biais inductifs, une pour la structure de circonscription et une autre pour la structure de dépendance.
Le biais inductif de circonscription encourage les modèles d'apprentissage profond à utiliser différentes unités (ou neurones) pour traiter séparément les informations à long terme et à court terme.
Cette séparation fournit un moyen pour les modèles d'apprentissage profond de construire les représentations hiérarchiques latentes à partir d'entrées séquentielles, dont une représentation de niveau supérieur est composée et peut être décomposée en une série de représentations de niveau inférieur.
Par exemple, sans connaître la structure de vérité fondamentale, notre modèle proposé apprend à traiter l'expression logique en composant des représentations de variables et d'opérateurs en représentations d'expressions selon sa structure syntaxique.
D'autre part, le biais inductif de dépendance encourage les modèles à trouver les relations latentes entre les mots dans la séquence d'entrée.
Pour le langage naturel, les relations latentes sont généralement modélisées sous la forme d'un graphe de dépendance orienté, où un mot a exactement un nœud parent et zéro ou plusieurs nœuds enfants.
Après avoir appliqué cette contrainte à un modèle de type transformateur, nous constatons que le modèle est capable d'induire des graphes orientés proches des annotations d'experts humains, et qu'il surpasse également le modèle de transformateur standard sur différentes tâches.
Nous pensons que ces résultats expérimentaux démontrent une alternative intéressante pour le développement futur de modèles d'apprentissage profond.


\anglais
\chapter*{Abstract}
The debate between connectionism and symbolism is one of the major forces that drive the development of Artificial Intelligence.
Deep Learning and theoretical linguistics are the most representative fields of study for the two schools respectively. 
While the deep learning method has made impressive breakthroughs and became the major reason behind the recent AI prosperity for industry and academia, linguistics and symbolism still holding some important grounds including reasoning, interpretability and reliability.

In this thesis, we try to build a connection between the two schools by introducing syntactic inductive biases for deep learning models.
We propose two families of inductive biases, one for constituency structure and another one for dependency structure.
The constituency inductive bias encourages deep learning models to use different units (or neurons) to separately process long-term and short-term information.
This separation provides a way for deep learning models to build the latent hierarchical representations from sequential inputs, that a higher-level representation is composed of and can be decomposed into a series of lower-level representations.
For example, without knowing the ground-truth structure, our proposed model learns to process logical expression through composing representations of variables and operators into representations of expressions according to its syntactic structure.
On the other hand, the dependency inductive bias encourages models to find the latent relations between entities in the input sequence.
For natural language, the latent relations are usually modeled as a directed dependency graph, where a word has exactly one parent node and zero or several children nodes.
After applying this constraint to a transformer-like model, we find the model is capable of inducing directed graphs that are close to human expert annotations, and it also outperforms the standard transformer model on different tasks.
We believe that these experimental results demonstrate an interesting alternative for the future development of deep learning models.


\anglais
\cleardoublepage
\pdfbookmark[chapter]{\contentsname}{toc}  
\tableofcontents

\cleardoublepage
\phantomsection  
\listoftables

\cleardoublepage
\phantomsection
\listoffigures

%
%
%

\chapter*{List of acronyms and abbreviations}
\begin{twocolumnlist}{.2\textwidth}{.7\textwidth}
  NLP & Natural Language Processing \\
  UF1 & Unlabeled F1 score for evaluating constituency parsing performance\\
  UAS  & Unlabeled (directed) Attachment Score for evaluating dependency parsing performance \\
  UUAS  &  Unlabeled Undirected Attachment Score for evaluating dependency parsing performance \\
  RNN & Recurrent Neural Network\\
  LSTM & Long-Short Term Memory\\
  CNN & Convolutional Neural Network\\
\end{twocolumnlist}


\chapter*{Acknowledgements}
The past five years at the University of Montreal and Mila have been the most valuable and unforgettable experience for me so far.
It's not easy to say goodbye to such a wonderful time, as well as the people and life.
When I started my Ph.D. study in September 2016, I only have a vague idea about the path and challenges in front of me.
In 2016, deep learning is already on its hype, some researchers suggested that its era will end in a few years. 
But the fact is that exciting breakthrough and challenge new questions still emerge every year. 
The technique gradually becomes an essential part of everyone's life.
From self-driving car to the photo album in a smartphone, they all include at least one deep learning model to provide some functions that you wouldn't expect 10 years ago.
It's fair to say that the development of Artificial Intelligence (AI, mostly deep learning) is reforming our society.
Although I have some different opinions on the future directions and still believe in them. 
But it doesn't stop me from feeling excited about what is happening over the last few years and now.
It's a great honor to witness and participate in this evolution in the front seat.
I would not have been able to make this journey without the help and support of many people and I feel deeply indebted to them.

My greatest thanks should go to my advisor Aaron Courville. 
It was a great privilege to work with one of the most important researchers in the deep learning community.
I can always rely on him to provide good advice for my research as well as my career.
I am not a student who always listens carefully to the professor's advice, but Aaron is always supportive. 
His support is the most important reason that allows me to focus on studying a relatively niche field.
He has a very insightful, high-level view of the field while he can also quickly understand details and understands the nature of the problems very well.
More importantly, he always encourages collaboration, not only between his students but also with other professors and external researchers.
Collaborations grant me access to lots of mental resources and allow me to understand problems from very different prospects.

I would like to thank Alessandro Sordoni and Siva Reddy -- the other two mentors in my Ph.D. study, for a lot of guidance and help throughout the last few years.
Alessandro is my host during my 3-years long part-time research internship at Microsoft Research (this program also provides many free meals and unexpected high life quality for a Ph.D. student).
He is a good criticizer and almost like a co-supervisor for me.
I can only have enough confidence in my work after passing his trial of insightful questions.
Siva is a charming person and knowledgeable NLP researcher.
He provides a lot of valuable linguistic-related advice, while I am in desperate need of this guidance and both Aaron and Alessandro focus their study on the machine learning field.
Personally, his attitude towards life also affects me. 
I want to be the same energetic and responsible person as him someday.
I want to thank Aaron, Alessandro, and Siva for all the guidance and patience that they have provided, and I will always be proud to be your student.

I also want to thank Shawn Tan and Zhouhan Lin.
They're the most important collaborators for me.
Almost all of my works are done in close collaborations with either Zhouhan or Shawn.
I want to thank Shawn for the countless discussions that we have had. 
He is capable of understanding the most complicated idea and formalize it with clear and rigorous math equations.
Together, we wrote Ordered Neurons \citep{shen2018ordered} and some other papers that this thesis is based on.
I want to thank Zhouhan for been the big buddy that helps me adapt to my Ph.D. life and booting my research.
Our works on unsupervised and supervised parsing \citep{shen2017neural, shen2018straight} leads me to the works presented in this thesis.
We are also very close friends. 
I want to thanks them for providing lots of support in my difficult time and tolerate my occasional aggressiveness.

I want to thank the Mila community. 
It gathers a lot of brilliant researchers and created a unique atmosphere of collaboration.
I met many interesting people in Mila. 
Among them, I especially want to thank Ying Zhang, Jie Fu, Jae Hyun Lim, Min Lin, Chin-Wei Huang, Saizheng Zhang, Yuchen Lu, Xing Chen, Amina Madzhun, Zhen Liu. 
I had a lot of joyful times with them. 
They also gave me a lot of help at various times.
Outside of Mila, I also met many great friends including Peng Lu, Yi Tay, Che Zheng, Xingdi Yuan, Lili Mou (just to name a few). 
I also want to thank my parents for their unwavering support and love during the last three decades.


 %
 %

\NoChapterPageNumber
\cleardoublepage

\anglais
\chapter{Introduction}


\section{The Evolution of Paradigm in NLP}
While research in Natural Language Processing (NLP) is now dominant by deep learning methods, the two fields were separately developed for decades.
Prior to the deep learning revolution, the difference between the two domains can trace back to the difference between connectionism and symbolism.

Before the 1990s, approaches for solving NLP problems were predominantly \textit{symbolic}~\citep{chao1968language}.
Symbolic systems directly model abstract concepts and the innate structure of the human mind.
Elementary semantics are represented by symbols. 
Complex semantics are represented by a group of symbols combined by operations and syntactic structures. 
Many early AI advances utilized a symbolic approach to AI programming, striving to create smart systems by modeling relationships and using symbols and programs to convey meaning.
Symbolic systems also allow straightforward generalization through assembling known rules and symbols into a new syntactic structure.
But symbolic systems have several defects: 
1) limited fault tolerance, a failure of a small component usually causes the entire system break;
2) they can't process inputs that include undefined rules or symbols resulting in a relatively limited learning capacity; 
3) applying symbolic methods to a new problem requires lots of human expertise.

In the 1990s, the paradigm shifts from symbolic methods to \textit{empirical} or \textit{statistical} methods~\citep{abney1996statistical}.
The empirical view assumes that language is a natural phenomenon whose effects are observable in the world as data. 
The best way to build a NLP model is to learn from the data.
Empirical methods nicely solved some defects of the symbolic methods. 
For example, a probabilistic framework can handle the ambiguity in natural language by assigning probabilities to different analyses.
Applying an empirical method to a new problem could be as simple as training the method on a set of pre-collected data.
These methods enjoyed great success in almost all problems in natural language processing until deep learning methods became prominent.

Starting from the middle of the 2010s, the paradigm shifts again to \textit{deep learning} methods, which try to process input with a powerful universal function approximator, without explicitly modeling discrete structure and operations.
The idea of deep learning can trace its roots back to the concept of connectionism.
Some advantages of the connectionist approach include its applicability to a broad array of functions, a structural approximation to biological neurons, low requirements for innate structure, and capacity for graceful degradation.
Recent progress shows that deep learning models could be trained in an end-to-end schema from human annotations or unsupervised losses.
Deep learning models can be easily applied to different tasks that have enough training data and clear input/output definitions.
The capacity of a neural model can be easily improved by increasing the number of parameters.
The data-driven feature and expansibility of the deep learning model make it very popular in industrial applications.
Recent studies show that deep learning-based NLP models have proven to be capable of learning a remarkable amount of syntax, despite having much weaker structural priors than Chomsky’s model of Universal Grammar. 

\section{Motivation}
Despite being very popular and empirically successful, the limitation of deep learning methods is starting to draw more attention:
1) they heavily rely on training data and computation resources, GPT-3~\citep{brown2020language} has 170 billion parameters and is trained on a dataset of about 500 billion tokens;
2) they fail to generalize to unexpected data points, for example, an unforeseen combination of known tokens and extra-long inputs~\citep{tay2020long};
3) it lacks explainability, state-of-the-art deep learning models are like black boxes, such that it is almost always impossible to understand why decisions are made, and it's also impossible to identify the source of an error or fix the error in a way that will not potentially hurt the other functionalities of the model~\citep{zhang2018visual}.
However, even with exponential increases in computing power and increasingly vast quantities of data, the improvement curve for predictive power is leveling off, suggesting that there is a cap to how far connectionism can take us.

One major difference between deep learning methods and many previous approaches is the assumption that there exist a syntax and a set of semantic functions. 
Syntax is the set of rules, principles, and processes that govern the structure of sentences (sentence structure) in a given language, usually including word order.
The syntax of a language describes the latent structure of a valid sentence, but does not provide any information about the meaning of that sentence.
The meaning given to a combination of symbols is handled by semantics.
In other words, for a given sentence $x_1, ..., x_T$, its syntactic structure can be seen as a computation graph $\mathbf{G}$ that is equipped with semantic functions.
The computation graph $\mathbf{G}$ takes the meaning of each token $x_i$ as input, and outputs the meaning of the sentence.
This assumption provides powerful generalizations and explainability.
The disentanglement of structure and functions allows the same set of semantic functions to be combined in many different ways to process different inputs and solve different tasks. 
The existence of a clear structure also provides rich and meaningful intermediate results.
These advantages provide a potential solution to the defects of deep learning. 

\section{Syntactic Inductive Biases}
In this thesis, we study the problem of building syntax-sensitive deep learning models.
In particular, we introduce \textbf{syntactic inductive biases} -- a new family of inductive biases that use a probabilistic relaxation of discrete syntactic structure to regularize the internal connection of deep learning models.

An \textit{inductive bias} of a learning algorithm is a set of assumptions that the learner uses to predict outputs of given inputs that it has not encountered \citep{mitchell1980need}.
In the field of deep learning, inductive bias usually refers to special neural network architecture designs, that leverage prior knowledge to regularize a more general version of deep learning model.
These special designs encourage the model to prioritise solutions with specific properties that requires hand-engineering in traditional methods.
One famous example of an inductive bias for deep learning is Convolutional Neural Network (CNN).
CNNs regularize multilayer perceptrons through weight sharing.
Multilayer perceptrons are fully connected networks, that is, each neuron in one layer is connected to all neurons in the next layer.
CNNs use convolution in place of general matrix multiplication in at least one of their layers.
CNNs learn to optimize the filters (or kernels) through automated learning, whereas in traditional methods these filters are hand-engineered. 
This independence from prior knowledge and human intervention in feature extraction is a major advantage.

Since there are two major classes of natural language syntax: dependency grammar and constituency grammar.
We introduce two respective types of inductive bias: \textit{constituency inductive bias} and \textit{dependency inductive bias}.

Constituency grammars model the assembly of one or several corresponded words.
From a syntactic point of view, a constituent is a word or a group of words that function as a single unit within a hierarchical structure.
From a semantic point of view, a constituent is a unit with a stand-alone meaning.
Although some syntactically well-formed constituents are nonsensical, e.g.``Colorless green ideas sleep furiously''.
For most NLP applications, a constituent should be syntactically and semantically well-formed.
Furthermore, in constituency trees, larger constituents are composed of smaller constituents.
The process of composition is modeled by compositional semantic functions.
Traditionally, constituency grammars treat the structure as primitive. 
Constituency grammars derive the functions from the constellation.
For instance, the object is identified as the NP appearing inside finite VP, and the subject as the NP appearing outside of finite VP.
The \textit{constituency inductive bias} should encourage deep learning models to:
1) induce the latent constituency structure of input sentences,
2) model the compositional semantic functions,
3) compute meaningful representations for constituents.

Dependency grammars model one-to-one correspondences between words.
In a dependency graph, the main verb is taken to be the structural center of a clause structure. 
All other syntactic units (words) are either directly or indirectly connected to the verb in terms of the directed links, which are called dependencies.
Dependency grammars have flatter tree structures than constituency grammars in part because they lack a finite verb phrase constituent, and they are thus well suited for the analysis of languages with free word order, such as Czech or Warlpiri.
Different from constituency grammars, dependency grammars treat the syntactic functions as primitive.
They posit an inventory of functions (e.g. subject, object, oblique, determiner, attribute, predicative, etc.). 
These functions can appear as labels on the dependencies in the tree structures.
The \textit{dependency inductive bias} should: 
1) explicitly model different syntactic functions as separate modules,
2) induce latent dependency edges between words,
3) compute meaningful representations for each node (words) in the dependency graph.

Furthermore, the idea of syntactic inductive bias is not exclusive for NLP tasks.
In fact, one school of thought sees syntax as a non-innate adaptation to innate cognitive mechanisms \citep{hawkins2004efficiency}.
Cross-linguistic tendencies are considered as being based on language users' preference for grammars that are organized efficiently, and on their avoidance of word orderings which cause processing difficulty.
This suggest that an effective syntactic inductive bias could capture some fundamental cognitive mechanisms.
In the real world, lots of tasks have sparse rewards and long time horizons, which typically pose significant challenges in reinforcement learning.
HIL and HRL handle this problem by introducing a temporal abstraction~\citep{stolle2002learning} -- a innate hierarchical structure over time.
This idea is coherent with constituency structure.
Based on this observation, we extend the application of our constituency inductive bias to Hierarchical Imitation Learning (HIL) and Hierarchical Reinforcement Learning (HRL) domain.
In the setting of HIL and HRL, we expect the constituency inductive bias can encourage the agent to:
1) segment a given task into meaningful subtasks;
2) model the subtasks as separate skills (neural network modules);
3) recompose these skills to solve a new tasks.

\section{Why not Supervised Parsing?}
Learning structures from treebanks has its advantages -- it is very well studied, a supervised parsing can achieve very high accuracy. 
But a trained parsing has the limitation of \textit{domain dependence}.
Parsers trained on English Penn Treebank~\citep{marcus1994penn} degrade substantially when applied to new genres and domains, and fail when applied to new languages.
This has spurred an area of research on parser adapation~\citep{ustun2020udapter, zeman2008cross, mcclosky2006reranking}.
Treebanks now exist for several domains and languages, but each treebank requires many resources and years of work to construct, and most languages are without treebank.

Taking the domain dependency problem to the extreme, we have \textit{modality} problem. 
Modern deep learning models, like Transformer~\citep{vaswani2017attention}, are considered as foundation models for AI~\citep{bommasani2021opportunities}.
A new deep learning technique could potentially be applied to many different modalities (e.g. vision, robotic, and language), and some multi-modality problems.
Many of these domains have latent structures.
For example, images can be parsed to a hierarchical structure~\citep{tu2005image}, hierarchical reinforcement learning~\citep{barto2003recent} is an important technique for robotic.
But most of these domains don't have a convention for human-labeled latent structure.
A model that requires structural supervision would have limited versatility.

A further limitation of treebanks is that they always have a fixed convention.
But the ``right'' kind of syntactic structure seems to depend heavily on the beneficiary task.
There have been studies comparing different kinds of syntactic structures for their usefulness on specific tasks~\citep{gildea-2004-dependencies}.
Finetuning pretrained language models on a specific task makes them learn a convention that suits better the targeted task~\citep{dai2021does}.
From this perspective, a method that can induce syntactic structure from downstream tasks could be more useful than a supervised parser.

\section{Thesis Outline}
Following the discussion in previous section, the rest of this thesis is organized around following themes:
\begin{itemize}
    \item In Chapter~\ref{cha:preliminaries}, we review some widely used neural network components and architectures. 
    These components and architectures are foundation of models introduced in this thesis.
    We then provide definitions to the tasks that we will use to evaluate syntactic inductive biases.
    
    \item In Chapter~\ref{cha:constituency}, we first review the history of constituency-augmented neural networks and constituency inductive biases.
    We then introduce \textit{Ordered Neurons} -- a constituency inductive bias for recurrent neural networks.
    Based on the ordered neurons, we further introduce two neural network architectures: ON-LSTM and Ordered Memory.
    We present experiment results on formal language tasks, language modeling, and unsupervised constituency parsing.
    
    \item In Chapter~\ref{cha:dependency}, we first review the history of dependency-augmented neural networks and dependency inductive biases.
    We then introduce \textit{dependency-constrained connection} -- an inductive bias for transformer or graph neural networks.
    We present experiment results on masked language modeling, unsupervised dependency parsing and semantic textual similarity tasks.
    
    \item In Chapter~\ref{cha:rl}, we first review the history of Hierarchical Imitation and Reinforcement Learning (HIRL).
    We then introduce the Option-Controller Network -- an HIRL model with a constituency inductive bias.
    In experiments, we perform behavior cloning from unstructured demonstrations coming from different tasks, and during the RL finetuning, we freeze the learned options and only re-initialize the controller.
    
    \item In Chapter~\ref{cha:conclusion},  we first conclude the thesis, then review three important research questions in this field.
    The first question is emerging better and more discrete structure.
    The second question discusses inducing reusable operators.
    The last question discusses systematic generalization based on the induced structure and operators.
    These questions still remain as open questions and yet to be answered in the future.
\end{itemize}

\section{Contributions}
The contributions of this thesis are summarized as follows:
\begin{itemize}
    \item I systematically studied syntactic inductive bias for deep learning methods.
    I studied a wide spectrum of neural network architecture design.
    I summarized our works and proposed two inductive biases: constituency inductive bias and dependency inductive bias.
    \item I were among the first to study neural unsupervised parsing. 
    I proposed several models that can achieve strong results on unsupervised dependency and constituency parsing.
    \item I also show that understanding the latent structure of input is essential for strong generalization on language tasks. 
    \item I further expand the application of syntactic inductive bias to hierarchical imitation and reinforcement learning domain.
    This application result in a highly modularized model that can effectively reuse learnt skills to solve new tasks. 
\end{itemize}

\section{Article Details}
This thesis includes materials from four papers, that I wrote as the first author or co-first author. 
This section provides a detailed list of these papers and explanations of my contribution to these papers:

\begin{itemize}
    \item \textbf{Ordered Neurons: Integrating Tree Structures into Recurrent Neural Networks.}
    Yikang Shen, Shawn Tan, Alessandro Sordoni, Aaron Courville.
    \textit{International Conference on Learning Representations, 2019.}
    
    \textit{Personal Contribution.} 
    I proposed the idea of Ordered Neurons; implemented the ON-LSTM model; wrote the introduction, model, and most of the experiment section; conducted the language model, unsupervised constituency parsing, and targeted syntactic evaluation experiments. Shawn Tan was intensively involved in the discussions that results in the idea, wrote the related work section, significantly contributed to the writing of other sections, and conducted the logical inference experiments. Alessandro Sordoni and Aaron Courville co-supervised the project and significantly contributed to the writing.

    \item \textbf{Ordered Memory.} Yikang Shen, Shawn Tan, Arian Hosseini, Zhouhan Lin, Alessandro Sordoni, Aaron Courville.
    \textit{Advances in Neural Information Processing Systems, 2019.}
    
    \textit{Personal Contribution.} 
    I proposed the idea of Ordered Memory; implemented the model; wrote most of the paper; conducted the ListOps experiment.
    Shawn was heavily involved in the discussion about model details; significantly contributed to the paper writing; conducted the logical inference experiment.
    Arian Hosseini conducted the sentiment analysis experiment.
    Zhouhan Lin was heavily involved in the discussion and paper writing. 
    Alessandro Sordoni and Aaron Courville co-supervised the project; were heavily involved in the discussion; contributed to the writing.
    
    \item \textbf{Unsupervised Dependency Graph Network.}
    Yikang Shen, Shawn Tan, Alessandro Sordoni, Peng Li, Jie Zhou, Aaron Courville.
    \textit{Annual Meeting of the Association for Computational Linguistics, 2022.}
    
    \textit{Personal Contribution.}
    I proposed the idea of UDGN; implemented the model; wrote the introduction, model, and most of the experiment section; conducted the language model, unsupervised parsing experiment, and ablation study.
    Shawn Tan wrote the related work section; conducted the language model and finetuning experiments. 
    Alessandro Sordoni and Aaron Courville co-supervised the project, were heavily involved in the discussion, contributed to the writing.
    Peng Li and Jie Zhou contributed to the writing.
    
    \item \textbf{Inducing Reusable Skills From Demonstrations with Option-Controller Network.}
    Siyuan Zhou, Yikang Shen, Yuchen Lu, Aaron Courville, Joshua B. Tenenbaum, Chuang Gan.
    
    \textit{Personal Contribution.}
    I proposed the idea of the Option-Controller Network; implemented the model; wrote the introduction and the model section. 
    Siyuan Zhou conducted the experiments; wrote the experiment section.
    Yuchen Lu and Chuang Gan were heavily involved in the discussion; significantly contributed to the writing. 
    Aaron Courville and Joshua B. Tenebaum co-supervised the project.
\end{itemize}

\chapter{Preliminaries} \label{cha:preliminaries}

\section{Neural Network Components and Architectures}
\subsection{Word embeddings}
The first key idea of word embeddings is to represent words as low-dimensional (e.g., 300), real-valued vectors. 
Before deep learning, it was common to represent a word as an index into the vocabulary, which is a notational variant of using one-hot word vectors: each word is represented as a high-dimensional, sparse vector where only one entry of that word is 1 and all other entires are 0's:
\begin{eqnarray*}
\mathbf{v}_{\text{car}} = [0, 0, \ldots, 0, 0, 1, 0, \ldots, 0]^{\intercal} \\
\mathbf{v}_{\text{vehicle}} = [0, 1, \ldots, 0, 0, 0, 0, \ldots, 0]^{\intercal}
\end{eqnarray*}

The biggest problem with these sparse vectors is that they don't share any semantic similarity between words, i.e., for any pair of different words $a, b$, $\cos(\mathbf{v}_a, \mathbf{v}_b) = 0$. 

Low-dimensional word embeddings effectively alleviated this problem and similar words can be encoded as similar vectors in space: $\cos(\mathbf{v}_{\text{car}}, \mathbf{v}_{\text{vechicle}}) < \cos(\mathbf{v}_{\text{car}}, \mathbf{v}_{\text{man}})$.
These word embeddings can be effectively learned from large unlabeled text corpora, based on the assumption that words occurred in similar contexts tend to have similar meanings (a.k.a. the \textit{distributional hypothesis})~\citep{harris1954distributional}. 
Indeed, learning word embeddings from text has a long-standing history and has been popularized by scalable algorithms and released sets of pretrained word embeddings such as \textsc{word2vec}~\citep{mikolov2013distributed}, \textsc{glove}~\citep{pennington2014glove} and \textsc{fasttext}~\citep{bojanowski2017enriching}.
In modern NLP models \citep{vaswani2017attention, devlin2018bert}, word embeddings are still essential components for representing the input and output tokens. 
But they are optimized together with the other components of neural network models.

\subsection{Recurrent neural networks}
\textit{Recurrent neural networks} are a class of neural networks which are suitable to handle sequences of variable length. More concretely, they apply a parameterized function recursively on a sequence $\mathbf{x}_1, \ldots, \mathbf{x}_n$:
\begin{equation}
    \mathbf{h}_t = f(\mathbf{h}_{t-1}, \mathbf{x}_t; \Theta)
\end{equation}
For NLP applications, we represent a sentence or a paragraph as a sequence of words where each word is transformed into a vector (usually through pre-trained word embeddings): $\mathbf{x} = \mathbf{x}_1, \mathbf{x}_2, \ldots, \mathbf{x}_n \in \R^d$ and $\mathbf{h}_t \in \R^h$ can be used to model the contextual information of $\mathbf{x}_{1:t}$.

Vanilla RNNs take the form:
\begin{equation}
    \mathbf{h}_t = \tanh(\mathbf{W}^{hh}\mathbf{h}_{t-1} + \mathbf{W}^{hx}\mathbf{x}_t + \mathbf{b}),
\end{equation}
where $\mathbf{W}^{hh} \in \R^{h \times h}, \mathbf{W}^{hx} \in \R^{h\times d}$, $\mathbf{b} \in \R^h$ are the parameters to be learned. To ease the optimization, many variants of RNNs have been proposed. Among them, long short-term memory networks (LSTMs)~\citep{hochreiter1997long} and gated recurrent units (GRUs)~\citep{cho2014learning} are the commonly used ones. Arguably, LSTM is still the most competitive RNN variant for NLP applications today and also our default choice for the neural models that we will describe. Mathematically, LSTMs can be formulated as follows:
\begin{eqnarray}
    \mathbf{i}_t & = & \sigma(\mathbf{W}^{ih}\mathbf{h}_{t-1} + \mathbf{W}^{ix}\mathbf{x_t} + \mathbf{b}^{i}) \\
    \mathbf{f}_t & = & \sigma(\mathbf{W}^{fh}\mathbf{h}_{t-1} + \mathbf{W}^{fx}\mathbf{x_t} + \mathbf{b}^{f}) \\
    \mathbf{o}_t & = & \sigma(\mathbf{W}^{oh}\mathbf{h}_{t-1} + \mathbf{W}^{ox}\mathbf{x_t} + \mathbf{b}^{o}) \\
    \mathbf{g}_t & = & \tanh(\mathbf{W}^{gh}\mathbf{h}_{t-1} + \mathbf{W}^{gx}\mathbf{x_t} + \mathbf{b}^{g}) \\
    \mathbf{c}_t & = & \mathbf{f}_t \odot \mathbf{c}_{t-1} + \mathbf{i}_t \odot \mathbf{g}_t \\
    \mathbf{h}_t & = & \mathbf{o}_t \odot \tanh(\mathbf{c}_t),
\end{eqnarray}
where $\mathbf{W}^{ih}, \mathbf{W}^{fh}, \mathbf{W}^{oh}, \mathbf{W}^{gh} \in \R^{h \times h}$, $\mathbf{W}^{ix}, \mathbf{W}^{fx}, \mathbf{W}^{ox}, \mathbf{W}^{gx} \in \R^{h \times d}$ and $\mathbf{b}^{i}, \mathbf{b}^{f}, \mathbf{b}^{o}, \mathbf{b}^{g} \in \R^h$ are the parameters to be learned.

Finally, a useful elaboration of an RNN is a \textit{bidirectional RNN}. The idea is simple: for a sentence or a paragraph, $\mathbf{x} = \mathbf{x}_1, \ldots, \mathbf{x}_n$, a forward RNN is used from left to right and then another backward RNN is used from right to left:
\begin{eqnarray}
    \overrightarrow{\mathbf{h}}_t & = & f(\overrightarrow{\mathbf{h}}_{t-1}, \mathbf{x}_t; \overrightarrow{\Theta}), \quad t = 1, \ldots, n\\
    \overleftarrow{\mathbf{h}}_t & = & f(\overleftarrow{\mathbf{h}}_{t+1}, \mathbf{x}_t; \overleftarrow{\Theta}), \quad t = n, \ldots, 1
\end{eqnarray}
We define $\mathbf{h}_t = [\overrightarrow{\mathbf{h}}_t; \overleftarrow{\mathbf{h}}_t] \in \R^{2h}$ as the concatenation of the hidden vectors from the RNNs in both directions. These representations can usefully encode information from both the left context and the right context and are suitable for general-purpose trainable feature-extracting component of many NLP tasks.

\subsection{Attention mechanism}
The third important component is an attention mechanism. 
It was first introduced in the \textit{sequence-to-sequence} (seq2seq) models \cite{sutskever2014sequence} for neural machine translation \citep{bahdanau2014neural,luong2015effective} and has later been extended to other NLP tasks.

Before introducing attention mechanism, if we want to predict the sentiment of a sentence, or translate a sentence of one language to the other, we apply recurrent neural networks to encode a single sentence (or the source sentence for machine translation): $\h_1, \h_2, \ldots, \h_n$ and use the last time step $\h_n$ to predict the sentiment label or the first word in the target language. 
The label probability or first-word distribution can be modeled by the softmax function:

\begin{equation}
  P(Y = y) =\mathrm{softmax} \left( \mathbf{W}_y\h_n \right) 
  = \frac{\exp(\mathbf{W}_y\h_n)}{\sum_{y'}{\exp\left(\mathbf{W}_{y'}\h_n\right)}}
\end{equation}

This requires the model to be able to compress all the necessary information of a sentence into a fixed-length vector. 
This information bottleneck could result in a loss of details. 
An attention mechanism is designed to solve this problem: instead of squashing all the information into the last hidden vector, it looks at the hidden vectors at all time steps and chooses a subset of these vectors adaptively:
\begin{eqnarray}
    \alpha_i & = & \frac{\exp\left(g(\h_i, \mathbf{w}; \Theta_g)\right)}{\sum_{i'=1}^{n}\exp\left(g(\h_{i'}, \mathbf{w}; \Theta_g)\right)} \label{eq:attention} \\
    \mathbf{c} & = & \sum_{i=1}^{n}{\alpha_i \h_i} \label{eq:context-vector}
\end{eqnarray}

Here $\mathbf{w}$ can be a task-specific vector learned from the training process, or taken as the current target hidden state in machine translation and $g$ is a parameteric function which can be chosen in various ways, such as dot product, bilinear product, or one hidden layer of an MLP:
\begin{eqnarray}
    g_{\text{dot}}(\h_i, \mathbf{w}) &=& {\h_i}^{\intercal}\mathbf{w} \\
    g_{\text{bilinear}}(\h_i, \mathbf{w}) &=& {\h_i}^\intercal\mathbf{W}\mathbf{w} \\
    g_{\text{MLP}}(\h_i, \mathbf{w}) &=& {\mathbf{v}}^\intercal\tanh(\mathbf{W}^h\h_i + \mathbf{W}^w\mathbf{w}) \label{eq:mlp-att}
\end{eqnarray}

An attention mechanism computes a similarity score for each $\h_i$ and then a softmax function is applied which returns a discrete probability distribution over all time steps. Thus $\alpha$ essentially captures which parts of the sentence are relevant and $\mathbf{c}$ aggregates over all the time steps with a weighted sum and can be used for final prediction. 
Interested readers are referred to \cite{bahdanau2014neural,luong2015effective} for more details.

Attention mechanisms have proved widely effective in numerous applications and become an integral part of neural NLP models. 
\cite{parikh2016decomposable} and \cite{vaswani2017attention} conjectured that attention mechanisms don't have to be used in conjunction with recurrent neural networks and can be built purely based on word embeddings and feed-forward networks while providing minimal sequence information. 
This class of models usually requires fewer parameters and is more parallelizable and scalable --- in particular, the \textsc{Transformer} model proposed in \cite{vaswani2017attention} has become a recent trend and our model also has a strong connection to it.
Recent developments of pretrained language model~\citep{devlin2018bert,brown2020language} have made the attention mechanism one of the most important components for modern NLP models.

\subsection{Memory Networks}
A memory network is a neural network extended with a memory component that can be read and written to. 
The model is trained to learn how to operate effectively with the memory component. 
The high-level view of a memory network is as follows:
\begin{itemize}
    \item There is a memory, \textbf{m}, an indexed array of objects (e.g. vectors or arrays of strings).
    \item An input feature map \textbf{I}, which converts the incoming input to the internal feature representation.
    \item A writing component \textbf{W} which updates old memories given the new input. 
    \item An output feature map \textbf{O}, which produces a new output in the feature representation space given the new input and the current memory state.
    \item A response component \textbf{R} which converts the output into the response format desired – for example, a textual response or an action.
\end{itemize}
\textbf{I}, \textbf{W}, \textbf{O} and \textbf{R} can all potentially be learned components and make use of any ideas from the existing machine learning literature.
\begin{figure}[h]
    \centering
    \includegraphics[width=0.6\linewidth]{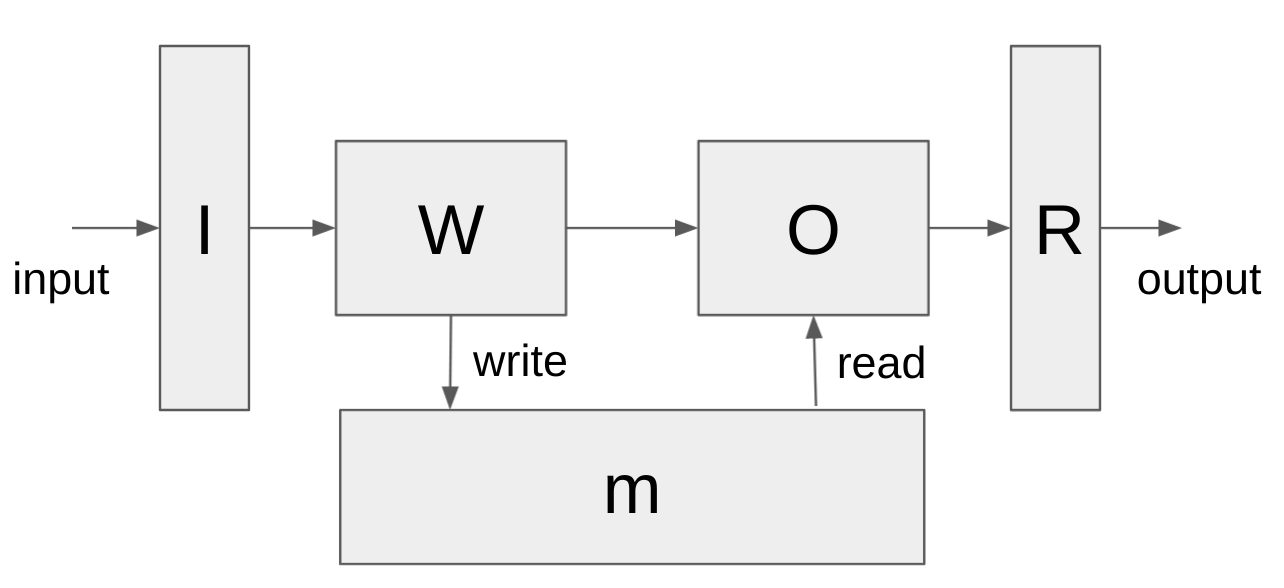}
    \caption{The architecture of a memory network}
    \label{fig:memory_network}
\end{figure}

\textbf{I} can make use of standard pre-processing such as parsing, coreference, and entity resolution. It could also encode the input into an internal feature representation by converting from text to a sparse or dense feature vector.
The simplest form of \textbf{W} is to introduce a function H which maps the internal feature representation produced by \textbf{I} to an individual memory slot and just updates the memory at H(\textbf{I}(x)). 
More sophisticated variants of \textbf{W} could go back and update earlier stored memories (potentially, all memories) based on the new evidence from the current input x. If the input is at the character or word level one could group inputs (i.e., by segmenting them into chunks) and store each chunk in a memory slot.
\textbf{O} Reads from memory and performs inference to deduce the set of relevant memories needed to perform a good response.
\textbf{R} would produce the actual wording of the question-answer based on the memories found by \textbf{O}. 
For example, \textbf{R} could be an RNN conditioned on the output of \textbf{O}.
When the components \textbf{I}, \textbf{W}, \textbf{O} and \textbf{R} are neural networks, the resulting system is a Memory Neural Network (MemNN). 

\subsection{Transformer}
\cite{vaswani2017attention} introduces a novel architecture called a Transformer, that uses the attention-mechanism we saw earlier. 
Like a LSTM-based sequence-to-sequence model, the Transformer is an architecture for transforming one sequence into another one with the help of two parts: an Encoder and a Decoder. 
But it differs from the previously existing sequence-to-sequence models because it does not imply any Recurrent Networks.

\begin{figure}[h]
    \centering
    \includegraphics[width=0.6\linewidth]{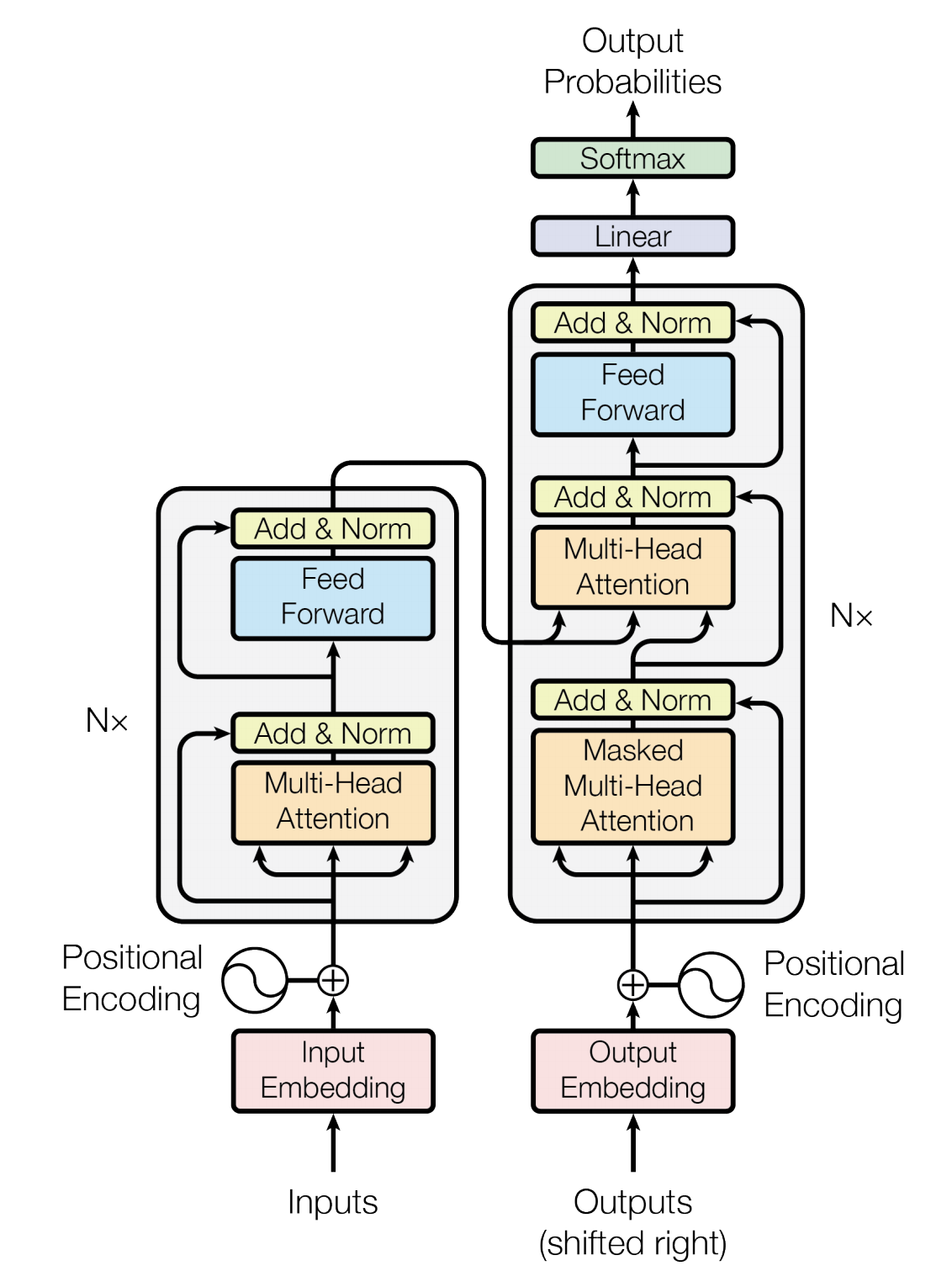}
    \caption{The Transformer architecture \citep{vaswani2017attention}.}
    \label{fig:transformer}
\end{figure}

The Transformer follows this overall architecture using stacked self-attention and point-wise, fully connected layers for both the encoder and decoder, shown in the left and right halves of Figure \ref{fig:transformer}, respectively.

The inputs and outputs (target sentences) are first embedded into an $d_k$-dimensional space.
One slight but important part of the model is the positional encoding of the different words. 
Since the transformer has no recurrent networks that can remember how sequences are fed into a model, we need to inform every word in the sequence of their positions.
These positions are represented by vectors and are added to the embedded representation of each word.

The attention function used in transformer can be described by the following equation:
\begin{equation}
    \mathrm{Attention}(Q,K,V) = \mathrm{softmax} \left( \frac{QK^T}{\sqrt{d_k}} \right) V
\end{equation}
Given the input matrix $X$, $Q=W_Q X$ is a matrix that contains the query, $K=W_K X$ are all the keys, and $V=W_V X$ are the values, $d_k$ is the dimension of word embeddings and hidden states. 
For self-attention, V consists of the same word sequence as Q. 
However, for the intra-attention that takes into account the encoder and the decoder sequences, V and K are different from the sequence represented by Q.

One set of $ \left(W_{Q},W_{K},W_{V}\right) $ matrices and the attention function form an attention head.
Each layer in a Transformer model has multiple attention heads.
While each attention head attends to the tokens that are relevant to each token, with multiple attention heads the model can do this for different definitions of "relevance". 
The computations for each attention head can be performed in parallel, which allows for fast processing. 
The outputs for the attention layer are concatenated to pass into the feed-forward neural network layers.

After the multi-headed attention in both the encoder and decoder, we have a pointwise feed-forward layer. 
This feed-forward network has identical parameters for each position, which can be described as a separate, identical linear transformation of each element from the given sequence.
After each multi-headed attention layer and pointwise feed-forward layer, the Transformer has a skip connection and layer normalization.
Figure~\ref{fig:transformer} shows the architecture of Transformer.

\subsection{Graph Neural Networks}
Graph neural network is first introduced in \cite{scarselli2008graph}.
Consider a graph $G = (V, E)$, where $|V| = N$ is the number of nodes in the graph and $|E| = N^e$ is the number of edges. 
$A \in R^{N \times N}$ is the adjacency matrix.
For graph representation learning, we use $\h_v$ as the hidden state of node $v$.

The graph structures are different from task to task. 
There are usually two scenarios: structural scenarios and non-structural scenarios. 
In structural scenarios, the graph structure is explicit in the applications, such as applications on molecules, physical systems, knowledge graphs, and so on. 
In non-structural scenarios, graphs are implicit so that we have to first build the graph from the task, such as building a fully-connected “word” graph for text or building a scene graph for an image. 
After we get the graph, the later design process attempts to find an optimal GNN model on this specific graph.
In chapter~\ref{cha:dependency}, we propose a third scenario: the model needs to induce the graph structure given an inductive bias.

Although graph structure could be very different, they usually can be categorized from three perspectives:
\begin{itemize}
    \item \textbf{Directed/Undirected Graphs.} 
    Edges in directed graphs are all directed from one node to another, which provides more information than undirected graphs. 
    Each edge in undirected graphs can also be regarded as two directed edges.
    \item \textbf{Homogeneous/Heterogeneous Graphs.} 
    Nodes and edges in homogeneous graphs have the same types, while nodes and edges have different types in heterogeneous graphs. 
    For example, social networks are mostly homogeneous graphs, because all nodes belong to the same type -- user.
    But an author-paper network is a heterogeneous graph because it includes two types of nodes: author and paper.
    Types for nodes play important roles in heterogeneous graphs and should be further considered.
    \item \textbf{Static/Dynamic Graphs.} 
    When input features or the topology of the graph vary with time, the graph is regarded as a dynamic graph. 
    The time information should be carefully considered in dynamic graphs.
\end{itemize}
The dependency graph can be categorized as a directed, homogeneous and static graph.
The direction is usually from a dependent to its head. 
Since the only type of nodes is words, the graph is homogeneous.
And a dependency graph does not change with time.

After identifying the graph type, we need to design the propagation module for a graph neural network.
The propagation module is used to propagate information between nodes so that the aggregated information could capture both feature and topological information. 
In propagation modules, the convolution operator and recurrent operator are usually used to aggregate information from neighbors while the skip connection is used to gather information from previous layers.
In Chapter \ref{cha:dependency}, we will propose a multi-channel competitive mechanism to propagate information that is inspired by dependency functions. 

\subsection{Recursive Neural Network}
A recursive neural network~\citep{socher2010learning} is a kind of deep neural network created by applying the same set of weights recursively over a structured input, to produce a structured prediction over variable-size input structures, or a scalar prediction on it, by traversing a given structure in topological order. 
Recursive neural networks, sometimes abbreviated as RvNNs, have been successful, for instance, in learning sequence and tree structures in natural language processing, where they are used mainly to learn continuous representations of phrases and sentences based on word embedding. 
RvNNs have first been introduced to learn distributed representations of structure, such as logical terms.

In the most simple architecture, nodes are combined into parents using a weight matrix that is shared across the whole network, and a non-linearity such as tanh. If $c_1$ and $c_2$ are n-dimensional vector representation of nodes, their parent will also be an n-dimensional vector, calculated as
\begin{equation}
    p_{1,2}=\tanh \left(W[c_{1};c_{2}]\right)
\end{equation}
Where W is a learned $n\times 2n$ weight matrix.

\section{Tasks for Evaluating Syntactic Inductive Biases}
\subsection{Language Modeling}
Language modeling (LM) is the use of various statistical and probabilistic techniques to determine the probability of a given sequence of words occurring in a sentence.
Given such a sequence, say of length m, a language model usually assigns a probability $P(w_{1},\ldots ,w_{m})$ to the whole sequence.
In this thesis, we use word-level language modeling task for two purposes: 
1) an unsupervised training method for other tasks, including unsupervised parsing and syntactic evaluation; 
2) a macroscopic evaluation of the model's ability to deal with various linguistic phenomena (e.g. co-occurrence, syntactic structure, verb-subject agreement, etc).
We expect that a good syntactic inductive bias should improve the model's performance in modeling natural language.

In recent years, two language modeling tasks have been widely used.
The first is \textbf{recurrent language modeling}, which requires that the model predicts the next token based on the previous context.
\begin{equation}
    p(w_t | w_{<t}) = f ( w_1, ..., w_{t-1})
\end{equation}
where $f(\cdot)$ is a neural network that uses word embeddings to make its predictions.
In this way, one can easily compute the probability for the entire sentence or document.
\begin{equation}
    P(w_{1},\ldots ,w_{m}) = p(w_1) p(w_1 | w_2) ... p(w_m | w_1, ..., w_{m-1})
\end{equation}
To train a recurrent language model, we can use gradient descent methods to optimize the log probability.
The performance of a recurrent language model is quantified by perplexity ($ppl$):
\begin{equation}
    ppl(w_{1},\ldots ,w_{m}) = P(w_{1},\ldots ,w_{m})^{-\frac{1}{m}} = \exp \left( -\frac{1}{m} \sum_{i=1}^m \log p(w_i | w_1, ..., w_{i-1}) \right)
\end{equation}
During the evaluation, we compare the perplexity of different models on the test set.

The other task is \textbf{masked language modeling}, which is widely used for pretraining on a large-scale dataset.
Under Masked Language Modelling, we typically mask a certain percentage of words in a given sentence and the model is expected to predict those masked words based on other words in that sentence.
Such a training scheme makes this model bidirectional in nature because the representation of the masked word is learned based on the words that occur to its left as well as its right.
\begin{equation}
    p(w_{masked} | w_{unmasked}) = f(w_{unmasked})
\end{equation}
Similar to the recurrent language model, a masked language model can also be optimized with gradient descent.
But it's not straightforward to compute the sentence or document probability from a masked language model.
The precision of prediction also attracts less interest.
Because the task is primarily used as a pre-training method before fine-tuning on downstream tasks.
However, for the same architecture, models with better pretraining loss usually have better fine-tuning performance \citep{devlin2018bert,liu2019roberta}.

\subsection{Unsupervised Parsing}
Work on unsupervised parsing (also known as grammar induction) attempts to find methods for syntactic parsing that do not require expensive and difficult-to-design expert labeled treebanks for training \citep{carroll1992two, klein2002generative, smith2005guiding}. 
Recent work on latent tree learning offers a new family of approaches to the problem \citep{yogatama2016learning, maillard2017jointly, choi2018learning}.
Latent tree learning models attempt to induce syntactic structure using the supervision from a non-parsing task such as language modeling and textual entailment.
Syntactic inductive biases can be used as tools to solve unsupervised parsing, while unsupervised parsing tasks are also good ways to examine the correlation between human-annotated structures and model-induced structures.
There are two major unsupervised parsing tasks: 1) \textbf{unsupervised constituency parsing} and 2) \textbf{unsupervised dependency parsing}.

\textbf{Unsupervised constituency parsing} compares the predicted constituency trees with human annotations.
The performance of an unsupervised constituency parser is often evaluated with the Unlabeled F1 (UF1) score.
It calculates the overlap of constituents proposed by the reference tree and induced tree:
\begin{equation}
    R = \frac{\text{\# of correct constituents in induced tree of s}}{\text{\# of constituents in reference tree of s}}
\end{equation}
\begin{equation}
    P = \frac{\text{\# of correct constituents in induced tree of s}}{\text{\# of constituents in induced tree of s}}
\end{equation}
\begin{equation}
    \text{UF1} = \frac{2PR}{P + R}
\end{equation}

\textbf{Unsupervised dependency parsing} compares the predicted dependency graphs with human annotations.
Unlabeled Attachment Score (UAS) is a standard evaluation metric in dependency parsing: the percentage of words that are assigned the correct syntactic head.
\begin{equation}
    \text{UAS} = \frac{\text{\# of words with correct heads in induced graph of s}}{\text{\# of words in s}}
\end{equation}

\subsection{Syntactic Evaluation}
Syntactic evaluations are designed for evaluating the grammaticality of the predictions of a language model.
The method provides hand-crafted minimal pairs of sentences that differ only in the main verb’s conjugation, then evaluates whether language models rate each grammatical sentence as more likely than its ungrammatical counterpart.
For example, given the two strings ``\textit{The keys to the cabinet are on the table}'' and ``\textit{The keys to the cabinet is on the table}'', a model that has learned the proper subject-verb number agreement rules for English should assign a higher probability to the grammatical plural verb in the first sentence than to the ungrammatical singular verb in the second \citep{linzen2016assessing}.
This method can be used to evaluate many different syntactic phenomenons, including Agreement, Licensing, Garden-Path Effects, Gross Syntactic Expectation, Center Embedding, and Long-Distance Dependencies \citep{hu2020systematic}.

\subsection{Synthetic Tasks}
Studying the parsing ability of syntactic inductive biases in natural language can be challenging due to the inherent complexities of natural language, like having several valid parses for a single sentence.
Another way to check whether a model can understand latent structure is using synthetic data generated from well-formed grammar.
There are three major advantages of using synthetic data:
1) the ground-truth structure is known and unique, so the parsing result can be accurately evaluated;
2) the solution of such task usually relies on the latent structure, which provides a strong pressure for the model to learn the grammar;
3) given the well-formed grammar, one can easily generate new data to evaluate the model's generalization ability.
Given these three advantages, several synthetic datasets are proposed to evaluate latent tree models.
They are usually in the form of logical or mathematical expressions.
\citet{nangia2018listops} proposed ListOps, which is in the style of prefix arithmetic. 
It is comprised of deeply nested lists of mathematical operations and a list of single-digit integers.
The task requires the model to compute the result of a given equation.
\citet{bowman2014recursive} proposed a logical inference dataset, comprised of propositional logical statements.
The task requires the model to reason over two logical statements and compute their relations.

\chapter{Constituency Inductive Bias: Ordered Neurons}
\label{cha:constituency}

A constituency grammar is hierarchically structured: smaller units (e.g., phrases) are nested within larger units (e.g., clauses). 
When a larger constituent ends, all of the smaller constituents that are nested within it must also be closed. 
While the standard recurrent networks and memory network architectures allows different neurons (memory slots) to track information at different time scales, it does not have an explicit bias towards modeling the hierarchy of constituents. 
In this chapter, we will introduce a constituency inductive bias and two instantiations of the idea. 

We start by reviewing the history of constituency-related models (Section~\ref{sec:constituency_prev}).
In Section~\ref{sec:ordered_neurons}, we propose ordered neurons -- a inductive bias that models the hierarchy of constituents through the assigning of an order to neurons.
We then proposed two models as instantiations of the inductive bias.
In Section~\ref{sec:on-lstm} and Section~\ref{sec:nle}, we propose ON-LSTM -- a variant of the LSTM model.
It use a vector of master input and forget gates to ensures that when a given neuron is updated, all the neurons that follow it in the ordering are also updated.
In Section~\ref{sec:ordered_memory} and Section~\ref{sec:fle}, we propose the Ordered Memory. 
It use a soft stack mechanism to enforce an order in memory slots.
Finally, we summarize recent advances in Section~\ref{constituency_conclusion}.

\section{Previous Approaches} \label{sec:constituency_prev}
\subsection{Constituency-Augmented Neural Networks}
Theoretically, RNNs and LSTMs can model data produced by context-free grammars and context-sensitive grammars \citep{gers2001lstm}.
However, recent results suggest that introducing structure information into neural networks is beneficial. \cite{kuncoro2018lstms} showed that RNNGs~\citep{dyer2016recurrent}, which have an explicit bias to model the syntactic structures, outperform LSTMs on the subject-verb agreement task~\citep{linzen2016assessing}.
In this thesis, we use a more extensive suite of grammatical tests recently provided by \cite{marvin2018targeted} to evaluate the grammar acceptability of our model.
\cite{bowman2014recursive,bowman2015tree} also demonstrate that tree-structured models are more effective for downstream tasks whose data was generated by recursive programs.
Interestingly,~\cite{shi2018tree} suggests that while the prescribed grammar tree may not be ideal, some sort of hierarchical structure, perhaps task dependent, might help.
\cite{kuncoro2020syntactic} shows that syntactic biases help large scale pre-trained models, like BERT, to achieve better language understanding.

\paragraph{\textbf{Recursive Neural Networks}}
There has been prior work leveraging tree structures for natural language tasks in the literature.
\cite{socher2010learning, alvarez2016tree, zhou2017generative, zhang2015top} use supervised learning on expert-labeled treebanks for predicting parse trees.
~\cite{socher2013recursive}, ~\cite{tai2015improved} and \cite{zhu2015long}, explicitly model the tree-structure using parsing information from an external parser.
Later,~\cite{bowman2016fast} exploited guidance from a supervised parser~\citep{klein2003accurate} in order to train a stack-augmented neural network.

Composition with recursive structures has been shown to work well for certain types of tasks.~\cite{pollack1990recursive} first suggests their use with distributed representations.
Later,~\cite{socher2013recursive} shows their effectiveness on sentiment analysis tasks.
Recent work has demonstrated that recursive composition of sentences is helpful for systematic generalisation \citep{bowman2015tree,bahdanau2018systematic}. \cite{kuncoro2018lstms} also demonstrate that architectures like RNNG~\citep{dyer2016recurrent} handle syntax-sensitive dependencies better for language-related tasks.

\paragraph{\textbf{Stack-Augmented Neural Networks}}
\cite{schutzenberger1963context} first showed an equivalence between push-down automata (stack-augmented automatons) and context-free grammars. \citet{knuth1965translation} introduced the notion of a shift-reduce parser that uses a stack for a subset of formal languages that can be parsed from left to right. This technique for parsing has been applied to natural language as well:~\cite{shieber1983sentence} applies it to English, using assumptions about how native English speakers parse sentences to remove ambiguous parse candidates. More recently, \cite{maillard2017jointly} shows that a soft tree could emerge from all possible tree structures through back propagation.

The idea of using neural networks to control a stack is not new.~\cite{zeng1994discrete} uses gradient estimates to learn to manipulate a stack using a neural network.~\cite{das1992learning} and~\cite{mozer1993connectionist} introduced the notion of a \emph{continuous stack} in order for the model to be fully differentiable.
Much of the recent work with stack-augmented networks built upon the development of neural attention \citep{graves2013generating,bahdanau2014neural,weston2014memory}.
\cite{graves2014neural} proposed methods for reading and writing using a head, along with a ``soft'' shift mechanism. Apart from using attention mechanisms, \cite{grefenstette2015learning} proposed a neural stack where the push and pop operations are made to be differentiable, which worked well in synthetic datasets. \cite{yogatama2016learning} proposes RL-SPINN where the discrete stack operations are directly learned by reinforcement learning.

\subsection{Constituency Inductive Biases}
The task of learning the underlying grammar from data is known as~\emph{grammar induction} ~\citep{chen1995bayesian, cohen2011unsupervised}. 
Early work incorporated syntactic structure in the context of language modeling~\citep{roark2001probabilistic, charniak2001immediate, chelba2000structured}.
More recently, there have been attempts at incorporating some structure for downstream tasks using neural models~\citep{grefenstette2015learning, sun2017neural,joulin2015inferring}.
Generally, these works augment a main recurrent model with a stack and focus on solving algorithmic tasks.
\cite{yogatama2018memory} focus on language modeling and syntactic evaluation tasks \citep{linzen2016assessing} but they do not show the extent to which the structure learnt by the model align with gold-standard parse trees. \cite{shen2017neural} introduced the Parsing-Reading-Predict Networks (PRPN) model, which attempts to perform parsing by solving a language modeling task. The model uses self-attention to compose previous states, where the range of attention is controlled by a learnt ``syntactic distance''. The authors show that this value corresponds to the depth of the parse tree. However, the added complexity in using the PRPN model makes it unwieldy in practice.

Another possible solution is to develop models with varying time-scales of recurrence as a way of capturing this hierarchy. \cite{el1996hierarchical, schmidhuber1991neural, lin1998learning} describe models that capture hierarchies at pre-determined time-scales.
More recently, \cite{koutnik2014clockwork} proposed Clockwork RNN, which segments the hidden state of a RNN by updating at different time-scales, while \cite{xu2016cached} rescales the forget gates at different pre-determined scales.
These approaches typically make a strong assumption about the regularity of the hierarchy involved in modelling the data.
\cite{chung2016hierarchical} proposed a method that, unlike the Clockwork RNN, would learn a multi-scale hierarchical recurrence.
However, the model still has a pre-determined depth to the hierarchy, depending on the number of layers. 
\cite{rippel2014learning} proposed to encourage a hierarchy in the representation units by applying ``nested'' dropout masks: units are not dropped independently at random but whenever a unit is dropped, all the units that follow in the ordering are also dropped. 
Our work can be seen as a soft and controllable version of the dropout, that we apply a monotonically decreasing mask to the hidden state.
Moreover, we propose to condition the update masks on the particular input and apply our overall model to sequential data. Therefore, our model can adapt the structure to the observed data, while both Clockwork RNN and nested dropout impose a predefined hierarchy to hidden representations.

\section{Ordered Neurons}\label{sec:ordered_neurons}
\begin{figure}[h]
\centering
\includegraphics[width=1.\linewidth]{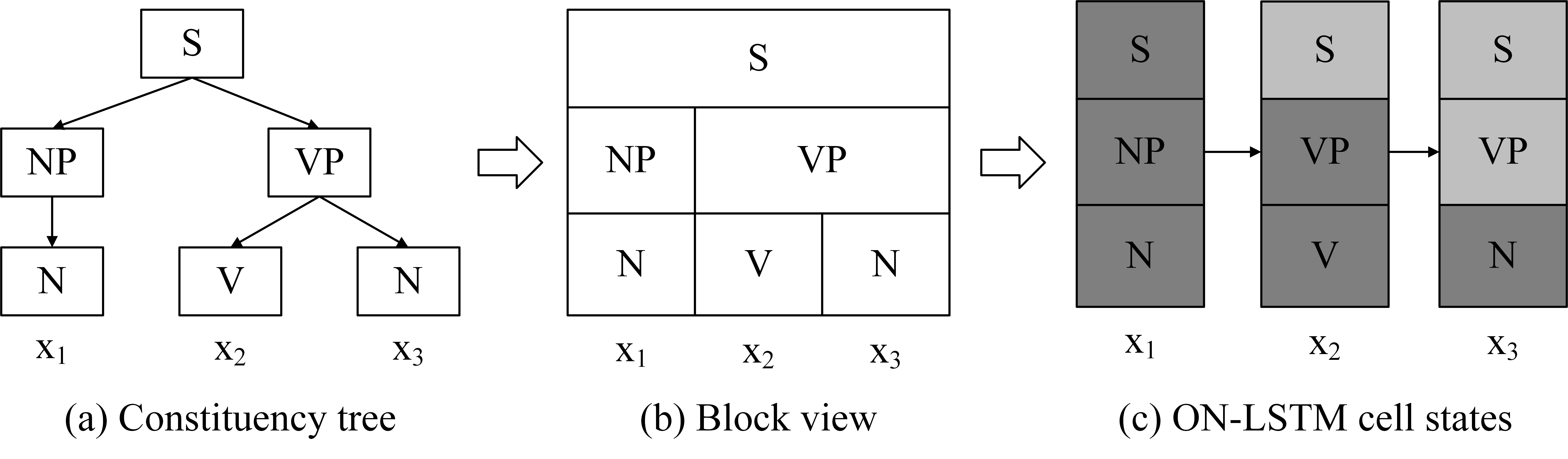}
\caption{
An instantiation of Ordered Neurons: ON-LSTM
A sequence of tokens $S = (x_1, x_2, x_3)$ and its corresponding constituency tree are illustrated in (a).
We provide a block view of the tree structure in (b), where both~$\mathrm{S}$ and~$\mathrm{VP}$ nodes span more than one time-step.
The representation for high-ranking nodes should be relatively consistent across multiple time-steps.
(c) Visualization of the update frequency of groups of hidden state neurons. At each time-step, given the input word, dark grey blocks are completely updated while light grey blocks are partially updated.
The three groups of neurons have different update frequencies. Topmost groups update less frequently while lower groups are more frequently updated.
}
\label{fig:tree}
\end{figure}

Given a sequence of tokens $S = (x_1, \ldots, x_T)$ and its corresponding constituency tree~(Figure~\ref{fig:tree}(a)), our goal is to infer the unobserved tree structure while processing the observed sequence,~i.e. while computing the hidden state $h_t$ for each time-step $t$.
At each time-step, $h_t$ would ideally contain information about all the nodes on the path between the current leaf node $x_t$ and the root $\mathrm{S}$. In Figure~\ref{fig:tree}(c), we illustrate how $h_t$ would contain information about all the constituents that include the current token $x_t$ even if those are only partially observed.
This intuition suggests that each node in the tree can be represented by a set of neurons in the hidden states. However, while the dimensionality of the hidden state is fixed in advance, the length of the path connecting the leaf to the root of the tree may be different across different time-steps and sentences. Therefore, a desiderata for the model is to dynamically reallocate the dimensions of the hidden state to each node.

Given these requirements, we introduce \emph{ordered neurons}, an inductive bias that forces neurons to represent information at different time-scales. In our model, high-ranking neurons contain long-term or global information that will last anywhere from several time-steps to the entire sentence, representing nodes near the root of the tree. Low-ranking neurons encode short-term or local information that only last one or a few time-steps, representing smaller constituents, as shown in Figure~\ref{fig:tree}(b). The differentiation between high-ranking and low-ranking neurons is learnt in a completely data-driven fashion by controlling the update frequency of single neurons: to erase (or update) high-ranking neurons, the model should first erase (or update) all lower-ranking neurons. In other words, some neurons always update more (or less) frequently than the others, and that order is pre-determined as part of the model architecture.

\section{ON-LSTM}\label{sec:on-lstm}
While natural languages' grammar remains an open question to study, many neural network models already show strong performance on many NLP tasks.
Thus, the most reasonable next step is adding the structural inductive bias to an established neural network model to improve its grammar acceptability.
In this section, we present ON-LSTM (``\emph{ordered neurons} LSTM'') -- a LSTM variant augmented with constituency inductive bias.
The difference with the LSTM is that we replace the update function for the cell state $c_t$ with a new function that will be explained in the following sections.
The forget gates $f_t$ and input gates $i_t$ are used to control the erasing and writing operation on cell states $c_t$, as before.
Since the gates in the LSTM act independently on each neuron, it may be difficult in general to discern a hierarchy of information between the neurons. To this end, we propose to make the gate for each neuron dependent on the others by enforcing the order in which neurons should be updated.

\subsection{Activation Function: $\cummax()$}
To enforce an order to the update frequency, we introduce a new activation function:
\begin{align}
    \hat{g} = \cummax (\ldots) = \mathrm{cumsum}(\mathrm{softmax}(\ldots))
\end{align}
where $\mathrm{cumsum}$ denotes the cumulative sum.
We will show that the vector $\hat{g}$ can be seen as the expectation of a binary gate $g=(0,...,0,1,...,1)$. This binary gate splits the cell state into two segments: the 0-segment and the 1-segment. Thus, the model can apply different update rules on the two segments to differentiate long/short-term information. Denote by~$d$ a categorical random variable representing the index for the first $1$ in $g$:
\begin{align}
    p(d) &= \mathrm{softmax}(\ldots)
\end{align}
The variable $d$ represents the split point between the two segments. We can compute the probability of the $k$-th value in $g$ being 1 by evaluating the probability of the disjunction of any of the values before the $k$-th being the split point, that is $d \leq k = (d=0) \vee (d=1) \vee \cdots \vee (d=k)$. Since the categories are mutually exclusive, we can do this by computing the cumulative distribution function:
\begin{align}
    p(g_k=1) = p(d \leq k) &= \sum_{i \leq k} p(d = i)
\end{align}
Ideally, $g$ should take the form of a discrete variable.
Unfortunately, computing gradients when a discrete variable is included in the computation graph is not trivial~\citep{schulman2015gradient}, so in practice we use a continuous relaxation by computing the quantity $p(d \leq k)$, obtained by taking a cumulative sum of the softmax.
As $g_k$ is binary, this is equivalent to computing $\mathbb{E}[g_k]$. Hence, $\hat{g} = \mathbb{E}[g]$. 

\subsection{Structured Gating Mechanism}
Based on the $\cummax()$ function, we introduce a master forget gate $\tilde{f}_{t}$ and a master input gate $\tilde{i}_{t}$:
\begin{align}
    \tilde{f}_{t} &= \cummax (W_{\tilde{f}}x_{t}+U_{\tilde{f}}h_{t-1}+b_{\tilde{f}}) \label{eq:masterforget} \\
    \tilde{i}_{t} &= 1 - \cummax (W_{\tilde{i}}x_{t}+U_{\tilde{i}}h_{t-1}+b_{\tilde{i}}) \label{eq:masterinput}
\end{align}
Following the properties of the $\cummax()$ activation, the values in the master forget gate are monotonically increasing from 0 to 1, and those in the master input gate are monotonically decreasing from 1 to 0.
These gates serve as high-level control for the update operations of cell states. Using the master gates, we define a new update rule:
\begin{align}
    \omega_t &= \tilde{f}_t \circ \tilde{i}_t \label{eq:overlap} \\
    \hat{f}_t &= f_t \circ \omega_t + (\tilde{f}_t - \omega_t) = \tilde{f}_t \circ (f_t \circ \tilde{i}_t + 1 - \tilde{i}_t) \label{eq:onforget} \\
    \hat{i}_t &= i_t \circ \omega_t + (\tilde{i}_t - \omega_t) = \tilde{i}_t \circ (i_t \circ \tilde{f}_t + 1 - \tilde{f}_t) \label{eq:oninput} \\
    c_{t} &= \hat{f}_{t}\circ c_{t-1} + \hat{i}_{t}\circ \hat{c}_{t} \label{eq:onupdate}
\end{align}
where $f_t$ and $i_t$ are the original forget and input gate in the LSTM.

In order to explain the intuition behind the new update rule, we assume that the master gates are binary:
\begin{itemize}
    \item The master forget gate $\tilde{f}_t$ controls the erasing behavior of the model. Suppose $\tilde{f}_t=(0,\dots,0,1,\dots,1)$ and the split point is $d^f_t$. Given the Eq. (\ref{eq:onforget}) and (\ref{eq:onupdate}), the information stored in the first $d^f_t$ neurons of the previous cell state $c_{t-1}$ will be completely erased.
    In a parse tree (e.g. Figure~\ref{fig:tree}(a)), this operation is akin to closing previous constituents.
    A large number of zeroed neurons,~i.e. a large $d^f_t$, represents the end of a high-level constituent in the parse tree, as most of the information in the state will be discarded. Conversely, a small $d^f_t$ represents the end of a low-level constituent as high-level information is kept for further processing.
    
    \item The master input gate $\tilde{i}_t$ is meant to control the writing mechanism of the model. 
    Assume that $\tilde{i}_t = (1,\dots,1,0,\dots,0)$ and the split point is $d^i_t$. 
    a large $d^i_t$ means that the current input $x_t$ contains long-term information that needs to be preserved for several time-steps.
    Conversely, a small $d^i_t$ means that the current input $x_t$ just provides local information that could be erased by $\tilde{f}_t$ in the next few time-steps.
    
    \item The product of the two master gates $\omega_t$ represents the overlap of $\tilde{f}_t$ and $\tilde{i}_t$.
    Whenever an overlap exists ($\exists k, \omega_{tk} > 0$), the corresponding segment of neurons encodes the incomplete constituents that contain some previous words and the current input word $x_t$.
    Since these constituents are incomplete, we want to update the information inside the respective blocks. 
    The segment is further controlled by the $f_t$ and $i_t$ in the standard LSTM model to enable more fine-grained operations within blocks.
    For example, in Figure~\ref{fig:tree}, the word $x_3$ is nested into the constituents $\mathrm{S}$ and $\mathrm{VP}$. At this time-step, the overlap gray blocks would represent these constituents, such that $\tilde{f}_t$ and $\tilde{i}_t$ can decide whether to reset or update each individual neurons in these blocks.
\end{itemize}

As the master gates only focus on coarse-grained control, modeling them with the same dimensions as the hidden states is computationally expensive and unnecessary.
In practice, we set $\tilde{f}_{t}$ and $\tilde{i}_{t}$ to be $D_m = \frac{D}{C}$ dimensional vectors, where $D$ is the dimension of hidden state, and $C$ is a chunk size factor.
We repeat each dimension $C$ times, before the element-wise multiplication with $f_t$ and $i_t$. The downsizing significantly reduces the number of extra parameters that we need to add to the LSTM. 
Therefore, every neuron within each $C$-sized chunk shares the same master gates.

\subsection{Syntactic Distance}
We first proposed Syntactic distance in \cite{shen2018straight} to quantify the process of splitting sentences into smaller constituents.
\begin{deff} \label{def:distance}
Let $\tree$ be a constituency tree for sentence $(w_1, ..., w_n)$.
The height of the lowest common ancestor for consecutive words $x_i$ and $x_{i+1}$ is $\Tilde{\tau}_i$. 
Syntactic distances $\distances=(\tau_1, ..., \tau_{n-1})$ are defined as a sequence of $n-1$ real scalars that share the same rank as $(\tilde{\tau}_1, ..., \tilde{\tau}_{n-1})$.
\end{deff}
In other words, each syntactic distance $d_i$ is associated with a split point $(i,i+1)$ and specify the relative order in which the sentence will be split into smaller components. 
Thus, any sequence of $n-1$ real values can unambiguously map to an unlabeled binary constituency tree with $n$ leaves through Algorithm \ref{alg:distance2tree} \citep{shen2018straight}.
As \cite{shen2018ordered, shen2018neural, wang2019tree} pointed out, the syntactic distance reflects the information communication between constituents. 
More concretely, a large syntactic distance $\tau_i$ represents that short-term or local information should not be communicated between $(x_{\leq i})$ and $(x_{>i})$.
While cooperating with appropriate neural network architectures, we can leverage this feature to build unsupervised constituency parsing models.

\begin{algorithm}[h]
    \centering
    \caption{Distance to binary constituency tree}\label{alg:distance2tree}
    \begin{algorithmic}[1]
    \Function{Constituent}{$\mathbf{w}$, $\mathbf{d}$}
    	\If {$\mathbf{d} = []$}
        	\State {$\tree \Leftarrow$ Leaf($\mathbf{w}$)}
        \Else
        	\State {$i \Leftarrow \mathrm{arg}\max_i (\mathbf{d})$}
            \State {$\mathrm{child}_l$ $\Leftarrow$ Constituent($\mathbf{w}_{\leq i}$, $\mathbf{d}_{<i}$)}
            \State {$\mathrm{child}_r$ $\Leftarrow$ Constituent($\mathbf{w}_{> i}$, $\mathbf{d}_{>i}$)}
            \State $\tree \Leftarrow \mathrm{Node}(\mathrm{child}_l, \mathrm{child}_r)$
        \EndIf
        \State \Return $\tree$
    \EndFunction
    \end{algorithmic}
\end{algorithm}

In ON-LSTM and other methods developed from Ordered Neurons, the syntactic distance has a more specific meaning.
It represent the amount of low-level information that has been erased at each steps.
To infer the tree structure of a sentence from a pre-trained model, we initialize the hidden states with the zero vector, then feed the sentence into the model as done in the language modeling task. At each time step, we compute an estimate of $d^f_t$:
\begin{equation}
    \hat{d}^f_t = \mathbb{E} \left[ d^f_t \right] = \sum_{k=1}^{D_m} k p_f(d_t=k) = \sum_{k=1}^{D_m} \sum_{i=1}^{k} p_f(d_t=k) = D_m - \sum_{k=1}^{D_m} \tilde{f}_{tk}
\end{equation}
where $p_f$ is the probability distribution over split points associated to the master forget gate and $D_m$ is the size of the hidden state. Given $\hat{d}^f_t$, we can use the top-down greedy parsing algorithm proposed in~\cite{shen2017neural} for unsupervised constituency parsing. We first sort the $\{\hat{d}^f_t\}$ in decreasing order. For the first $\hat{d}^f_i$ in the sorted sequence, we split the sentence into constituents $((x_{<i}), (x_{i}, (x_{>i})))$. Then, we recursively repeat this operation for constituents $(x_{<i})$ and $(x_{>i})$, until each constituent contains only one word.

\section{Natural Language Experiments} \label{sec:nle}
In this section, we study the performance of ON-LSTM in natural language settings.
We first train and evaluate the model on language modeling task.
We then compared the syntactic structures induced by ON-LSTM with human-annotated structures.
Finally, We test its grammar acceptability on a synthetic dataset.
Results show that the proposed inductive bias can indeed improve model's grammar acceptability while preserve the strong performance on language modeling.

\subsection{Language Modeling}
Word-level language modeling is a macroscopic evaluation of the model's ability to deal with various linguistic phenomena (e.g. co-occurence, syntactic structure, verb-subject agreement, etc).
We evaluate our model by measuring perplexity on the Penn TreeBank (PTB)~\citep{marcus1993building, mikolov2012statistical} task.

For fair comparison, we closely follow the model hyper-parameters, regularization and optimization techniques introduced in AWD-LSTM \citep{merityRegOpt}.
Our model uses a three-layer ON-LSTM model with 1150 units in the hidden layer and an embedding of size 400. 
For master gates, the downsize factor $C=10$.
The total number of parameters is 25 millions, which is 1 millions more then the AWD-LSTM.
The extra parameters are used for computing the master gates.
We manually searched some of the dropout values for ON-LSTM based on the validation performance.
The values used for dropout on the word vectors, the output between LSTM layers, the output of the final LSTM layer, and embedding dropout where (0.5, 0.3, 0.45, 0.1) respectively. 
A weight-dropout of 0.45 was applied to the recurrent weight matrices.

\begin{table}[h]
\small
\centering
\scalebox{0.85}{
\begin{tabular}{l|ccc}
\toprule
\bf Model & \bf Parameters & \bf Validation &  \bf Test \\
\midrule
\cite{zaremba2014recurrent} - LSTM (large) & 66M & $82.2$ & $78.4$ \\
\cite{gal2016theoretically} - Variational LSTM (large, MC) & 66M & $-$ & $73.4$ \\
\cite{kim2016character} - CharCNN & 19M & $-$ & $78.9$ \\
\cite{merity2016pointer} - Pointer Sentinel-LSTM & 21M & $72.4$ & $70.9$ \\
\cite{grave2016improving} - LSTM & $-$ & $-$ & $82.3$ \\
\cite{grave2016improving} - LSTM + continuous cache pointer & $-$ & $-$ & $72.1$ \\
\cite{inan2016tying} - Variational LSTM (tied) + augmented loss & 51M & $71.1$ & $68.5$ \\
\cite{zilly2016recurrent} - Variational RHN (tied) & 23M & $67.9$ & $65.4$ \\
\cite{zoph2016neural} - NAS Cell (tied) & 54M & $-$ & $62.4$ \\
\cite{shen2017neural} - PRPN-LM & $-$ & $-$ & $62.0$ \\
\cite{melis2017state} - 4-layer skip connection LSTM (tied) & 24M & $60.9$ & $58.3$ \\
\cite{merityRegOpt} - AWD-LSTM - 3-layer LSTM (tied) & 24M & $60.0$ & $57.3$ \\
\midrule
\textbf{ON-LSTM} - 3-layer (tied) & 25M & $\bf 58.29 \pm 0.10$ & $\bf 56.17 \pm0.12$ \\
\midrule
\cite{yang2017breaking} - AWD-LSTM-MoS* & 22M & $56.5$ & $54.4$ \\
\bottomrule
\end{tabular}
}
\caption{
Single model perplexity on validation and test sets for the Penn Treebank language modeling task.
Models labelled \textit{tied} use weight tying on the embedding and softmax weights \citep{inan2016tying, press2017using}.
Models labelled * focus on improving the softmax component of RNN language model. Their contribution is orthogonal to ours.
}
\label{table:PTBresults}
\end{table}

As shown in Table~\ref{table:PTBresults}, our model performs better than the standard LSTM while sharing the same number of layers, embedding dimensions, and hidden states units.
Recall that the master gates only control how information is stored in different neurons. It is interesting to note that we can improve the performance of a strong LSTM model without adding skip connections or a significant increase in the number of parameters.

\subsection{Unsupervised Constituency Parsing}
The unsupervised constituency parsing task compares the latent stree structure induced by the model with those annotated by human experts.
Following the experiment settings proposed in \cite{htut2018grammar}, we take our best model for the language modeling task, and test it on WSJ10 dataset and WSJ test set.
WSJ10 has 7422 sentences, filtered from the WSJ dataset with the constraint of 10 words or less, after the removal of punctuation and null elements \citep{klein2002generative}.
The WSJ test set contains 2416 sentences with various lengths.
It is worth noting that the WSJ10 test set contains sentences from the training, validation, and test set of the PTB dataset, while WSJ test uses the same set of sentences as the PTB test set.

\begin{table}[h]
\small
\centering
\scalebox{0.85}{
\setlength{\tabcolsep}{2.8 pt} 
\begin{tabular}{lllccccccc}
\toprule
 & \bf \multirow{3}{*}{\shortstack[l]{Training \\Data}} & \bf \multirow{3}{*}{\shortstack[l]{Vocab \\Size}} & \multicolumn{2}{c}{\bf Parsing F1} & \bf \multirow{3}{*}{\shortstack[l]{Depth \\WSJ}} & \multicolumn{4}{c}{\bf \multirow{2}{*}{\shortstack{Accuracy on WSJ by Tag}}} \\ 
\bf Model & & & \bf WSJ10 & \bf WSJ & & \bf \multirow{2}{*}{\shortstack{ADJP}} & \bf \multirow{2}{*}{\shortstack{NP}} & \bf \multirow{2}{*}{\shortstack{PP}} & \bf \multirow{2}{*}{\shortstack{INTJ}} \\
\bf  &  &  & \bf $\mu\:(\sigma)$ & $\mu\:(\sigma)$ &  &   &   &   &   \\
 \midrule
PRPN-UP &  AllNLI Train & 76k & 66.3 (0.8) &  38.3 (0.5) & 5.8 & 28.7 & 65.5 & 32.7 &  \it 0.0 \\
PRPN-LM & AllNLI Train & 76k & 52.4 (4.9) &  35.0 (5.4) & 6.1 & 37.8 & 59.7 & \bf 61.5 & \bf 100.0 \\
\midrule
PRPN-UP  & WSJ Train & 15.8k &  62.2 (3.9) &  26.0 (2.3) & 5.8 & 24.8 & 54.4 & 17.8 &  \it 0.0 \\
PRPN-LM & WSJ Train & 10k & 70.5 (0.4) &  37.4 (0.3) & 5.9 & 26.2 &  \bf 63.9 & 24.4 &   \it 0.0 \\
\midrule
\textbf{ON-LSTM} 1st-layer & WSJ Train & 10k & 35.2 (4.1) & 20.0 (2.8) & 5.6 & 38.1 & 23.8 & 18.3 & \bf 100.0  \\
\textbf{ON-LSTM} 2nd-layer & WSJ Train & 10k & 65.1 (1.7) & \bf 47.7 (1.5) & 5.6 & \bf 46.2 & 61.4 & 55.4 & \it 0.0  \\
\textbf{ON-LSTM} 3rd-layer & WSJ Train & 10k & 54.0 (3.9) & 36.6 (3.3) & 5.3 & 44.8 & 57.5 & 47.2 & \it 0.0  \\
\midrule 
CCM   &  WSJ10 Full & -- & 71.9 & -- & -- & -- & -- & -- & -- \\
DMV+CCM  & WSJ10 Full & -- & 77.6 &  -- & -- & -- & -- & -- & -- \\
UML-DOP &   WSJ10 Full & -- & \bf 82.9 & -- & -- & -- & -- & -- & --  \\
\midrule
Random Trees & --  & -- & 31.7 (0.3) & 18.4 (0.1) & 5.3 &17.4 & 22.3 & 16.0 & 40.4 \\
Balanced Trees & -- & -- & 43.4 (0.0) & 24.5 (0.0) & 4.6 & 22.1 & \textit{20.2} & \textit{9.3} & 55.9 \\
Left Branching & --  & -- & \it 19.6 (0.0) & \it 9.0 (0.0) & 12.4 & -- & -- & -- & --  \\
Right Branching &  -- & -- & 56.6 (0.0) & 39.8 (0.0) &  12.4 & -- & -- & -- & --  \\
\bottomrule 
\end{tabular}
}
\caption{ Unlabeled parsing F1 results evaluated on the full WSJ10 and WSJ test set. 
Our language model has three layers, each of them provides a sequence of $\hat{d}^f_t$. 
We provide the parsing performance for all layers.
Results with RL-SPINN and ST-Gumbel are evaluated on the full WSJ~\citep{williams2017broad}. 
PRPN models are evaluated on the WSJ test set~\citep{htut2018grammar}. We run the model with 5 different random seeds to calculate the average F1. The~\textit{Accuracy} columns represent the fraction of ground truth constituents of a given type that correspond to constituents in the model parses.
We use the model with the best F1 score to report ADJP, NP, PP, and INTJ.
WSJ10 baselines are from~\cite[CCM]{klein2002generative}, \cite[DMV+CCM]{klein2005natural}, and \cite[UML-DOP]{bod2006all}.
As the WSJ10 baselines are trained using POS tags, they are not strictly comparable with the latent tree learning results.
Italics mark results that are worse than the random baseline.
}
\label{tab:wsj-table}
\end{table} 

The performance is shown in Table \ref{tab:wsj-table}.
The second layer of ON-LSTM achieves state-of-the-art unsupervised constituency parsing results on the WSJ test set, while the first and third layers do not perform as well.
One possible interpretation is that the first and last layers may be too focused on capturing local information useful for the language modeling task as they are directly exposed to input tokens and output predictions respectively, thus may not be encouraged to learn the more abstract tree structure.
Since the WSJ test set contains sentences of various lengths which are unobserved during training, we find that ON-LSTM provides better generalization and robustness toward longer sentences than previous models.
We also see that the ON-LSTM model can provide strong results for phrase detection, including ADJP (adjective phrases), PP (prepositional phrases), and NP (noun phrases).
This feature could benefit many downstream tasks, like question answering, named entity recognition, co-reference resolution, etc.

\subsection{Targeted Syntactic Evaluation}
Targeted syntactic evaluation tasks have been proposed in~\cite{marvin2018targeted}. It is a collection of tasks that evaluate language models along three different structure-sensitive linguistic phenomena: subject-verb agreement, reflexive anaphora and negative polarity items.
Given a large number of minimally different pairs of English sentences, each consisting of a grammatical and an ungrammatical sentence, a language model should assign a higher probability to a grammatical sentence than an ungrammatical one.

Using the released codebase\footnote{\url{https://github.com/BeckyMarvin/LM_syneval}. We notice that the test set generated from the code is different from the one used in the original paper~\citep{marvin2018targeted}. Therefore, our results are not strictly comparable with the results in~\cite{marvin2018targeted}.} and the same settings proposed in~\cite{marvin2018targeted}, we train both our ON-LSTM model and a baseline LSTM language model on a 90 million word subset of Wikipedia.
Both language models have two layers of 650 units, a batch size of 128, a dropout rate of 0.2, a learning rate of 20.0, and were trained for 40 epochs.
The input embeddings have 200 dimensions and the output embeddings have 650 dimesions.

\begin{table}[h]
\centering
\scalebox{0.80}{
\begin{tabular}{l c c c c}
\toprule
& ON-LSTM & LSTM\\
\midrule
\large Short-Term Dependency \\
\midrule
\textsc{Subject-verb agreement:}\vspace{0.2em}\\
Simple                                          & 0.99 & \bf 1.00  \\
In a sentential complement                      & 0.95 & \bf 0.98  \\
Short VP coordination                           & 0.89 & \bf 0.92 \\
In an object relative clause                    & 0.84 & \bf 0.88  \\
In an object relative (no \textit{that})        & 0.78 & \bf 0.81  \\
\midrule
\textsc{Reflexive anaphora:}\vspace{0.2em}\\
Simple                                          & \bf 0.89 & 0.82  \\
In a sentential complement                      & \bf 0.86 & 0.80  \\
\midrule
\textsc{Negative polarity items:}\vspace{0.2em}\\
Simple (grammatical vs. intrusive)              & 0.18 & \bf 1.00  \\
Simple (intrusive vs. ungrammatical)            & \bf 0.50 & 0.01 \\
Simple (grammatical vs. ungrammatical)          & 0.07 & \bf 0.63\\
\midrule
\large Long-Term Dependency \\
\midrule
\textsc{Subject-verb agreement:}\vspace{0.2em}\\
Long VP coordination                            & \bf 0.74 & \bf 0.74  \\
Across a prepositional phrase                   & 0.67 & \bf 0.68  \\
Across a subject relative clause                & \bf 0.66 & 0.60  \\
Across an object relative clause                & \bf 0.57 & 0.52  \\
Across an object relative (no \textit{that})    & \bf 0.54 & 0.51  \\
\midrule
\textsc{Reflexive anaphora:}\vspace{0.2em}\\
Across a relative clause                        & 0.57 & \bf 0.58  \\
\midrule
\textsc{Negative polarity items:}\vspace{0.2em}\\
Across a relative clause (grammatical vs. intrusive)    & 0.59 & \bf 0.95  \\
Across a relative clause (intrusive vs. ungrammatical)    & \bf 0.20 & 0.00  \\
Across a relative clause (grammatical vs. ungrammatical)    & \bf 0.11 & 0.04  \\
\bottomrule
\end{tabular}
}
\caption{Overall accuracy for the ON-LSTM and LSTM on each test case. ``Long-term dependency'' means that an unrelated phrase (or a clause) exist between the targeted pair of words, while ``short-term dependency'' means there is no such distraction.}
\label{table:syneval}
\end{table}

Table \ref{table:syneval} shows that the ON-LSTM performs better on the long-term dependency cases, while the baseline LSTM fares better on the short-term ones.
This is possibly due to the relatively small number of units in the hidden states, which is insufficient to take into account both long and short-term information.
We also notice that the results for NPI test cases have unusually high variance across different hyper-parameters.
This result maybe due to the non-syntactic cues discussed in \cite{marvin2018targeted}.
Despite this, ON-LSTM actually achieves better perplexity on the validation set.

\section{Ordered Memory}\label{sec:ordered_memory}

While modern neural network models achieve surprising performance on many natural language tasks, our later experiment sections will show that they fall short on some formal language tasks.
Especially when standard neural network models face a systematic generalization setting, they tend suffer a dramatic performance drop.
This exposes the problem that these models fail to capture the internal mechanism of formal languages.
In this section, we present \textit{Ordered Memory} -- a new model that is developed from ordered neurons to solve formal language tasks.


\begin{figure}
    \centering
  \includegraphics[width=0.7\linewidth]{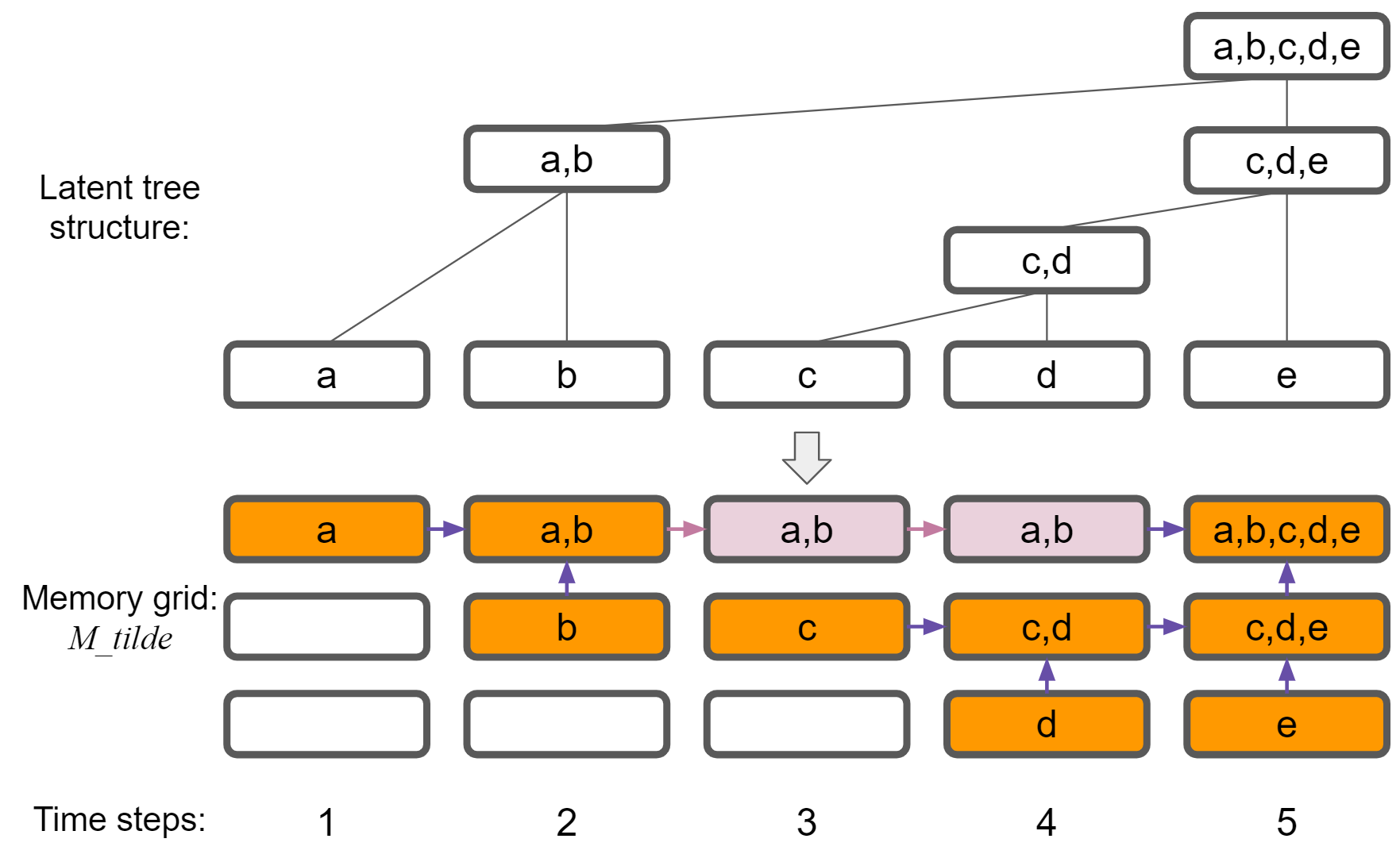}
  \caption{ The grid view of a tree structure. Blue arrows represent composing children into parent. Pink arrows represent copying from previous time-step. Orange slots are memories generated at the current time-step. Pink slots are memories copied from previous time-step.}
  \label{fig:grid_view}
\end{figure}

Ordered Memory is another instantiation of ordered neurons that groups neurons into memory slots and assigns a predetermined order to the memory slots.
The model explicitly models constituency structure through memory writing and erasing operations.
OM maps the latent syntax into a $T \times N$ memory grid $\tilde{M}$, where $T$ is the length of input sequence and $N$ is the maximum number of memory slots. 
Figure~\ref{fig:grid_view} gives an intuition of what the grid contains.
Empty blocks in the figure represent memory slots that can be discarded during inference.
Ideally, the memory network should generate the $t$-th column of the grid $\tilde{M}_t$ at time-step $t$.
But generating $\tilde{M}_t$ requires the model to have access about the tree structure which is usually latent. 
The OM model actively maintains its memory as a stack and processes the input from left to right, with a one-step lookahead in the sequence.
This allows the OM model to decide the local structure more accurately, much like a shift-reduce parser ~\citep{knuth1965translation}.

\begin{figure}[h]
  \centering
  \includegraphics[width=0.7\linewidth]{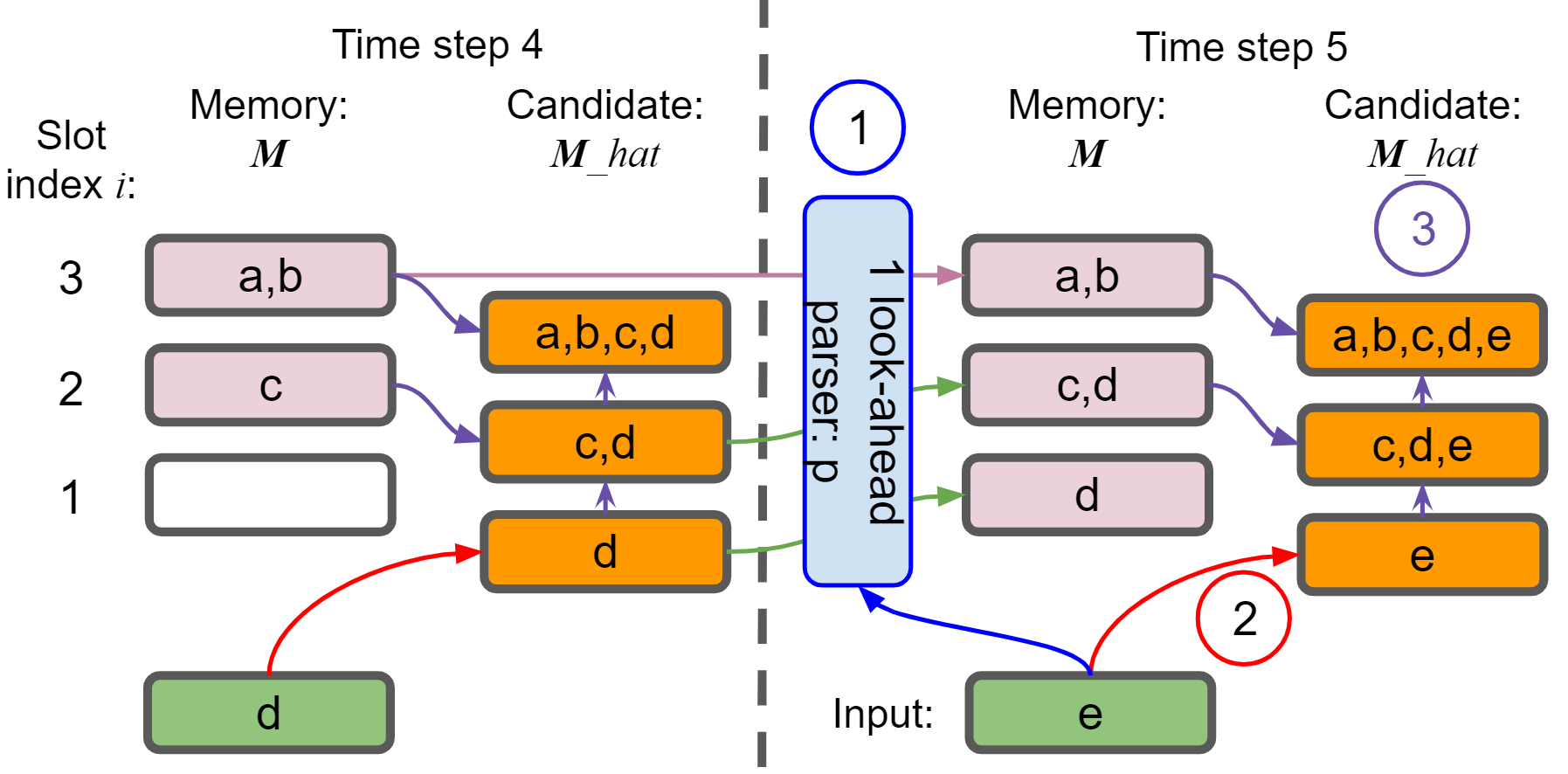}
  \caption{ The transition from time-step 4 to 5.
  (1) The one-step look-ahead parser combines $\hat{M}_{t-1}$ and $M_{t-1}$ considering on the current input $x_t$,
  in this example, the split point of $\hat{M}_{t-1}$ and $M_{t-1}$ is $i=2$.
  (2) Current input $x_t$ is written into the lower slot of new candidate memory $\hat{M}_{t}^{i-1}$. 
  (3) The rest of new candidate memories $\hat{M}_{t}^{\geq i}$ are generated with bottom-up recurrent composition.
  }
  \label{fig:om_timestep}
\end{figure}

At a given point $t$ in the input sequence $\boldsymbol{x}$ (the $t$-th time-step), we have a memory of candidate sub-trees spanning the non-overlapping sub-sequences in $x_1,\hdots, x_{t-1}$, with each sub-tree being represented by one slot in the memory stack.
We also maintain a memory stack of sub-trees that contains $x_1,\hdots, x_{t-2}$.
We use the input $x_t$ to choose its parent node from our previous candidate sub-trees.
The descendant sub-trees of this new sub-tree (if they exist) are removed from the memory stack, and this new sub-tree is then added.
We then build the new candidate sub-trees that include $x_t$ using the current input and the memory stack.
Figure \ref{fig:om_timestep} provides an example of one time-step in OM.
In what follows, we describe the OM model in detail.
To facilitate a clearer description, a discrete attention scheme is assumed, but only ``soft" attention is used in both the training and evaluation of this model.

Let $D$ be the dimension of each memory slot and $N$ be the number of memory slots. At time-step $t$, the model takes four inputs:

\begin{itemize}
    \item \textbf{Memory $M_{t-1}$}: a matrix of dimension $N \times D$, where each occupied slot is a distributed representation for a node spanning a subsequence in $x_1, .., x_{t-2}$ conditioned on $x_{t-1}$, i.e. $M_{t-1}$ represents a one-step look-ahead parser stack. 
    It's represented by gray blocks in Figure~\ref{fig:om_timestep}.

    \item \textbf{Candidate memory $\hat{M}_{t-1}$}: a matrix of dimension $N \times D$ contains representations for all possible new nodes at time-step $t-1$. 
    At next time-step $t$, the model will decide whether or not to write these candidates into memory $M_{t}$ conditioned on $x_{t}$.
    They are represented by orange blocks in Figure~\ref{fig:om_timestep}. 
    If the model is making correct parsing decisions, then $M_{t} = \tilde{M}_{t-1}$.
    
    \item \textbf{Memory mask $\cp_{t-1}$}: $\cp_{t-1} \in \{0, 1\}^N$, where each entry indicates whether the respective slot in $\hat{M}_{t-1}$ is occupied by a candidate,~e.g., if $\cp_{t-1} = (0, 1, 1)$, then the occupied slots are $\hat{M-1}_t^{\geq 2}$. At time-step t, the model can only choose a candidate from masked slots to write into the memory $M_{t}$.
    
    \item \textbf{Input} $x_t$: a vector of dimension $D_{in}$ that represent the input at time-step $t$.
\end{itemize}

The model first transforms $x_t$ to a $D$ dimension vector.
\begin{equation}
    \tilde{x}_t = LN(W x_t + b) \label{eq:input_proj}
\end{equation}
where $LN(\cdot)$ is the layer normalization function \citep{ba2016layer}.

To select the candidate representations from $\hat{M}_{t-1}$, the model uses $\tilde{x}_t$ as its query to attend on $\hat{M}_{t-1}$:
\begin{align}
    p_t &= \mathrm{Att}(\tilde{x}_t, \hat{M}_{t-1}, \  \cp_{t-1}) \\
    \cp^{i}_t &= \sum_{j\leq i}p^{j}_t \\
    \rcp^{i}_t &=  \sum_{j\geq i}p^{j}_t \label{eq:rcp}
\end{align}
where $\mathrm{Att}(\cdot)$ is a masked attention function, $\cp_{t-1}$ is the mask, $p_t$ is a distribution over different memory slots in $\hat{M}_{t-1}$, and $p_t^j$ is the probability on the $j$-th slot.
The attention mechanism will be described in section \ref{sec:stickbreaking}.
Intuitively, $p_t$ can be viewed as a pointer to the head of the stack, $\cp_t$ is an indicator value over where the stack exists, and $\rcp_t$ is an indicator over where the top of the stack is and where it is non-existent. 

To compute $M_{t}$, we combine $\hat{M}_{t-1}$ and $M_{t-1}$ through:
\begin{eqnarray}
    M_{t}^{i} = M_{t-1}^{i} \cdot (1 - \rcp)^{i} + \hat{M}_{t-1}^{i} \cdot \rcp_t^{i}, \quad \forall i \label{eq:write_m}
\end{eqnarray}

\begin{algorithm}[h]
    \centering
    
    \caption{
    Ordered Memory algorithm. The attention function $\mathrm{Att}(\cdot)$ is defined in section \ref{sec:stickbreaking}. The recursive cell function $\mathrm{cell}(\cdot)$ is defined in section \ref{sec:cell}.}
    \label{algo:OM}
    \begin{algorithmic}[1]
    \Function{OM}{$x_1, ..., x_T$}
    	\State initialize $M_{0}, \hat{M}_0$
        \For{$i\leftarrow 1$ to $T$}
            \State $\tilde{x}_t = LN(W x_t + b)$
            \State $p_t = \mathrm{Att}(\tilde{x}_t, \hat{M}_{t-1},   \cp_{t-1})$
            \State $\cp^{i}_t = \sum_{j\leq i}p^{j}_t$
            \State $\rcp^{i}_t =  \sum_{j\geq i}p^{j}_t$
            \State $\hat{M}^0_t = \tilde{x}_t$
            \For{$i\leftarrow 1$ to $N$}
                \State $M_{t}^{i} = M_{t-1}^{i} \cdot (1 - \rcp_t)^{i} +
                \hat{M}_{t-1}^{i} \cdot \rcp_t^{i}$\;
                \State $o_t^i = \cell(M_{t}^i, \hat{M}_{t}^{i-1})$
                \State $\hat{M}_t^i = \tilde{x}_t \cdot (1-  \cp_t)^{i} + o_t^i \cdot   \cp_t^{i}$
            \EndFor
        \EndFor
        \State \Return {$o_T^N$}
    \EndFunction
    \end{algorithmic}
\end{algorithm}

Suppose $p_t$ points at a memory slot $y_t$ in $\hat{m}$. 
Then $\rcp_t$ will write $\hat{M}_{t-1}^i$ to $M_{t}^i$ for $i \leq y_t$, and $(1 - \rcp_t)$ will write $M_{t-1}^i$ to $M_{t}^i$ for $i > y_t$.
In other words, Eqn. \ref{eq:write_m} copies everything from $M_{t-1}$ to the current timestep, up to where $p_t$ is pointing.

We believe that this is a crucial point that differentiates our model from past stack-augmented models like \cite{yogatama2016learning} and \cite{joulin2015inferring}.
Both constructions have the 0-th slot as the top of the stack, and perform a convex combination of each slot in the memory / stack given the action performed.
More concretely, a distribution over the actions that is not sharp (e.g. 0.5 for pop) will result in a weighted sum of an un-popped stack and a pop stack, resulting in a blurred memory state.
Compounded, this effect can make such models hard to train.
In our case, because $(1 - \rcp_t)^{i}$ is non-decreasing with $i$, its value will accumulate to 1 at or before $N$.
This results in a full copy, guaranteeing that the earlier states are retained.
This full retention of earlier states may play a part in the training process, as it is a strategy also used in \cite{gulcehre2017memory}, where all the memory slots are filled before any erasing or writing takes place.

To compute candidate memories for time-step $t$, we recurrently update all memory slots with
\begin{align}
    o^i &= \cell(M_{t}^i, \hat{M}_{t}^{i-1}) \label{eq:cell} \\
    \hat{M}_t^i &= \tilde{x}_t \cdot (1 - \cp_t)^{i+1} + o^i_t \cdot \cp_t^{i},  \forall i
    \label{eq:write_mhat}
\end{align}
where $\hat{M}^0_t$ is $x_t$.
The $\cell(\cdot)$ function can be seen as a recursive composition function in a recursive neural network \citep{socher2013recursive}. We propose a new cell function in section \ref{sec:cell}. 

The output of time-step $t$ is the last memory slot $\hat{M}_t^N$ of the new candidate memory, which summarizes all the information from $x_1, ..., x_t$ using the induced structure.
The pseudo-code for the OM algorithm is shown in Algorithm \ref{algo:OM}.

\subsection{Masked Attention} \label{sec:stickbreaking}
Given the projected input $\tilde{x}_t$ and candidate memory $\hat{M}_{t-1}^{i}$.
We feed every $( \tilde{x}_t, \hat{M}_{t-1}^{i} )$ pair into a feed-forward network:
\begin{eqnarray}
    \alpha_t^i &=& \frac{ \mathbf{w}^{Att}_2 ~ \mathrm{tanh} \left( \mathbf{W}^{Att}_1  
    \left[ 
    \begin{matrix}
        \hat{M}_{t-1}^i \\ 
        \tilde{x}_t
    \end{matrix} 
    \right] 
    + b_1 \right)+ b_2 }{\sqrt{N}} \\
    \beta_t^i &=& \exp \left( \alpha_t^i - \max_j \alpha_t^j \right)
\end{eqnarray}
where $\mathbf{W}^{Att}_1$ is $N \times 2N$ matrix, $\mathbf{w}^{Att}_2$ is $N$ dimension vector, and the output $\beta^i_t$ is a scalar. 
The purpose of dividing by $\sqrt{N}$ is to scale down the logits before softmax is applied, a technique similar to the one seen in \cite{vaswani2017attention}.
We further mask the $\beta_t$ with the cumulative probability from the previous time-step to prevent the model attending on non-existent parts of the stack:
\begin{eqnarray}
    \hat{\beta}_t^i = \beta_t^i   \cp_{t-1}^{i+1} \label{eq:cum_prod}
\end{eqnarray}
where $  \cp_{t-1}^{N+1} = 1$ and $  \cp_0^{\leq N}=0$.
We can then compute the probability distribution:
\begin{eqnarray}
    p_t^i = \frac{\hat{\beta}_{t}^i}{\sum_j \hat{\beta}_{t}^j} 
    \label{eq:stickbreaking}
\end{eqnarray}
This formulation bears similarity to the method used for the multi-pop mechanism seen in \cite{yogatama2018memory}.

\subsection{Gated Recursive Cell} \label{sec:cell}
Instead of using the recursive cell proposed in TreeLSTM \citep{tai2015improved} and RNTN \citep{socher2010learning}, we propose a new gated recursive cell, which is inspired by the feed-forward layer in Transformer \citep{vaswani2017attention}. 
The inputs $M^{i}_{t}$ and $\hat{M}_t^{i-1}$ are concatenated and fed into a fully connect feed-forward network:
\begin{equation}
    \left[
    \begin{matrix}
        \vg_t^i \\ 
        \hg_t^i \\
        \cg_t^i \\
        u_t^i
    \end{matrix}
    \right]=  \mathbf{W}^{Cell}_2~ \mathrm{ReLU} \left( \mathbf{W}^{Cell}_1  \left[ 
    \begin{matrix}
        \hat{M}_t^{i-1} \\ 
        M^i_{t} 
    \end{matrix} 
    \right] 
    + b_1 \right) + b_2
\end{equation}
Like the TreeLSTM, we compute the output with a gated combination of the inputs and $u_t^i$:
\begin{eqnarray}
    o_t^i &=& LN (\sigma(\vg^i_t) \odot \hat{M}^{i-1}_t
                    + \sigma(\hg^i_t) \odot M^{i}_{t} 
                    + \sigma(\cg^i_t) \odot u^i_t ) \label{eq:gated_sum}
\end{eqnarray}
where $\vg_t^i$ is the vertical gate that controls the input from previous slot, $\hg_t^i$ is horizontal gate that controls the input from previous time-step, $cg_t^i$ is cell gate that control the $u^i_t$, $o^i_t$ is the output of cell function, and $LN(\cdot)$ share the same parameters with the one used in the Eqn. \ref{eq:input_proj}.

\subsection{Relations to ON-LSTM and Shift-reduce Parser}\label{sec:relationship}
Ordered Memory is implemented following the principles introduced in Ordered Neurons.
Our model is related to ON-LSTM in several aspects:
1) The memory slots are similar to the chunks in ON-LSTM, when a higher ranking memory slot is forgotten/updated, all lower ranking memory slots should likewise be forgotten/updated;
2) ON-LSTM uses the monotonically non-decreasing master forget gate to preserve long-term information while erasing short term information, the OM model uses the cumulative probability $\cp_t$;
3) Similarly, the master input gate used by ON-LSTM to control the writing of new information into the memory is replaced with the reversed cumulative probability $\rcp_t$ in the OM model.

At the same time, the internal mechanism of OM can be seen as a continuous version of a shift-reduce parser. 
At time-step $t$, a shift-reduce parser could perform zero or several reduce steps to combine the heads of stack, then shift the word $t$ into stack.
The OM implement the reduce step with Gated Recursive Cell. 
It combines $\hat{M}_t^{i-1}$, the output of previous reduce step, and $M_t^i$, the next element in stack, into $\hat{M}_t^{i}$, the representation for new sub-tree.
The number of reduce steps is modeled with the attention mechanism. 
The probability distribution $p_t$ models the position of the head of the stack after all necessary reduce operations are performed.
The shift operations is implemented as copying the current input word $x_t$ into memory.

\begin{algorithm}[h]
    \centering
    
    \caption{Shift-reduce parsing algorithm for generate parsing tree from Ordered Memory model. Here we greedily choose the $\mathrm{argmax}(p_t)$ as the head of stack for each slot.}
    \label{algo:parsing}
    \begin{algorithmic}[1]
    \Function{Constituent}{$(w_1, p_1), ..., (w_T, p_T)$}
        \State $\mathrm{queue}=[w_2,...,w_T]$
        \State $\mathrm{stack}=[w_1]$
        \State $h=\mathrm{argmax}(p_1)-1$
        \For{$i\leftarrow 2$ to $T$}
            \State $y_i=\mathrm{argmax}(p_i)$
            \State $d = y_i - h$
            \If{$d > 0$}
                \For{$j\leftarrow 1$ to $d$}
                    \If{$\mathrm{len}(\mathrm{stack}) < 2$}
                        \State \textbf{Break}
                    \EndIf
                    \State $e_1=\mathrm{stack}.\mathrm{pop}()$
                    \State $e_2=\mathrm{stack}.\mathrm{pop}()$ 
                    \State $\mathrm{stack}.\mathrm{push}(\mathrm{node}(e_1,e_2))$
                \EndFor
            \EndIf
            \State $\mathrm{stack}.\mathrm{push}(\mathrm{queue}.\mathrm{popleft}())$
            \State $h=y_i-1$
        \EndFor
        \While{$\mathrm{len(stack)}>2$}
            \State $e_1=\mathrm{stack}.\mathrm{pop}()$
            \State $e_2=\mathrm{stack}.\mathrm{pop}()$
            \State $\mathrm{stack}.\mathrm{push}(\mathrm{node}(e_1,e_2))$
        \EndWhile
        \State \Return $\mathbf{T}$
    \EndFunction
    \end{algorithmic}
\end{algorithm}

The upshot of drawing connections between our model and the shift-reduce parser is interpretability: We can approximately infer the computation graph constructed by our model with Algorithm \ref{algo:parsing}.
The algorithm can be used for the latent tree induction tasks in \cite{williams2018latent}.

\section{Formal Language Experiments}\label{sec:fle}

Formal languages are artificial languages, constructed from a well-defined grammar.
Many formal language tasks requires the model to have a comprehensive understanding about the grammar.
It's also straightforward to produce special training and test set, so that the model need to transfer the learned syntax and operators to out-of-distribution data points.
In this section, we will focus on test OM performance on two tasks: Logical Inforence~\citep{bowman2014recursive} and ListOps~\citep{nangia2018listops}.
Experiment results show that OM outperformance all baseline models, including ON-LSTM.
On out-of-distribution test sets, OM also demonstrates a symbolic-like systematical generalization ability.

\subsection{Logical Inference}
\begin{table}[h]
\small
    \caption{
    Partitions of the Logical Inference task from \cite{bowman2014recursive}. Each partitions include a training set filtered out all data points that match the rule indicated in \textbf{Excluded}, and a test set formed by matched data points.
    }
    \centering
    \begin{tabular}{r l r l}
    \toprule
    \textbf{Part.} & \textbf{Excluded} & 
    \textbf{Training set size}  &
    \textbf{Test set example} \\
    \midrule
   
    A & \texttt{* ( and (not a) ) *} & 
        128,969 &
        \texttt{f (and (not a))} \\
    B & \texttt{* ( and (not *) ) *} &
        87,948  &
        \texttt{c (and (not (a (or b))))}  \\
    C & \texttt{* ( \{and,or\} (not *) ) *} &
          51,896 &
        \texttt{a (or (e (and c)))}\\
    \midrule
    Full &  & 135,529 &  \\
    \bottomrule
    \end{tabular}

    \label{tab:ablation_tests}
\end{table}

\begin{figure}[h]
\small
\begin{center}
    \begin{forest}
		shape=coordinate,
		where n children=0{
			tier=word
		}{},
		nice empty nodes
[[[a], [[or], [[not], [d]]]], [[and], [[not], [[[not], [b]], [[and], [c]]]]]]
	\end{forest}
	\quad
	\begin{forest}
		shape=coordinate,
		where n children=0{
			tier=word
		}{},
		nice empty nodes
[[[[a], [[or], [[not], [d]]]], [and]], [[not], [[[[not], [b]], [and]], [c]]]]
	\end{forest}
	\quad 
	\begin{forest}
		shape=coordinate,
		where n children=0{
			tier=word
		}{},
		nice empty nodes
 [[[a], [[or], [[not], [d]]]], [[[and], [not]], [[[[not], [b]], [and]], [c]]]]
	\end{forest}
\end{center}

\caption{
Variations in induced parse trees under different runs of the logical inference experiment. 
The left most tree is the ground truth and one of induced structures.
We have removed the parentheses in the original sequence for this visualization.
It is interesting to note that the different structures induced by our model are all valid computation graphs to produce the correct results.
}
\label{fig:trees}
\end{figure}
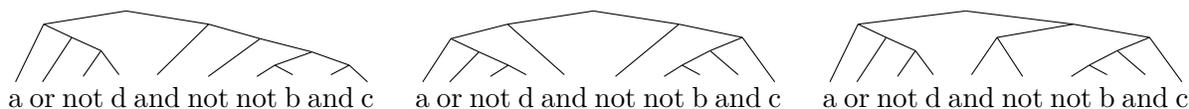



The logical inference task described in~\cite{bowman2015tree} has a vocabulary of six words and three logical operations, \texttt{or}, \texttt{and}, \texttt{not}. 
The task is to classify the relationship of two logical clauses into seven mutually exclusive categories.
We use a multi-layer perceptron (MLP) with $(h_1, h_2, h_1 \circ h_2, |h_1 - h_2|)$ as input, where $h_1$ and $h_2$ are the $\hat{M}^N_T$ of their respective input sequences.
We compare our model with LSTM, RRNet \citep{jacob2018learning}, Tranformer \citep{vaswani2017attention}, Universal Transformer \citep{dehghani2018universal}, TreeLSTM \citep{tai2015improved}, TreeRNN \citep{bowman2015tree}, and TreeCell.
We used the same hidden state size for our model and baselines for proper comparison. 
The model is trained on sequences containing up to $6$ operations and tested on sequences with higher number (7-12) of operations.

The Transformer models were implemented by modifying the code from the Annotated Transformer\footnote{\url{http://nlp.seas.harvard.edu/2018/04/03/attention.html}}.
The number of Transformer layers are the same as the number of slots in OM.
Unfortunately, we were not able to successfully train a Transformer model on the task, resulting in a model that only learns the marginal over the labels.
\cite{tran2018importance} achieves similar results, suggesting this could be a problem intrinsic to self-attention mechanisms for this task.

\begin{table}[h]
\small
    \centering
    \setlength{\tabcolsep}{3.5pt} 
    \caption{
    Test accuracy of the models, trained on operation lengths of $\leq 6$, with their out-of-distribution results shown here (lengths 7-12).
    We ran 5 different runs of our models, giving the error bounds in the last row.
    The $F_1$ score is the parsing score with respect to the ground truth tree structure. 
    The TreeCell is a recursive neural network based on the Gated Recursive Cell function proposed in section \ref{sec:cell}.
    For the Transformer and Universal Transformer, we follow the entailment architecture introduced in \cite{radford2018improving}. 
    The model takes \texttt{<start> sentence1 <delim> sentence2 <extract>} as input, then use the vector representation for \texttt{<extract>} position at last layer for classification.
    $^*$The results for RRNet were taken from \cite{jacob2018learning}.
    }
    \begin{tabular}{l c cccccc  c ccc}
    \toprule
    \textbf{Model} &&  \multicolumn{6}{c}{\textbf{Number of Operations}} 
          &&  \multicolumn{3}{c}{\textbf{Sys. Gen.}} \\
          
          && 7 & 8 & 9 & 10 & 11 & 12
          && A & B & C \\
    \midrule
    \multicolumn{12}{l}{\emph{Sequential sentence representation}}\\
    LSTM     && 88 & 84 & 80 & 78 & 71 & 69 && 84 & 60 & 59 \\
    RRNet*    && 84 & 81 & 78 & 74 & 72 & 71 && -- & -- & -- \\
    \cmidrule{1-12}
    \multicolumn{12}{l}{\emph{Inter sentence attention}}\\
    Transformer && 51 & 52 & 51 & 51 & 51 & 48 && 53	 & 51	& 51\\
    Universal Transformer && 51 & 52 & 51 & 51 & 51 & 48 && 53	 & 51	& 51 \\
    \cmidrule{1-12}
    \multicolumn{12}{l}{\emph{Our models}}\\
    ON-LSTM  && 91 & 87 & 85 & 81 & 78 & 75 && 70 &	63 & 60 \\
    OM && 98 $\pm$ 0.0 & 97 $\pm$ 0.4 & 96 $\pm$ 0.5 & 94 $\pm$ 0.8 & 93 $\pm$ 0.5 & 92 $\pm$ 1.1
             && 94 &	91 &	81\\
    \cmidrule{1-12}
    \multicolumn{12}{l}{\emph{Recursive NN + ground-truth structure}}\\
    TreeLSTM && 94 & 92 & 92 & 88 & 87 & 86 && 91 & 84 & 76 \\
    TreeCell && 98 & 96 & 96 & 95 & 93 & 92 && 95 & 95 & 90 \\
    TreeRNN  && 98 & 98 & 97 & 96 & 95 & 96 && 94 &	92 & 86 \\
    \bottomrule
    \end{tabular}
    \label{tab:proplogparse}
\end{table}

\paragraph{\textbf{Length Generalization Tests}}
The TreeRNN model represents the best results achievable if the structure of the tree is known.
The TreeCell experiment was performed as a control to isolate the performance of using the $\cell(\cdot)$ composition function versus using both using $\cell(\cdot)$ and learning the composition with OM.
The performance of our model degrades only marginally with increasing number of operations in the test set, suggesting generalization on these longer sequences never seen during training.

\paragraph{\textbf{Parsing results}}
OM achieve an unsupervised parsing performance $84.3 \pm 14.4$.
There is a variability in parsing performance over several runs under different random seeds, but the model's ability to generalize to longer sequences remains fairly constant.
The model learns a slightly different method of composition for consecutive operations.
Perhaps predictably, these are variations that do not affect the logical composition of the subtrees.
The source of different parsing results can be seen in Figure \ref{fig:trees}.
The results suggest that these latent structures are still valid computation graphs for the task, in spite of the variations.

\paragraph{\textbf{Systematic Generalization Tests}}
Inspired by \cite{loula2018rearranging}, we created three splits of the original logical inference dataset with increasing levels of difficulty.
Each consecutive split removes a superset of the previously excluded clauses, creating a harder generalization task.
Each model is then trained on the ablated training set, and tested on examples unseen in the training data.
As a result, the different test splits have different numbers of data points.
Table \ref{tab:ablation_tests} contains the details of the individual partitions.

The results are shown in the right section of Table \ref{tab:proplogparse} under Sys. Gen.
Each column labeled A, B, and C are the model's aggregated accuracies over the unseen operation lengths.
As with the length generalization tests, the models with the known tree structure performs the best on unseen structures, while sequential models degrade quickly as the tests get harder.
Our model greatly outperforms all the other sequential models, performing slightly below the results of TreeRNN and TreeCell on different partitions.

Combined with the parsing results, and our model's performance on these generalization tests, we believe this is strong evidence that the model has both (i) learned to exploit symmetries in the structure of the data by learning a good $\cell(\cdot)$ function, and (ii) learned where and how to apply said function by operating its stack memory.

\subsection{ListOps}
\begin{figure}[h]
    \centering
    \begin{subfigure}[b]{0.45\textwidth}
        \small
        \begin{tabular}{l cc }
            \toprule
            \textbf{Model} & \textbf{Accuracy}  & $F_1$\\
            \midrule
            \multicolumn{3}{l}{\emph{Baselines}} \\
            LSTM*                   & 71.5\(\pm\)1.5  & --\\
            RL-SPINN*               & 60.7\(\pm\)2.6  & 71.1 \\
            Gumbel Tree-LSTM*       & 57.6\(\pm\)2.9  & 57.3 \\
            Transformer             & 57.4\(\pm\)0.4  & -- \\
            Universal Transformer   & 71.5\(\pm\)7.8  & -- \\
            \cite{havrylov2019cooperative}  & 99.2\(\pm\)0.5 & --\\
            \midrule
            OM & \bf 99.97\(\pm\)0.014 & \bf 100 \\
            \bottomrule
        \end{tabular}
        \caption{}
        \label{table:listops comparison}
    \end{subfigure}
    \quad
    \begin{subfigure}[b]{0.45\textwidth}
        \includegraphics[width=1\textwidth]{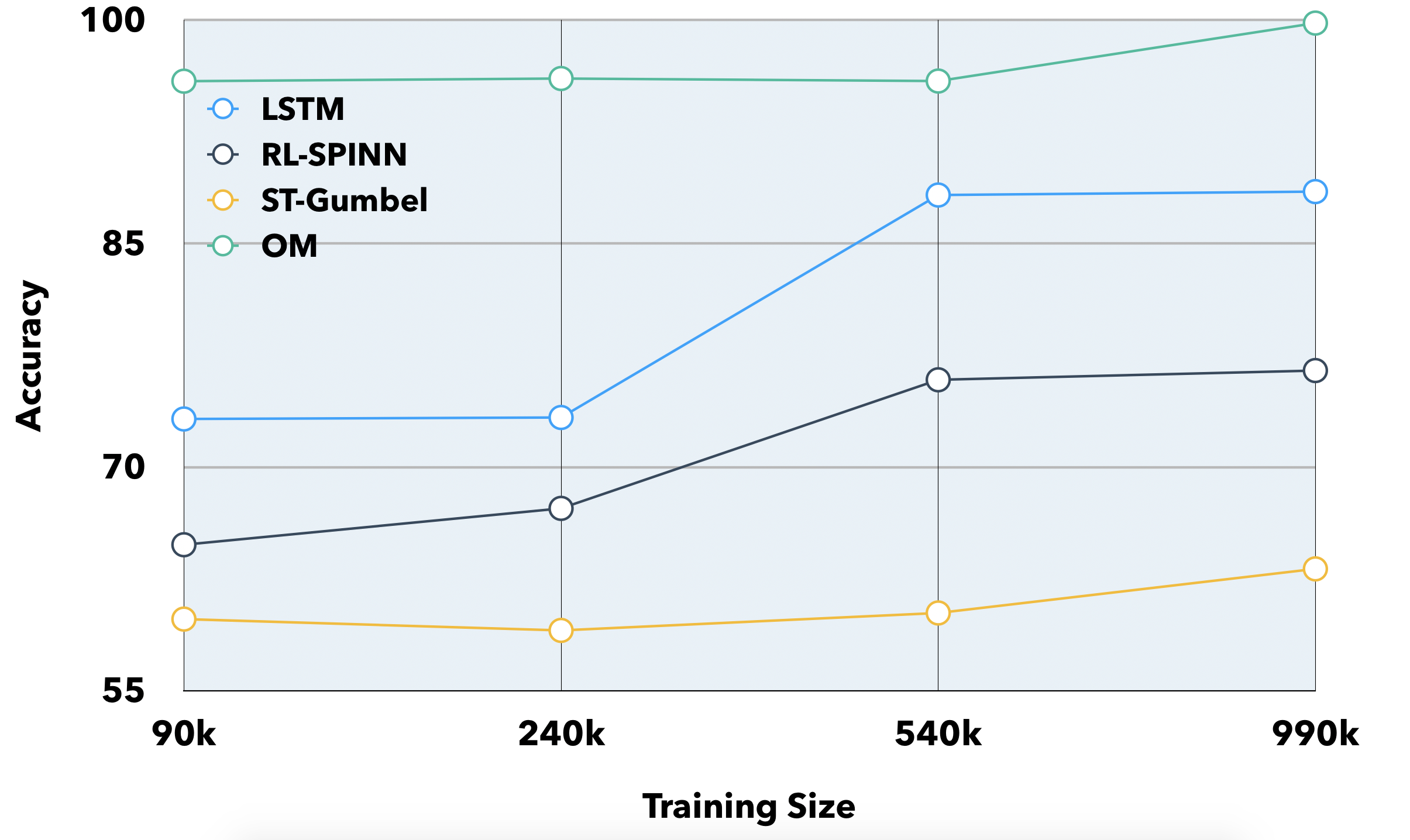}
        \caption{}
        \label{fig:listops_gen}
    \end{subfigure}
    \caption{
    (a) shows the accuracy of different models on the ListOps dataset. All models have $128$ dimensions. Results for models with * are taken from \cite{nangia2018listops}.
    (b) shows our model accuracy on the ListOps task when varying the the size of the training set.
    }
\end{figure}

\cite{nangia2018listops} build a dataset with nested summary operations on lists of single digit integers.
The sequences comprise of the operators \texttt{MAX}, \texttt{MIN}, \texttt{MED}, and \texttt{SUM\_MOD}.
The output is also an integer in $[0, 9]$
As an example, the input: \texttt{[MAX 2 9 [MIN 4 7 ] 0 ]} has the solution \texttt{9}.
As the task is formulated to be easily solved with a correct parsing strategy, the task provides an excellent test-bed to diagnose models that perform tree induction.
The authors binarize the structure by choosing the subtree corresponding to each list to be left-branching: the model would first take into account the operator, and then proceed to compute the summary statistic within the list. A right-branching parse would require the entire list to be maintained in the model's hidden state. 

OM achieves 99.9\% accuracy, and an $F_1$ score of 100\% on the model's induced parse tree~(See Table \ref{table:listops comparison}).
This result is consistent across 3 different runs of the same experiment. In \cite{nangia2018listops}, the authors perform an experiment to verify the effect of training set size on the latent tree models.
As the latent tree models (RL-SPINN and ST-Gumbel) need to parse the input successfully to perform well on the task, the better performance of the LSTM than those models indicate that the size of the dataset does not affect the ability to learn to parse much for those models.
Our model seems to be more data efficient and solves the task even when only training on a subset of 90k examples (Fig.~\ref{fig:listops_gen}).

\section{Recent Advances}\label{constituency_conclusion}
After ON-LSTM was first presented in 2018, many new developments on constituency inductive bias and unsupervised constituency parsing have been introduced.
In URNNG \citep{kim2019unsupervised}, amortized variational inference was applied between a recurrent neural network grammar (RNNG) \citep{dyer2016recurrent} decoder and a tree structure inference network, which encourages the decoder to generate reasonable tree structures. 
DIORA \citep{drozdov2019unsupervised} proposed using inside-outside dynamic programming to compose latent representations from all possible binary trees. 
The representations of inside and outside passes from the same sentences are optimized to be close to each other.
The compound PCFG \citep{kim2019compound} achieves grammar induction by maximizing the marginal likelihood of the sentences which are generated by a probabilistic context-free grammar (PCFG).
Neural L-PCFG \citep{zhu2020return} demonstrated that PCFG can benefit from modeling lexical dependencies. 
The Neural L-PCFG induces both constituents and dependencies within a single model.
Tree Transformer \citep{wang2019tree} share a similar inductive bias as ordered neurons.  
It adds extra locality constraints to the Transformer encoder's self-attention to encourage the attention heads to follow a tree structure such that each token can only attend on nearby neighbors in lower layers and gradually extend the attention field to further tokens when climbing to higher layers.
\chapter{Dependency Inductive Bias: Unsupervised Dependency Graph Network}
\label{cha:dependency}

In this chapter, we focus on the other facet of syntax: Dependency grammar.
Dependency is the notion that linguistic units, e.g. words, are connected to each other by directed links.
The natural of dependency graph makes it compatible with modern graph neural networks.
However the actually graph structure is latent, which means the ground-truth structure is usually unavailable to model.
Hence, we propose a dependency inductive bias to induce the structure, and a dependency graph network to model the information propagation between words.

We start from reviewing related works which either augment neural networks with dependency graph or induce the dependency graph from raw data (Section~\ref{sec:dependency_prev}).
In Section~\ref{sec:udgn}, we introduce Unsupervised Dependency Graph Network (UDGN) that includes a Dependency Graph Network and a parser.
The Dependency Graph Network (DGN) uses different channels and a competitive mechanism to model information propagation on different types of dependency edges.
Alongside the DGN, an independent parsing module provides a soft and differentiable dependency mask to constrain the information propagation.
In Section~\ref{sec:dependency_exp}, we train UDGN with a Masked Language Model (MLM) objective.
We find that the model learns to induce dependency trees.
It achieves strong performance on MLM and state-of-the-art results in unsupervised dependency parsing on the Wall Street Journal treebank.
We also perform analysis on the mechanisms that give rise to this behavior and evaluate the model's potential for finetuning on a downstream task.

\section{Previous Approaches}\label{sec:dependency_prev}

\subsection{Dependency-Augmented Models}
In many Transformer-based models, attention masks are often used to limit the input tokens that a particular timestep can attend over.
This attention mask can be viewed as an adjacency matrix over a graph whose nodes are the input tokens.
From this perspective, Transformers are a form of Graph Convolution network (GCN; \citep{kipf2016semi}) --- specifically, a Graph Attention Network (GAT; \citep{velivckovic2017graph}), as it attends over the features of its neighbors.
Several works have made this connection, and integrated dependency structures into transformers \citep{ahmad2020gate, wang2019tree,tang2020dependency}.
Results from \cite{omote2019dependency} and \cite{deguchi2019dependency} suggest that embedding these structures can improve translation models.

However, these dependency parses may not always be present to be used as input to the model.
\cite{strubell2018linguistically} trains the self-attention to attend the syntactic governor (head) of a particular token, resulting in a model that does not require dependency structure as input during inference time. 
We take a further step in our work and attempt to learn these structures in an unsupervised fashion from the MLM objective.


\subsection{Unsupervised Dependency Parsing}
Previous works on unsupervised dependency parsing are primarily based on the dependency model with valence (DMV) \citep{klein2004corpus} and its extension \citep{daume2009unsupervised, gillenwater2010sparsity}. 
To effectively learn the DMV model for better parsing accuracy, a variety of inductive biases and handcrafted features, such as correlations between parameters of grammar rules involving different part-of-speech (POS) tags, have been proposed to incorporate prior information into learning.
The most recent progress is the neural DMV model \citep{jiang2016unsupervised}, which uses a neural network model to predict the grammar rule probabilities based on the distributed representation of POS tags.
However, most existing unsupervised dependency parsing algorithms require the gold POS tags to ge provided as inputs. These gold POS tags are labeled by humans and can be potentially difficult (or prohibitively expensive) to obtain for large corpora.
\cite{spitkovsky2011unsupervised} proposed to overcome this problem with unsupervised word clustering that can dynamically assign tags to each word considering its context.
\cite{he2018unsupervised} overcame the problem by combining DMV model with invertible neural network to jointly model discrete syntactic structure and continuous word representations.

Dependency Model with Valence (DMV; \citep{klein2004corpus}) is the basis of several unsupervised dependency parsing methods \citep{daume2009unsupervised,gillenwater2010sparsity}.
\cite{jiang2016unsupervised} updates the method using neural networks to predict grammar rule probabilities.
These methods require additional Part-of-Speech (POS) information.
\cite{spitkovsky2011unsupervised} tackled the issue by performing clustering to assign tags to each word by considering its context.
In our proposal, the parser predicts pseudo-probabilities for POS tags for each word.
These probabilities are then used for predicting the head word.

\section{Unsupervised Dependency Graph Network}\label{sec:udgn}

\begin{figure}[t]
    \centering
    \includegraphics[width=0.5\textwidth]{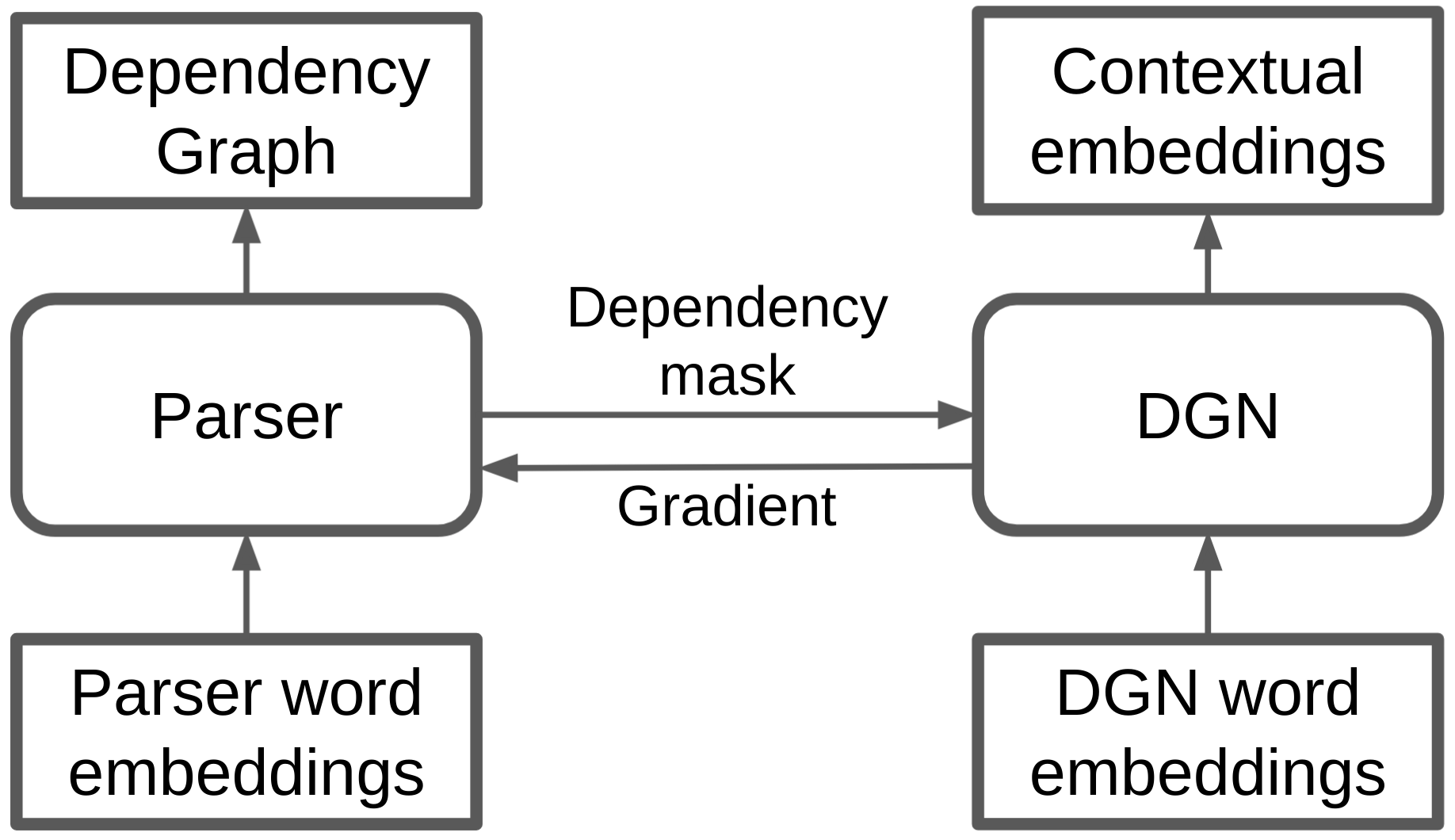}
    \caption{The architecture of Unsupervised Dependency Graph Network (UDGN).
    The model includes a parser and a Dependency Graph Network (DGN).
    Given an input sentence, the parser can predict the dependency relation between tokens and generate a soft mask to approximate the undirected dependency graph.
    The DGN takes the sentence and mask as input, and output contextual word embeddings.
    Since the mask is soft, the gradient can be backpropagated from the DGN into the parser.
    Thus UDGN can induce grammar while training on downstream tasks.
    }
    \label{fig:udgn_framework}
\end{figure}

In this section, we propose the Unsupervised Dependency Graph Networks (UDGN).
The UDGN includes two components:
\begin{itemize}
    \item A parser, which computes the dependency head probability distribution $p_i$ for each word $w_i$ in the input sentence, and then converts it to a matrix of edge probability $\mask_{ij}$ that approximates an undirected dependency graph;
    \item A multilayer Dependency Graph Network (DGN) that propagates information between words to compute a contextualized embedding $h_i$ for each word $w_i$. 
    It use $\mask_{ij}$ to control information propagation between word pair $(w_i, w_j)$.
\end{itemize}

As shown in Figure \ref{fig:udgn_framework}, the parser computes a dependency head distribution for each token and then converts it to a soft dependency mask $m_{ij}$. The DGN takes $m_{ij}$ and the sentence as input and uses a competitive mechanism to propagate information between tokens.

While training with the masked language modeling objective, the gradient can flow through the DGN to the parser network through its dependence on $\mask_{ij}$.
As a result, UDGN can induce a dependency grammar while solely relying on the masked language modeling objective.

To better model and induce dependency relations, we propose three key components for the DGN: 
\begin{itemize}
    \item A competitive mechanism is proposed to control information propagation between words. 
    The competitive mechanism controls several channels.
    A channel is a function that models a specific type of information propagation.
    Given a word pair $(w_i,w_j)$, channels will compete to get more weight to propagate information from word $w_j$ to $w_i$.
    The mechanism is inspired by syntactic functions in dependency graphs.
    \item A gated non-linear network is designed to parameterize each channel.
    The gating mechanism allows each word to filter information extracted from other tokens.
    And the non-linear function can increase the capacity of the channel.
    \item Relative position biases are proposed to model the sequential order of words.
    The bias allows some channels to focus on extracting information from positions before the current position, while other channels focus on future positions.
\end{itemize}

\begin{figure*}[h]
    \centering
    \includegraphics[width=1\textwidth]{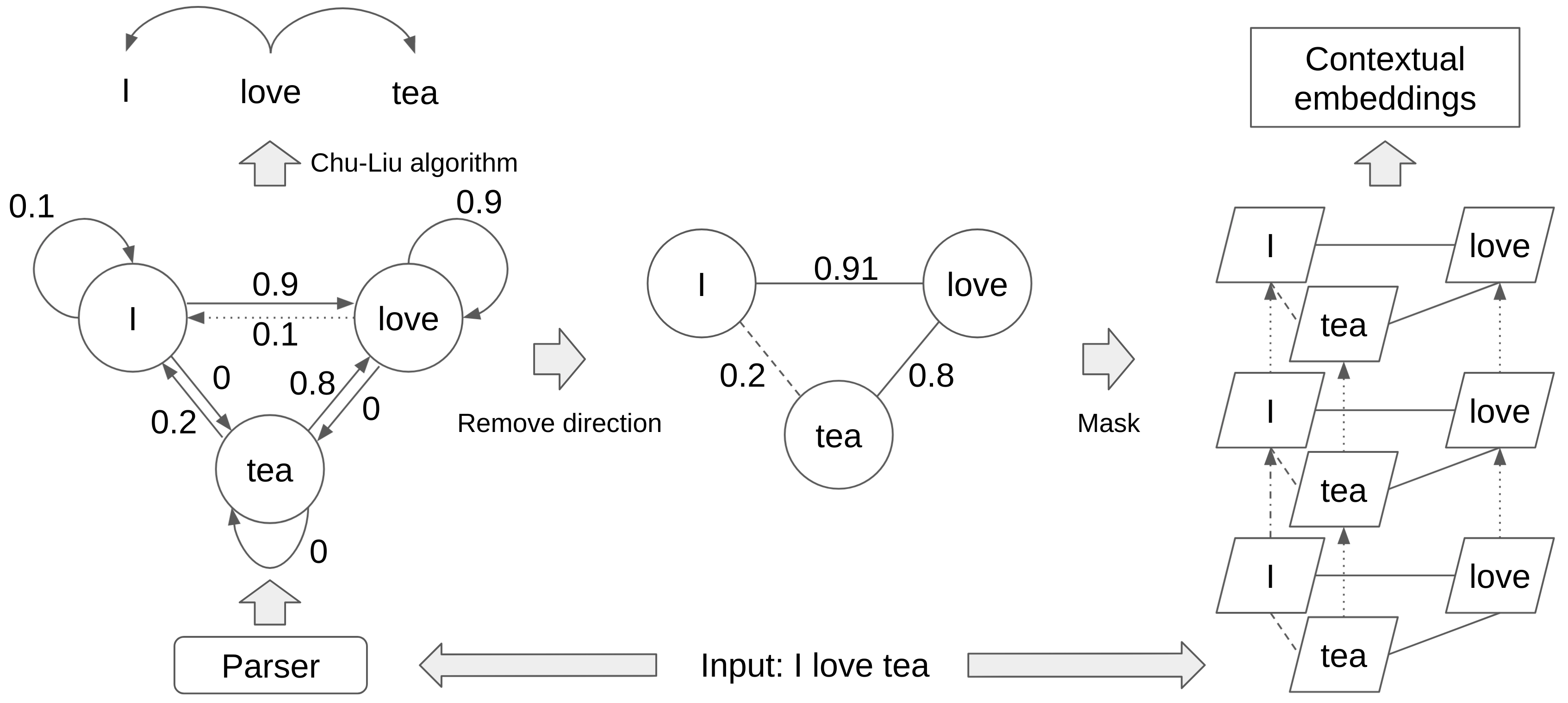}
    \caption{Details of the UDGN.
    Given the input sentence, the parser (left) produces a dependency head distribution for each token.
    These distributions form a distribution matrix $p_{ij}$.
    During inference, the Chu-Liu algorithm generates the most likely dependency graph given $p_{ij}$.
    While training, however, we remove the direction of dependency in $p_{ij}$ and obtain an undirected dependency mask $m_{ij}$ (middle).
    $m_{ij}$ is symmetric and with zeroes along the diagonal.
    The DGN (right) takes $m_{ij}$ and the sentence as input and uses a competitive mechanism to propagate information between tokens.
    Inside each layer of the DGN, every node (token) will extract information from all other nodes.
    $m_{ij}$ controls the amount of information being propagated between nodes. 
    If $m_{ij}$ is small then less information will be communicated between $x_i$ and $x_j$, and vice versa.
    After several layers, DGN outputs the contextual embedding for each token.
    These embeddings can be used either to predict missing tokens or as features for downstream tasks.}
    \label{fig:udgn_detail}
\end{figure*}

\subsection{Head Selective Parser}
We use a simplified version of the dependency neural selection parser \citep{zhang2016dependency} that only predicts unlabelled dependency relations. 
The parser takes the sentence $\mathbf{s}=w_1 w_2 ... w_T$ as input, and, for each token $w_i$, it produces a distribution $p_i$ over all tokens in the sentence, resulting in a $T \times T$ weight matrix.

The parser first maps the sequence of tokens $w_1 w_2 ... w_T$ into a sequence of embeddings $[\x_1, \x_2, ..., \x_T]$. 
The standard word embedding method usually maps the input token $w_i$ to a unique vector $\e_{w_i}$.
Under this setting, we found that the unsupervised parsing performance becomes increasingly unstable with increasing vocabulary size.
We hypothesize that the randomly initialized embedding for low-frequency tokens confuse the parser and cause an extra optimization problem.
On the other hand, low-frequency tokens usually share similar grammatical roles.
So we propose an extra embedding component $\e^{\mathrm{tag}}_{k}$ that is shared across tokens with similar grammatical roles:
\begin{align}
    \x_i &= \e_{w_i} + \sum_k p^{\mathrm{tag}}_{w_i k} \e^{\mathrm{tag}}_{k} \\
    p^{\mathrm{tag}}_{w_i k} &= \mathrm{softmax}(\tau_{w_i k})
\end{align}
where $e_{w_i}$ is the original word embedding, $p^{\mathrm{tag}}_{w_i k}$ is a probability distribution that associate the $w_i$ with different $\e^{\mathrm{tag}}_{k}$.

Then the word embeddings are fed into a stack of a Bidirectional LSTM:
\begin{equation}
    \h_i = \mathrm{BiLSTM}(\x_i)
\end{equation}
$\h_i$ is the output of the BiLSTM at $i$-th timestep. 
Linear transforms are applied to the output of the BiLSTM to extract head and dependent information.
\begin{align}
    \head_i &= \weight_{\mathrm{H}} \h_i + \bias_{\mathrm{H}} \\
    \dep_i &= \weight_{\mathrm{D}} \h_i + \bias_{\mathrm{D}}
\end{align}
To map the head and dependents, we use bilinear attention:
\begin{align}
    e_{ij} &= \frac{\dep_i \head_j}{\sqrt{D}} \\
    p_{ij} &= \frac{\mathrm{exp}(e_{ij})}{\sum_k \mathrm{exp}(e_{ik})}
\end{align}
where $p_{ij}$ is the probability that $w_i$ depends on $w_j$, $D$ is the dimension of hidden states.
To extract the most likely directed dependency graph from the matrix $p_{ij}$, one can use the Chu-Liu/Edmonds' algorithm \cite{chu1965shortest}. 

Conceptually, this bears a lot of similarity to Dependency Neural Selection (\textsc{DeNSe}; \cite{zhang2016dependency}).
In both cases, dependency parsing is reformulated as head selection, without ensuring a tree structure.
During inference for parsing, the same logits can be used in a spanning tree algorithm to retrieve a valid tree for the sentence.

\subsection{Dependency Mask}
Given the dependency probabilities, Structformer \citep{shen2020structformer} uses a weighted sum of matrix $p$ and $p^\top$ to produce a mask for self-attention layers in the transformer.
We found that simply using the adjacency matrix of the undirected dependency graph provides better parsing results and perplexities.
However, simply using the sum of the matrix and its transpose to create a symmetric weight matrix does not ensure that the attention mask has values < 1.
When $p_{ij = 1}$ and $p_{ji} = 1$, for instance, the mask violates the constraints of a dependency mask.
Thus, we treat $p_{ij}$ and $p_{ji}$ as parameters for independent Bernoulli variables, and we compute the probability that either $w_i$ depends on $w_j$ \textbf{or} $w_j$ depends on $w_i$.
\begin{align}
    \mask_{ij} &= p(i \rightarrow j \quad \mathbf{or} \quad j \rightarrow i) \nonumber \\
           &= p_{ij} + p_{ji} - p_{ij} \times p_{ji}
    \label{eqn:fuzzy_or}
\end{align}

\subsection{Dependency Graph Network}

\begin{figure}[h]
    \centering
    \begin{subfigure}[t]{0.45\textwidth}
        \centering
        \includegraphics[width=\textwidth]{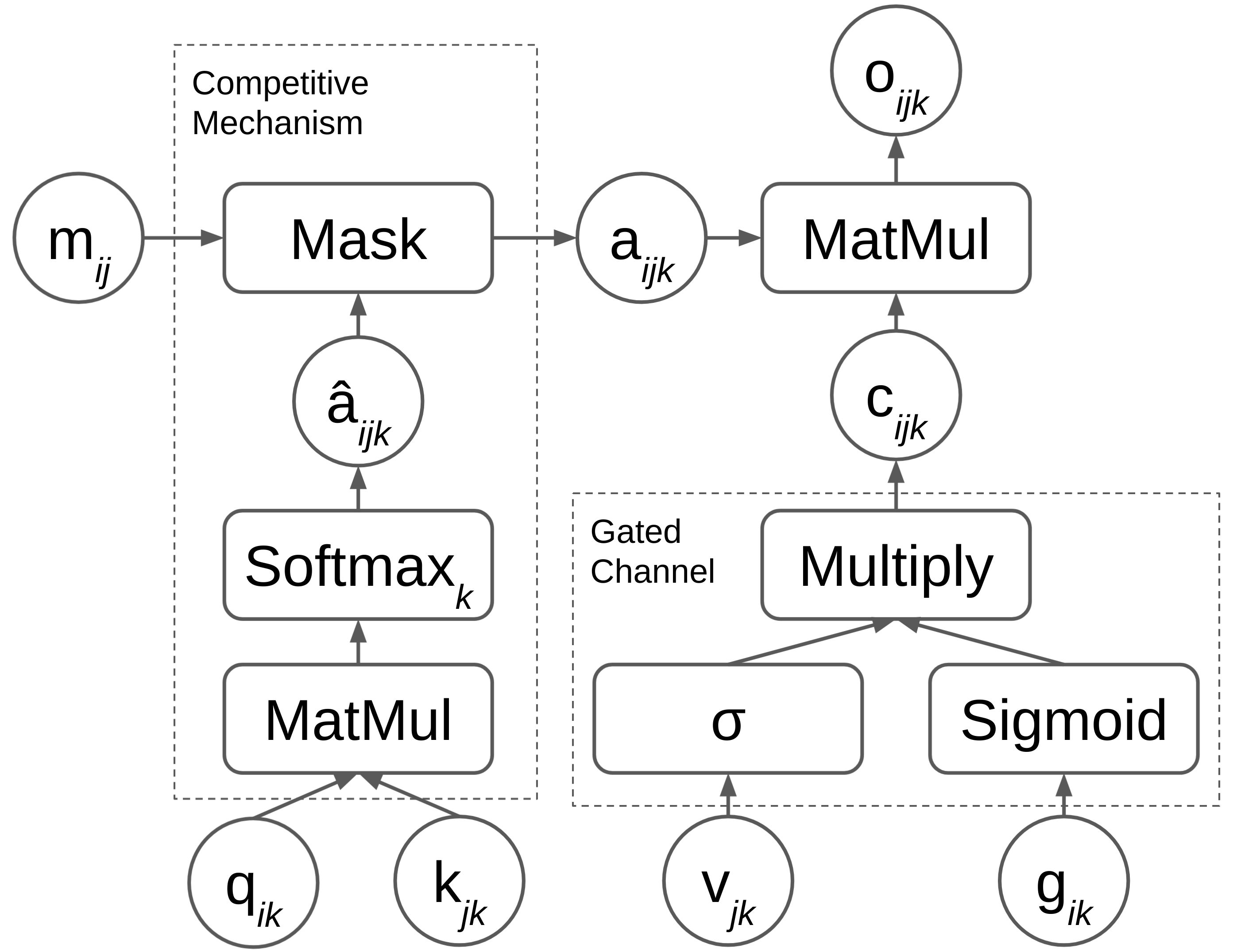}
        \caption{Competitive Mechanism and Gated Channel}
        \label{fig:gsan}
    \end{subfigure}
    \hfill
    \begin{subfigure}[t]{0.5\textwidth}
        \centering
        \includegraphics[width=\textwidth]{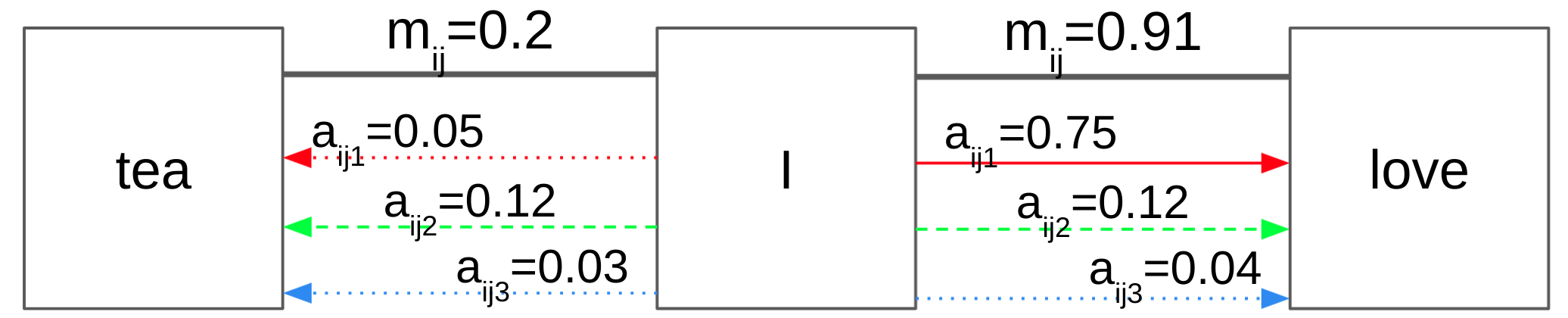}
        \caption{A example of the competitive mechanism.}
    \end{subfigure}
    \caption{For a given pair of nodes $(i, j)$, the competitive mechanism takes $\query_{i\cdot}, \key_{j\cdot}$ as input, output a probability distribution $\hat{a}_{ij}$ across different channels.
        This allows the model to select a channel for the information propagation from $j$ to $i$.
        Then the probability $\hat{a}_{ijk}$ is multiplied by dependency mask $m_{ij}$ to get $a_{ijk}$.
        The mask $m_{ij}$ functions as a macro gate to control the amount of information propagate between the node pair.
        $\hat{a}_{ijk}$ is the micro gate that controls the amount of information propagate from $j$ to $i$ through $k$-th channel.
        Each channel takes $\val_{jk}, \gate_{ik}$ as inputs and outputs respectively to represent the information that is propagated via the $k$-th channel.
        $\gate_{ik}$ allows the receiving node $i$ to filter the information.
        }
\end{figure}

To better use the dependency information, we propose a new Dependency Graph Network (DGN) with Competitive Mechanism. 
One DGN layer includes several gated channels and a competitive mechanism.
The gated channels can process and propagate information from one node to another.
Different channels can learn to process and propagate different types of information.
The competitive mechanism is designed to select the correct channel to propagate information between a specific pair of nodes.

We take inspiration from the linguistic theory that dependencies are associated with different syntactic functions.
These functions can appear as labels, e.g. ATTR (attribute), COMP-P (complement of preposition), and COMP-TO (complement of to).
However, DGN learns functions from training tasks, which in our experiments is the masked language model objective.
Since these objectives tend to be statistical in nature, these functions may not be correlated with ground truth labels given by human experts.

Inside each layer, the input vector $h_i^{l-1}$ is first projected into $N$ groups of vectors, where $N$ is the number of channels. 
Each group contains four different vectors, namely, query $\query$, key $\key$, and gate $\gate$:
\begin{align}
\left[ \begin{matrix}
\query_{ik} \\
\key_{ik} \\
\val_{ik} \\
\gate_{ik} \\
\end{matrix} \right] 
&= \weight_{\mathrm{channel}_k} \h_i^{l-1} + \bias_{\mathrm{channel}_k} 
\end{align}

\paragraph{Competitive Mechanism}

\cite{lamb2021transformers} proposed the idea of using a competition method to encourage channels to specialize over training iterations to achieve independence.
In this work, we view these gated channels as mechanisms.
Their function is to propagate information from one node to another. 
To satisfy the independence requirement, a competitive mechanism is designed to assign a channel to each pair of nodes $(i, j)$.
However discrete assignment is hard to optimize, we replace it with a soft relaxation:
\begin{align}
    e_{ijk} &= \frac{\query_{ik} \key_{jk}}{\sqrt{D}} \\
    \hat{a}_{ijk} &= \mathrm{softmax}_k (e_{ijk}) \label{eq:att_softmax}
\end{align}
where $\hat{a}_{ijk}$ is the probability that the $k$-th channel is assigned to propagate information from the $j$-th token to the $i$-th token.
To obtain the actual channel weights, we multiply the probability of edge existence with the probability of choosing a specific attention head:
\begin{align}
    a_{ijk} = \hat{a}_{ijk} \times \mask_{ij}
\end{align}
where $a_{ijk}$ is the attention weight from the $i$-th token to the $j$-th token for $k$-th attention head.

\paragraph{Gated Channel}
To model the information propagation from node $j$ to node $i$, we proposed a gated channel:
\begin{align}
    \cell_{ijk} = \sigma(\val_{jk}) \odot \mathrm{sigmoid}(\gate_{ik})
\end{align}
where $\sigma$ is a non-linear activation function, and gates $\mathrm{sigmoid}(\gate)$ allows the $i$-th token to filter the extracted information.
We also found that the gate effectively improves the model's ability to induce latent dependency structures that are coherent to human-annotated trees.
The activation function can be chosen from a wide variety of functions, including the identity function, tanh, ReLU, and ELU, etc.
According to our experiments, we found that tanh function provides the best overall performance.

At the end, a matrix multiplication is used to aggregate information from different positions.
\begin{align}
    \out_{{ik}} &= \sum_j a_{ijk} \cell_{ijk} 
\end{align}
Then, the output $\out$ from different channels are concatenated, and then projected back to the hidden state space with a linear layer.
\begin{align}
    \h_i^l = \h_i^{l-1} + \weight_o 
    \left[ \begin{matrix}
    \out_{i1} \\
    \vdots \\
    \out_{in} \\
    \end{matrix} \right] + \bias_o
\end{align}
where $\h_i^l$ is the output of the $l$-th gated self attention layers. 
The shared hidden state space can be seen as the shared global workspace \cite{goyal2021coordination} for different independent mechanisms.

\subsection{Relative Position Bias}
Transformer models use positional encoding to represent the absolute position for each token.
In DGN, we only model whether the token is before or after the current token.
The motivating intuition is the association of different channels with different directions.
In equation \ref{eq:att_softmax}, we can introduce a relative position bias:
\begin{align}
    \hat{a}_{ijk} &= \mathrm{softmax}_k (e_{ijk} + b^{lr}_k) \\
    b^{lr}_k &= \begin{cases}
        b_k^l, & i > j \\
        b_k^r, & i < j 
    \end{cases}
\end{align}
where $b_k^l$ and $b_k^r$ are trainable parameters. 
The relative position bias allows the attention head $k$ to prioritize forward or backward directions.
A mere forward and backward differentiation may seem weak compared to other parameterizations of positional encoding \cite{vaswani2017attention,shaw2018self}, but in conjunction with the dependency constraints, this method is a more effective way to model the relative position in a tree structure. 
Compared to positional encoding, this relative position bias achieves stronger masked language modeling and parsing performance.

\section{Experiments} \label{sec:dependency_exp}
In the experiment section, we first train the UDGN on the masked language modeling task, then evaluate it on masked language modeling and unsupervised parsing. 
Our experimental results show that UDGN can effectively induce the latent dependency graph from raw corpus, and achieve competitive performance on language modeling tasks.
Furthermore, all three key components of DGN play important roles in both grammar induction and language modeling.
We also finetuned the pretrained UDGN on Semantic Textual Similarity (STS) tasks.
Our experiments also show that UDGN outperforms a transformer-based model that is trained on the same corpus.

\subsection{Masked Language Modeling}

Masked Language Modeling (MLM) is a macroscopic evaluation of the model’s ability to deal with various semantic and linguistic phenomena (e.g. co-occurrence, syntactic structure, verb-subject agreement, etc). 
The performance of MLM is evaluated by measuring perplexity on masked words. 
We trained and evaluated our model on 2 different datasets: the Penn TreeBank (PTB) and BLLIP.
In our MLM experiments, each token has an independent chance to be replaced by a mask token \texttt{<mask>}, except that we never replace \texttt{<unk>} token.

\begin{table}[h]
    \centering
    \small
    \begin{tabular}{l c c c c c}
    \toprule
        \multirow{2}{*}{Model} & \multirow{2}{*}{PTB} & BLLIP & BLLIP & BLLIP \\
         &  & -SM & -MD & -XL \\
    \midrule
        Transformer & 68.9 & 44.6 & 22.8 &  17.0 \\
        StructFormer & 64.8 & 43.1 & 23.4 & 16.8 \\
        UDGN & 60.4 & 40.2 & 24.2 & 19.7 \\
    \bottomrule
    \end{tabular}
    \caption{Masked Language Model perplexities on different datasets.}
    \label{tab:mlm}
    \vspace{-0.3cm}
\end{table}

\paragraph{PTB} The Penn Treebank \cite{marcus1993building} is a standard dataset for language modeling \citep{mikolov2012statistical} and unsupervised constituency parsing \citep{shen2018ordered, kim2019compound}.
It contains 1M words (2499 stories) from Wall Street Journal.
Following the setting proposed in \citet{shen2020structformer}, we preprocess the Penn Treebank dataset by removing all punctuations, lower case all letters, and replaces low frequency tokens (< 5) with \texttt{<unk>}. The preprocessing results in a vocabulary size of 10798 (including \texttt{<unk>}, \texttt{<pad>} and \texttt{<mask>}).

\paragraph{BLLIP} The Brown Laboratory for Linguistic Information Processing dataset is a large Penn Treebank-style parsed corpus of approximately 24 million sentences from Wall Street Journal.
We train and evaluate UDGN on four splits of BLLIP: BLLIP-XS (40k sentences, 1M tokens), BLLIP-SM (200K sentences, 5M tokens), BLLIP-MD (600K sentences, 14M tokens), and BLLIP-LG (2M sentences, 42M tokens). 
Following the same setting proposed in \citet{hu2020systematic} for sentence selection, resulting in each BLLIP split being a superset of smaller splits.
All models are then tested on a shared held-out test set (20k sentences, 500k tokens).
To make the mask language modeling and parsing results comparable, we use a shared vocabulary for all splits.
Just like the PTB dataset, we preprocess the BLLIP dataset by removing all punctuations and lower case all letters.
The shared vocabulary is obtained by counting word frequencies on BLLIP-LG dataset and select the words that appear more than 27 times.
The resulting vocabulary size is 30232  (including \texttt{<unk>}, \texttt{<pad>} and \texttt{<mask>}), and covers more than 98\% tokens in BLLIP-LG split.

The word mask rate when training on both corpora is 30\%.
In Section \ref{sec:maskparsing}, we further explore the relationship between mask rate and parsing results.
Other hyperparameters are tuned separately for each model and dataset. 
The masked language model results are shown in Table \ref{tab:mlm}. 
UDGN outperforms the baselines on smaller datasets (PTB, BLLIP-SM), but underperforms against baselines trained on large datasets (BLLIP-MD, BLLIP-LG).
However, in Section \ref{sec:finetuning}, we find that the UDGN pretrained on BLLIP-LG dataset can achieve stronger performance when finetuned on a downstream task.
This may suggest that our model learns more generic contextual embeddings.


\subsection{Unsupervised Dependency Parsing}

\begin{table}[h]
    \centering
    \small
    \begin{tabular}{l l}
    \toprule
        Methods & UAS\\
        \midrule
        DMV \citep{klein2004corpus} &  35.8 \\
        E-DMV \citep{headden2009improving} & 38.2 \\
        UR-A E-DMV \citep{tu2012unambiguity} & 46.1 \\
        CS* \citep{spitkovsky2013breaking} & 64.4* \\
        Neural E-DMV \citep{jiang2016unsupervised} & 42.7 \\
        Gaussian DMV \citep{he2018unsupervised} & 43.1 (1.2) \\
        INP \citep{he2018unsupervised} & 47.9 (1.2) \\
        Neural L-PCFGs \citep{zhu2020return} & 40.5 (2.9) \\
        StructFormer \citep{shen2020structformer} & 46.2 (0.4) \\
        UDGN (Chu-Liu) & \bf 50.2 (1.5) \\
        UDGN (Argmax)$^\dagger$ & 52.5 (0.7) \\
        \bottomrule
    \end{tabular}
    \caption{Dependency Parsing Results on WSJ test set without gold POS tags. 
    Starred entries (*) benefit from additional punctuation-based constraints.
    Daggered entries ($\dagger$) takes the argmax of head distribution without a tree constraint.
    Baseline results are from \citet{he2018unsupervised}.
    UAS stands for Unlabeled Attachment Score.
    Unsupervised dependency parsing results with the knowledge of gold POS tags are excluded from this table.
    }
    \label{tab:dependency}
\end{table}












\begin{figure}
    \centering
    \includegraphics[width=0.5\linewidth]{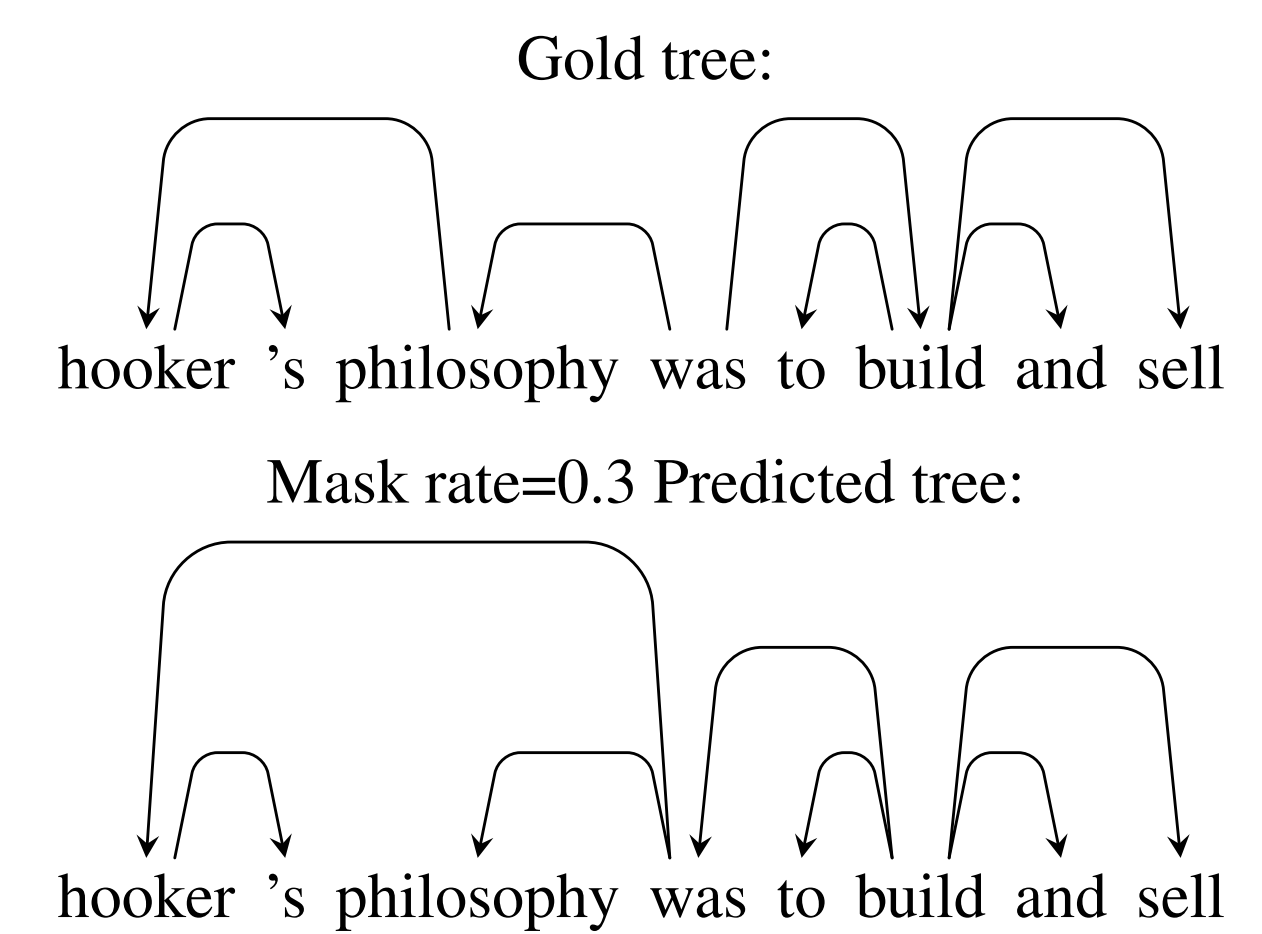}
    \caption{A example of gold tree and model generated dependency tree. 
    }
    \label{fig:dep_exp_1}
\end{figure}

We convert the human-annotated constituency trees from the Wall Street Journal test set \cite{marcus1993building} to dependency trees and use the unlabelled attachment score (UAS) as our metric.
To derive valid trees from the attention mask, we use the Chu-Liu \cite{chu1965shortest} (or Edmonds' \citealt{edmonds1967optimum})  algorithm to obtain the maximum directed spanning tree.
We also report the $\mathrm{argmax}$ over the $p$ --- we take the word at the maximum $p$ value for each word the word $i$. 
This can result in non-tree structures, but we believe that this metric gives a better indication of how often the parser predicts the right head of each word.
Following previous research \citep{shen2020structformer}, we use the model trained on the preprocessed PTB dataset (no punctuations), and test its parsing performance on section 23 of the WSJ corpus.
Punctuation is ignored during the evaluation.

Table \ref{tab:dependency} shows that our model outperforms baseline models.
This result suggests that, given our minimum inductive bias (a token must attach to another, but the graph is not necessarily a tree), 
predicting missing tokens implicitly learns a good graph that correlates well with human-annotated dependency trees.
This may suggest that some of the dependency relations proposed by linguists correspond with efficient ways of propagating information through the sentence.
Figure \ref{fig:dep_exp_1} shows a parsing example of our model after training with different mask rates.



\subsection{Correlation Between Channels and Dependency Types}

\begin{table}[t]
    \centering
    \small
    \begin{tabular}{cccccccccc}
    \toprule
        Models & prep & pobj & det & compound & nsubj & amod \\
    \midrule
        UDGN & 0.65(0.12) & 0.60(0.11) & 0.68(0.15) & 0.42(0.04) & 0.50(0.06) & 0.39(0.07) \\
        StructFormer & 0.39(0.05) & 0.38(0.07) & 0.57(0.03) & 0.33(0.01) & 0.25(0.06) & 0.26(0.01) \\
        Transformer & 0.43(0.00) & 0.46(0.03) & 0.46(0.12) & 0.30(0.01) & 0.39(0.15) & 0.26(0.02) \\
    \bottomrule
    \end{tabular}
    \caption{
        The \emph{pearson correlation coefficients} between most frequent dependency types and their most correlated channel.
        All results are average across four random seeds, standard derivation are in parentheses.
        Types are arrange from the highest frequency to lower frequency. 
    }
    \label{tab:pearson_back}
\end{table}


In this section, we test the correlation between channels and dependency types.
We consider each dependency edge $i \rightarrow j$ ($i$ depends on $j$) in the ground truth structure as a data point.
Given all the edges, we can obtain three sets of quantities: channel probabilities $A^k = \{ \hat{a}^k_{ji} \}$ and type values $Y^l = \{ y^l_{ij} \}$.
$\hat{a}^k_{ij}$ is a real value between 0 and 1, represents the probability that channels $k$ is used to model the information propagation from the child $i$ to the parent $j$.
Details about this value can be found at Equation~\ref{eq:att_softmax}.
$y^l_{ij}$ is a binary value, represents whether the label $l$ is assigned to edge $i \rightarrow j$.
We can then compute Pearson Correlation Coefficient (PCC) for every pair of $A^k$ and $Y^l$ across all ground truth edges $\{ i \rightarrow j \}$:
\begin{equation}
    \rho_{A^k,Y^l}={\frac {\mathrm{cov} (A^k, Y^l)}{\sigma_{A^k}\sigma_{Y^l}}}
\end{equation}
where $\mathrm{cov}(\cdot)$ is the covariance function, $\sigma_{\cdot}$ is the standard deviation of the respective variable.
Hence, $\rho_{A^k,Y^l}$ measures the correlation between channel $k$ and dependency type $l$.
$\rho_{A^k,Y^l} > 0$ means that the model tends to use channel $k$ for propagating information from child to parent for dependency edges of the type $l$. 
Here, we only consider the information propagation from child to parent even though information can propagate in both directions in masked language models. 

Table~\ref{tab:pearson_back} shows the PCC between the most frequent dependency types and their most correlated channels.
We can observe that all three models have channels that are positively correlated to human-annotated dependency types.
This result is coherent with the observation of \citet{htut2019attention}.
Meanwhile, the UDGN achieves a significantly better correlation than the StructFormer and the Transformer.
This confirms our intuition that competitive gated channels can better induce dependency types.

\subsection{Ablation Experiments}

\paragraph{\textbf{Mask rate and parsing interaction}} \label{sec:maskparsing}

\begin{figure}[h]
    \centering
    \includegraphics[width=0.4\textwidth]{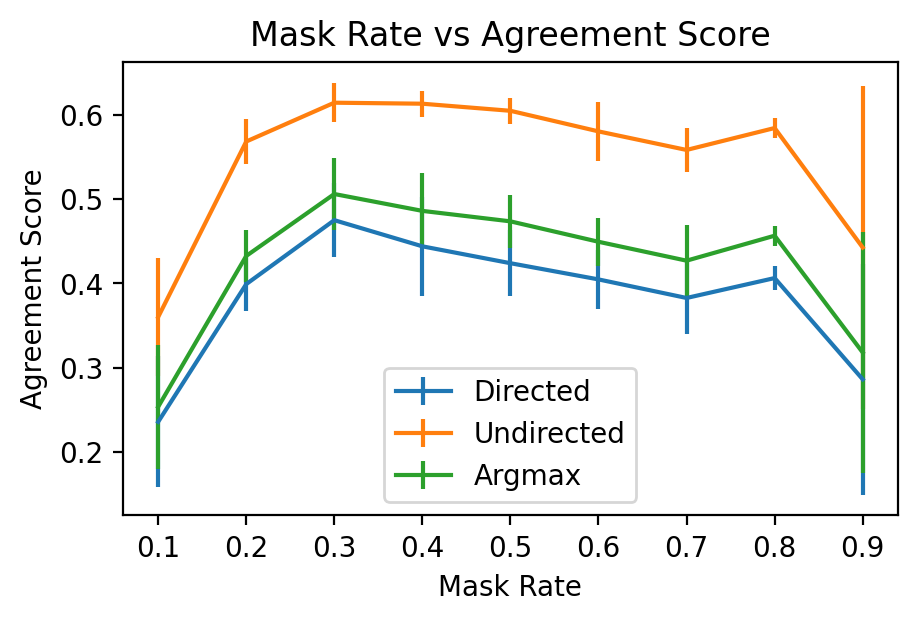}
    \caption{Relationship between parsing performance and mask rate for MLM.}
    \label{fig:mask_rate}
\end{figure}

One of the more surprising findings in our experiments with this architecture was the relationship between the word mask rate in the MLM task and how much the resulting parse trees corresponded to the ground-truth parse trees.

We trained 5 models for different word masking rates from 0.1 to 0.9, in 0.1 increments, and computed the $\mathrm{argmax}$, UAS, and undirected UAS (UUAS) scores for each of these models.
Figure \ref{fig:mask_rate} shows the plot for these results.

Firstly, we observe that the acceptable range of masking rate for achieving a decent UUAS score was fairly large: the optimal was at about 0.3, but values of 0.2 up to 0.8 worked to induce tree structures that resulted in fairly good undirected trees.
Secondly, as we move away from the optimum of 0.3-0.4, the variance of our results increases, with the highest variance when we mask at a rate of 0.9.
Finally, our model supplies the attention mask as a symmetric matrix--- the directionality of the mask is decimated when we perform Equation  \ref{eqn:fuzzy_or}.
Consequently, we find that the variance of the UAS is higher than UUAS as the connectivity of the nodes in the tree is more important than the direction of the connection in our architecture.

\paragraph{\textbf{Effects of Model Components}}

\begin{table}[h]
    \centering
    \small
    \begin{tabular}{l c c c c c c}
    \toprule
        \multirow{2}{*}{Model} & MLM & \multicolumn{2}{c}{Argmax} & \multicolumn{2}{c}{Chu-Liu} \\
         & PPL & UAS & UUAS & UAS & UUAS\\
        \midrule
        UDGN & 60.4(0.8) & 52.5(0.7) & 58.8(0.9) & 50.2(1.5) & 61.2(0.4)\\
        - Nonlinear & 61.2(1.0) & 49.5(1.1) & 56.8(1.4) & 45.6(2.0) & 60.8(1.4)\\
        - Gates & 69.5(1.9) & 31.5(2.2) & 40.7(0.3) & 26.1(2.1) & 48.9(0.5) \\
        - Competition + Sigmoid & 73.6(3.1) & 44.7(1.9) & 54.4(1.9) & 40.4(1.6) & 56.6(2.1) \\
        - Competition + Single channel &  663.1(18.6) & 3.2(0) & 6(0) & 1.3(0) & 6.1(0)\\
        - relative pos bias + pos enc & 65.2(3.4) & 47.1(7.3) & 55.4(4.1) & 44.8(7.2) & 58.2(5.2) \\
        \bottomrule
    \end{tabular}
    \caption{The performance of UDGN after removing different components. 
    ``- Nonlinear'' means remove the tanh activation function gated channels.
    ``- relative pos bias + pos enc'' means using a trainable positional encoding to replace the relative position bias.
    ``- Gates'' means remove the gate $\gate$ in gated channels.
    ``- Competition + Sigmoid'' means using a non-competitive sigmoid function to replace the competitive softmax in the competitive mechanism.
    ``- Competition + Single channel'' means using a single big channel to replace multi-channels in competitive mechanism, and the number of total remains the same.
    UUAS stands for Undirected Unlabeled Attachment Score.}
    \label{tab:ablation}
\end{table}

Table \ref{tab:ablation} shows the model's performance when individual components are removed.
The most significant decrease is caused by replacing multiple channels with one single big channel in each layer.
Although the total number of parameters remains the same, MLM and parsing performance is greatly affected.
The parser for this model cannot induce structure and predicts a uniform distribution over the other tokens.
The second, and more important, parsing performance decrease is caused by removing the gating mechanism.
This change forces each channel to always extract the same information from a given key node $h_j$, regardless of the query node $h_i$.
This has a similar effect as the previous change, reducing the diversity of different functions that can be modeled by channels.
These two observations may suggest that the diversity of information propagation function (multiple channels) is essential to induce a meaningful structure.

Another interesting observation is that relative position bias helps the model to achieve better perplexity and parsing performance in comparison with positional encoding. 
This may suggest that the combination of dependency graphs and relative position is more informative than absolute positions.

\paragraph{\textbf{Corpus Size}}
\begin{table}[h]
    \centering
    \small
    \begin{tabular}{l c c c c c c c}
    \toprule
        \multirow{2}{*}{Dataset} & \multirow{2}{*}{\#tokens} & MLM & \multicolumn{2}{c}{Argmax} & \multicolumn{2}{c}{Chu-Liu} \\
         & & PPL & UAS & UUAS & UAS & UUAS\\
        \midrule
        BLLIP-XS & 1M & 133.7(3.1) & 51.4(2.0) & 57.6(1.6) & 47.9(2.7) & 61.2(1.6) \\
        BLLIP-SM & 5M & 40.2(0.8) & 53.7(2.5) & 60.7(0.6)& 50.9(5.3) & 65.1(1.6)\\
        BLLIP-MD & 14M & 24.2(0.5) & 50.5(6.1) & 59.8(2.9) & 47.7(8.1) & 63.0(4.2) \\
        BLLIP-LG & 42M & 19.7(0.3) & 45.6(2.9) & 61.7(1.8) & 41.6(4.2) & 62.5(1.6) \\
        \bottomrule
    \end{tabular}
    \caption{The performance of UDGN after trained on different BLLIP splits. 
    Since they share the same vocabulary and test set, results are comparable.
    While UAS have a high variance, UUAS remain stable across different corpus sizes.
    Since DGN only use an undirected dependency mask, the choice of dependency direction could be arbitrary.}
    \label{tab:datasize}
\end{table}

Table \ref{tab:datasize} shows UDGN's performance after training on datasets of different sizes.
While the MLM performance improves significantly, the unsupervised parsing performance (UUAS) remains stable.
This may suggest that syntax can be acquired with a relatively small amount of data. 
It is possible then, that where extra data helps is in terms of semantic knowledge, like common sense.

\subsection{Fine-tuning}
\label{sec:finetuning}

\begin{table}[h]
    \centering
    \small
    \begin{tabular}{lcccccccc}
    \toprule
       Model & STS12 & STS13 & STS14 & STS15 & STS16 & STS-B & SICK-R & Avg. \\
    \midrule
        Transformer     & 76.17 & 61.48 & 73.97 & 74.35 & 53.72 & 64.26 & 80.00 & 69.14 \\
    \midrule
        UDGN      & 80.51 & 75.02 & 80.54 & 82.16 & 64.73 & 72.49 & 81.94 & 76.77 \\
        UDGN (Freeze parser) & 77.71 & 71.17 & 78.71 & 82.30 & 66.04 & 70.13 & 82.17 & 75.46 \\
    \bottomrule
    \end{tabular}
    \caption{
        Sentence embedding performance on STS tasks.
        All models are pretrained on BLLIP-LG, and finetuned on STS.
        The sentence embeddings are obtained by averaging the output vector across all positions.
        Freeze parser means that the parameters for the parser are not updated during finetuning.
    }
    \label{tab:main_sts}
\end{table}

In this experiment, the goal was to determine if a better representation of semantics can be encoded if the model was constrained for structure.
We pretrain a UDGN model on the BLLIP-XL dataset, and then finetune it on the STS-B \cite{cer2017semeval} dataset.
For a controlled experiment, we compare the results we attain with the previously mentioned Transformer model.
We then evaluate the resulting classifier on the STS 2012-2016 \cite{agirre2012semeval,agirre2013sem, agirre2014semeval,agirre2015semeval,agirre2016semeval}, the SICK-Relatedness \cite{marelli2014sick} dataset, and STS-B \cite{cer2017semeval}.
These datasets were downloaded and prepared using the scripts from Infersent \cite{conneau2017supervised}.
We then report the Spearman correlation score for each dataset (the `all' setting in \citealt{gao2021simcse}).

We find that the UDGN model performs better overall compared to the transformer model.
While these are not state-of-the-art results on these tasks, the purpose of our comparison was to examine the benefit of the UDGN model over the Transformer when trained on the same dataset, without conflating the effects of the pretraining dataset size.
Other models trained on more data exist, with better performance on these tasks.

\section{The Future of Dependency-based Models}
UDGN shows that dependency grammars have strong compatibility with transformer like models. 
Replacing the transformer in BERT-like models with UDGN seems to be a natural next step. 
Experimental result also suggest that UDGN seems capable of achieving better finetuning performance compared to the Transformer.
Beside simply improving performance, there may be other benefits.
Such as the ability to use it as an unsupervised dependency parser.
Structures revealed by this unsupervised parser could have interest to some linguistician and help researchers to evaluate the quality of language model.
Furthermore, the separation of parser and DGN provides a better schema for cross-lingual language models (XLMs).
Given that most languages can be parsed into the same universal dependency grammar, we could image a XLM that has a separate parser for each language and a shared DGN.
The language-specific parser should parse the input sentence into a language-agnostic dependency graph, and the shared DGN compute contextualized embedding from the graph.
Another potential application is unsupervised or semi-supervised knowledge extraction.
The unsupervised dependency parser provide a way to extract noisy relations from the training corpus. 
With some filtering and post-processing, these relations could form a large scale knowledge graph. 
\chapter{Beyond Natural Language: Hierarchical Imitation and Reinforcement Learning (HIRL)}
\label{cha:rl}

Acquiring primitive skills from demonstrations and reusing them to solve a novel long-horizon task is a hallmark in human intelligence.
For example, after learning the necessary skills (e.g., steering wheel, changing lanes) at a driving school, one could be capable of driving across the country by recombining the learned skills, which has a much longer time-scale than driving school practice.
On the other hand, the idea of decompose a long-horizon task to a sequence of skills is very coherent with the idea of parsing sentences, if two hypothesis are satisfied:
1) the agent can only execute one skill at a time;
2) once a skill is in execution, it must keep executing until it's finished.
Under this two hypothesis, we can view the normal task-skill hierarchy as a two level tree structure, and a complicated multi-level hierarchy as a multi-level tree structure.

In this chapter, we propose \emph{Option-Control Network} (OCN) -- a new HIRL model that can decompose a long-horizon task to several subtasks and separately model these subtasks as reusable skills.
We start from reviewing related works in HIRL (Section~\ref{sec:hirl_related_works}).
We then introduce OCN (Section~\ref{sec:hirl_ocn}).
The model is developed from Ordered Memory Policy Network (OMPN)~\citep{lu2021learning}. 
It includes the ordered neurons inductive bias to build a innate hierarchical structure.
Finally, we present the experiment results in two different settings (Section~\ref{sec:hirl_exp}). 
These experiment results show that OCN and effectively induce and reuse skills in the Craft environment~\citep{andreas2017modular}.

\section{Previous Approaches} \label{sec:hirl_related_works}
One general approach is to leverage the additional \emph{unstructured demonstrations} during pretraining, e.g., compILE~\citep{kipf2019compile} pretrains a VAE~\citep{kingma2013auto} on the demonstrations and uses an action decoder for finetuning. 
Our work is in this line of research.

Learning to solve temporally extended tasks is an important question for Hierarchical Reinforcement Learning (HRL). 
Different temporal abstractions are proposed to achieve structured exploration and transfer to a long-horizon task, including option frameworks~\citep{sutton1999between}, HAM~\citep{parr1998reinforcement} and max-Q~\citep{dietterich2000hierarchical}. 
With the popularity of neural nets, recent works propose to use a bi-level neural network such as option critics~\citep{bacon2017option}, feudal networks~\citep{vezhnevets2017feudal}, generative models with latents~\citep{hiro}, and modulated networks~\citep{pashevich2018modulated}. 
These models can be furthered combined with hindsight memory~\citep{levy2018hierarchical} to increase the sample efficiency. Our work can also be viewed as designing a specific neural architecture for HRL. 

However, a pure HRL method suffers from serious exploration challenges when learning from scratch~\citep{gupta2019relay}:
it takes a significant amount of samples for random walk to induce a good temporal abstraction that leads to positive rewards at the beginning of training.
A general approach to tackle this problem is to introduce a pretraining phase to ``warm up'' the policy. 
Recent works propose to pretrain the policy with an intrinsic diversity reward~\citep{eysenbach2018diversity} or language abstraction~\citep{jiang2019language}, which is shown to be useful in the HRL.
Other works~\citep{le2018hierarchical, levy2018hierarchical, gupta2019relay} have focused on learning useful skills in a pretraining phase first, and then reusing these skills when finetuning with HRL in the new environment. 
However, these methods either assume the existence of goal-conditioned policies or access to environments, which limits the practical values of these approaches. 
One general approach is to leverage the additional \emph{unstructured demonstrations} during pretraining, e.g., compILE~\citep{kipf2019compile} pretrains a VAE~\citep{kingma2013auto} on the demonstrations and uses an action decoder for finetuning. 
Our work is in this line of research.

Recent works build upon this ``imitation - finetune'' paradigm. 
With the prevalence of goal-conditioned policies~\citep{schaul2015universal,kaelbling1993learning, levy2018hierarchical} in robotics, these methods leverage demonstrations with relabelling technique to pretrain the low-level policy~\citep{gupta2019relay} or a generative model~\citep{lynch2020learning}. 
However, they exploit the fact that the ending state of a trajectory segment can be described as a point in the goal space. Hence it is difficult to apply them beyond goal-conditioned policies. 
CompILE~\citep{kipf2019compile} treats the segment boundaries as latent variables, and their model can be trained end-to-end with soft trajectory masking. 
However, CompILE requires specifying the number of segments, which is a much more limiting constraint than that required by OCN. 
Nevertheless, it is designed to be a general method so we use it as our main baseline.
Modular policy networks~\citep{andreas2017modular, shiarlis2018taco} are also used in this paradigm, where each subtask corresponds to a single modular policy. 
However, in this setting, the demonstration needs to be segmented beforehand, which requires additional human labor. 
On the contrary, our work focused on using unstructured demonstrations. 
OptionGAN~\citep{henderson2018optiongan} proposes a Mixture-of-Expert (MoE) formulation and performs IRL on the demonstration. 
However, without an explicit termination function, the learnt expert networks do not provide time-extended actions for the high-level controller. 
As a result, this method still suffers from problems of exploration with sparse rewards (as also seen in our experimental comparison with an MoE baseline).

Extracting meaningful trajectory segments from the unstructured demonstration is the focus of Hierarchical Imitation Learning (HIL). 
These works can be summarized as finding the optimal behavior hierarchy so that the behavior can be better predicted~\citep{solway2014optimal}.
DDO~\citep{fox2017multi} proposes an iterative EM-like algorithm to discover multiple levels of options, and it is applied in the continuous action space~\citep{krishnan2017ddco} and program modelling~\citep{fox2018parametrized}. 
VALOR~\citep{achiam2018variational} extends this idea by incorporating powerful inference methods like VAE~\citep{kingma2013auto}. 
Directed-Info GAIL~\citep{sharma2018directed} extracts meaning segments by maximizing the mutual information between the subtask latent variables and the generated trajectory.
Ordered Memory Policy Network (OMPN)~\citep{lu2021learning} proposes a hierarchical inductive bias to infer the skill boundaries. The above works mainly focus on skill extraction, so it is unclear how to use the segmented skills for RL finetuning. Although OCN shares a similar inductive bias with OMPN, OCN replaces the continuous hidden states communication with a softmax distribution over multiple low-level modules (options). This enables OCN to model different subtasks with different options and to effectively reuse them in a new task. 

\section{Option-Controller Network}\label{sec:hirl_ocn}

\begin{figure}[h]
    \centering
    \includegraphics[width=\linewidth]{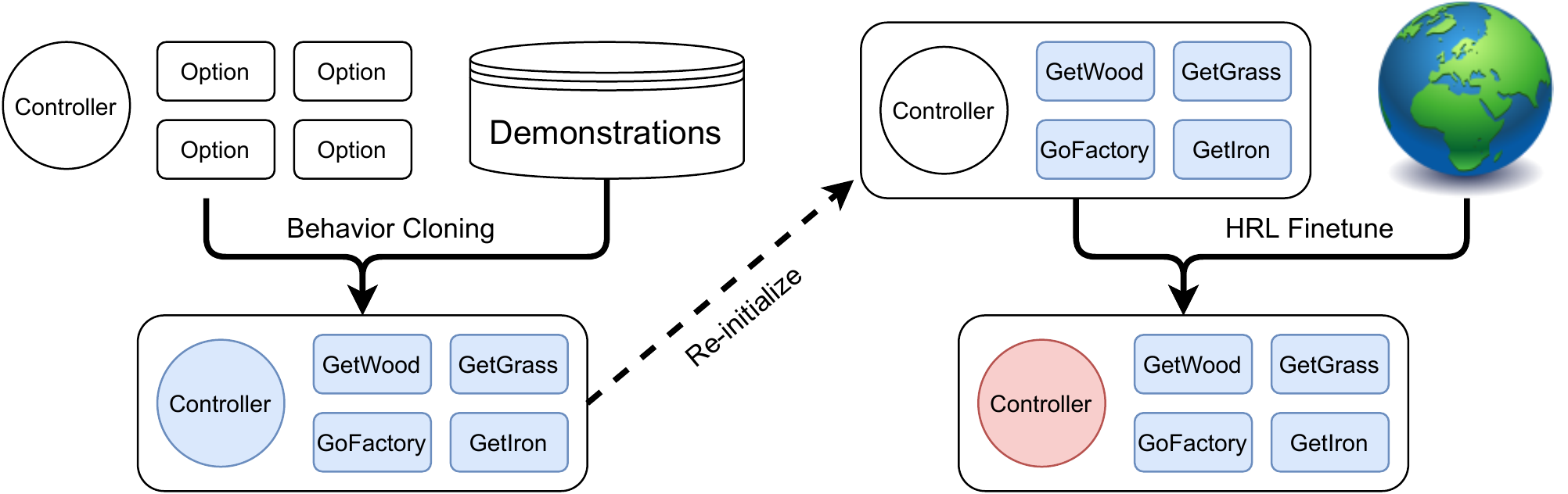}
    \caption{The training pipeline of OCN. Our model is composed of a controller (circle) and a options pool (rectangles). 
    The controller and options are randomly initialized, which means each option does not correspond to a meaningful subtask. After behavior cloning, both options and controllers are induced (marked blue) and the options correspond to meaningful subtasks from demonstrations (e.g., get wood).
    Then we freeze the parameters in the options and re-initialize the controller. 
    The controller is trained to adapt to the new environment with HRL (marked red).
    }
    \label{fig:option-controller}
\end{figure}

In this section, we propose \emph{Option-Control Network} (OCN). 
Unlike previous works, our method does not require generative models~\citep{eysenbach2018diversity}, goal-conditioned policies~\citep{gupta2019relay}, pre-specified policy sketch~\citep{shiarlis2018taco} or constraints on the number of segments~\citep{kipf2019compile}, making our approach conceptually simple and general.
An OCN includes a set of $N$ options $\left\{ \option_1, ..., \option_N \right\}$ and a controller $\controller$.
As shown in figure \ref{fig:option-controller}, the OCN starts by using the controller to choose an option to execute the first subtask.
Once the subtask is done, the controller will choose another option to execute the second subtask, and so on, until the goal of the task is achieved.
Inspired by OM and OMPN~\citep{lu2021learning}, we use the ordered neurons inductive bias to enforce the hierarchical constraint between the controller and the options so that the high-level controller is updated less frequently than the low-level options while keeping the model end-to-end differentiable. 

Another intuition behind the OCN is making controllers and options running independently.
So options can be easily reused by another controller to solve a new task.
Each component directly takes raw observation as input and only interacts with other components through the probability distribution $p_t$. 
In other words, an option can always execute the induced skill by itself, regardless of the higher-level task. 
A controller can consider options as black boxes with an on/off switch so the controller can achieve structured exploration.
Given a new (possibly) complicated and long-horizon task and a set of learnt options, the controller can then quickly find a solution from the induced option space, enabling structured exploration.

As shown in Figure~\ref{fig:option-controller}, OCN can jointly learn options and controllers with multitask behavior cloning from unstructured demonstrations. 
When given a new task, one could perform HRL finetuning by re-initializing the controller and freezing the options. 
This enables our model to generalize combinatorially to unforeseen conjunctions \citep{denil2017programmable}.

\subsection{Option and Controller}
\begin{figure}[h]
    \centering
    \includegraphics[width=\linewidth]{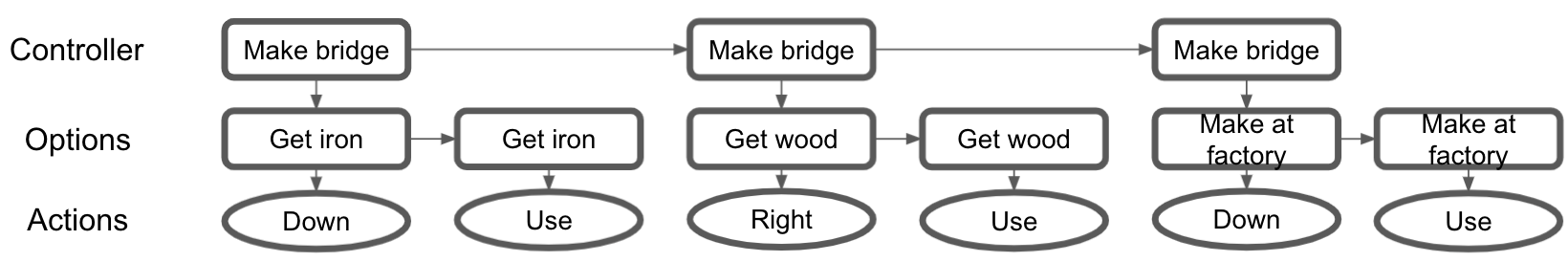}
    \caption{An example of OCN. 
    The controller $\controller$ models the task \textit{make bridge}.
    Three options separately model subtasks \textit{get iron}, \textit{get wood} or \textit{make at factory}.}
    \label{fig:ocn_example}
\end{figure}

\textbf{Option} As shown in the middle of Figure~\ref{fig:ocn_example}, an option $\option_i$ models a skill that can solve one specific subtask, for example \textit{get wood}, \textit{get iron} or \textit{make at workbench}.
It can be described as:
\begin{equation}
    \prob^{\option}_{i,t}, \hid^{\option}_{i,t}, e_{i,t} = \option_i(\x_t, \hid^{\option}_{i,t-1})
\end{equation}
where $\x_t$ is the observation at time step $t$, and $\hid^{\option}_{i,t-1}$ is the hidden state of the respective option at time step $t-1$;
$\hid^{\option}_{i,t}$ is the hidden state of $\option_i$ at time step $t$; 
$e_{i,t}$ is a scalar between 0 and 1, represents the probability that the current option is done;
$\prob^{\option}_{i,t}$ is a distribution of actions, including \textit{move up}, \textit{move left} and \textit{use}. 
These actions are the smallest elementary operations that an agent can execute.
During the execution of an option, if probability $e_{i,t}$ is 0, the option will keep executing the current subtask; if $e_{i,t}$ is 1, the option will stop the execution and return to the controller for the next subtask. In our work, each option maintains a separate set of parameters.

\textbf{Controller} As shown at the top of Figure~\ref{fig:ocn_example}, a controller $\controller$ models a higher level task, like \textit{make bed}, \textit{make axe}, or \textit{make bridge}.
Each of these tasks can be decompose to a sequence of subtasks. For example, \textit{make bridge} can be decompose to 3 steps: 1) \textit{get iron}, 2) \textit{get wood}, 3) \textit{make at factory}.
Thus a controller can also be represented as:
\begin{equation}
    \prob^{\controller}_{t}, \hid^{\controller}_{t}, e^{\controller}_t = \controller(\x_t, \hid^{\controller}_{t-1})
\end{equation}
where $\prob^{\controller}_{t}$ is a distribution over the set of options $\left\{ \option_i \right\}$, $\hid^{\controller}_{t}$ is the hidden state for controller,
$e^{\controller}_t$ is the probability that the current task is done.
In this OCN architecture, we don't need the $e^{\controller}_t$, since the environment will provide signal(reward) once the task is done. 
However, OCN can be easily expanded to a multi-level model. In this multi-levels model, a set of multiple controllers become options for a higher-level controller, and their respective tasks become subtasks for a more complicated task.

\textbf{Cell Network}
In OCN, options and controllers share the same format for input and output.
Thus, we parameterize them with the same neural network architecture.
To model the policy of controllers and options, we proposed the following cell network:
\begin{align}
    \hat{h}_t &= \mathrm{MLP} \left(
    \left[ \begin{matrix}
    x_t,
    h_{t-1}
    \end{matrix} \right] \right)  \\
    p_t &= \mathrm{softmax}(\mathbf{W}_{\mathrm{act}} \hat{h}_t + \mathbf{b}_{\mathrm{act}}) \\
    h_t &= \mathrm{tanh}(\mathbf{W}_{\mathrm{hid}} \hat{h}_t + \mathbf{b}_{\mathrm{hid}}) \\
    e_t &= \mathrm{sigmoid}(\mathbf{w}_{\mathrm{end}} \hat{h}_t + b_{\mathrm{end}})
\end{align}
$x_t$ is the raw observation, the shape of the vector depends on the environment.
$h_{t-1}$ is the recurrent hidden state of size $d_{\mathrm{hid}}$, it allows the model to remember important information from previous time steps.
$\mathrm{MLP}$ is a multi-layer neural network of Depth $l_{\mathrm{MLP}}$ and hidden size $d_{\mathrm{MLP}}$.
We use $\mathrm{tanh}$ as activation function for $\mathrm{MLP}$.
$\hat{h}_t$ is a vector of size $d_{\mathrm{MLP}}$.
$\mathbf{W}_{\mathrm{act}}$ is a matrix of size $n_{\mathrm{act}} \times d_{\mathrm{MLP}}$, where $n_{\mathrm{act}}$ is number of actions.
$\mathbf{W}_{\mathrm{hid}}$ is a matrix of size $d_{\mathrm{hid}} \times d_{\mathrm{MLP}}$.
$\mathbf{w}_{\mathrm{end}}$ is a vector of size $d_{\mathrm{MLP}}$.
Following the fast and slow learning idea proposed in \cite{madan2021fast}, we introduce a temperature term $T$ to controller's softmax function:
\begin{equation}
    p^{\controller}_t = \mathrm{softmax} \left( \frac{\mathbf{W}_{\mathrm{act}} \hat{h}_t + \mathbf{b}_{\mathrm{act}}}{T} \right)
\end{equation}
A large temperature $T$ allows the option to output smoother distribution at the beginning of training. 
It also reduces the scale of gradient backpropagated into the controller.
This results in the controller changes and updates slower than options.
We found $T$ makes OCN become more stable in imitation learning and converge to a better hierarchical structure.

\subsection{Option-Controller Framework} 

\begin{figure}[h]
    \centering
    \begin{subfigure}[t]{0.25\textwidth}
         \centering
         \includegraphics[width=\textwidth]{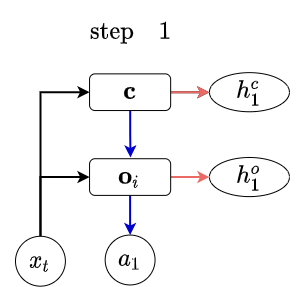}
         \caption{First time step}
         \label{fig:step1}
    \end{subfigure}
    \hfill
    \begin{subfigure}[t]{0.32\textwidth}
         \centering
         \includegraphics[width=\textwidth]{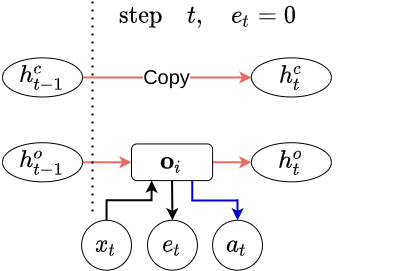}
         \caption{Inside a subtask}
         \label{fig:not_end}
    \end{subfigure}
    \hfill
    \begin{subfigure}[t]{0.37\textwidth}
         \centering
         \includegraphics[width=\textwidth]{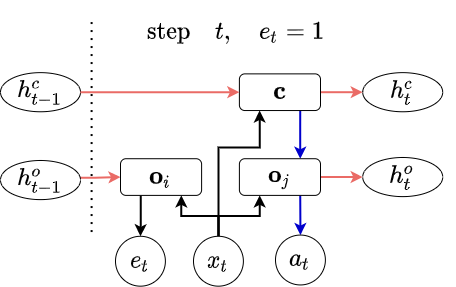}
         \caption{Switching between subtasks}
         \label{fig:end}
     \end{subfigure} 
    \caption{The three different phase of OCN: 
    (a) At the first time step, the controller selects an option $\option_i$;
    The option $\option_i$ outputs the first action $\action_1$. 
    (b) If the previous option $\option_i$ predict that the subtask is not finish;
    The option $\option_i$ then continue outputs action $\action_t$;
    The controller hidden state is copied from previous time step. 
    (c) If the previous option $\option_i$ predict that the subtask is done;
    The controller then selects a new option $\option_j$ and updates the controller hidden state;
    The new option $\option_j$ outputs action $\action_t$.
    Blue arrows represent probability distributions output by controller and options.
    Red arrows represent recurrent hidden states between time steps.}
    \label{fig:my_label}
\end{figure}

Given the definition for options and controllers, we can further formulate OCN.
As shown in Figure \ref{fig:step1}, at the first time step, the controller computes a probability distribution over options for the first subtask, and options execute their first steps:
\begin{align}
    \prob^{\controller}_{1}, \hid^{\controller}_{1} 
    &= \controller(\x_1, \hid^{\controller}_0) \\
    \prob^{\option}_{i,1}, \hid^{\option}_{i,1}, e_{i,1} 
    &= \option_i(\x_1, \hid^{\option}_{i,0}) \\
    \prob^{\action}_1 &= \sum_i p^{\controller}_{1,i} \prob^{\option}_{i,1}
\end{align}
where $h^{\controller}_0$ and $h^{\option}_{i,0}$ are initial hidden states for controller and options, $p^{\action}_1$ is a distribution for actions.
The output $p^{\action}_1$ is formulated as a mixture of experts, where experts are options and the gating model is the controller.

At time steps $t > 1$, the options first execute one step to decide whether this subtask is done.
If the subtask is unfinished, the option then outputs an action distribution, as shown in Figure \ref{fig:not_end}:
\begin{align}
    \hat{\prob}^{\option}_{i,t}, \hat{\hid}^{\option}_{i,t}, e_{i,t}
    &= \option_i(\x_t, \hid^{\option}_{i,t-1}) \\
    e_t &= \sum_i p^{\controller}_{t-1, i} e_{i,t} \\
    \hat{\prob}^{\action}_{i,t} &= \sum_i p^{\controller}_{t-1, i} \hat{\prob}^{\option}_{i,t}
\end{align}
where $e_t$ is the probability that the previous subtask is done and $\hat{\prob}^{\action}_{i,t}$ is the action distribution if the subtask is not done.
If the previous subtask is done, the controller $\controller$ need to select a new option distribution for the next subtask and reinitialize the option, as shown in Figure \ref{fig:end}:
\begin{align}
    \prob'^{\controller}_{t}, \hid'^{\controller}_{t} 
    &= \controller(\x_t, \hid^{\controller}_{t-1}) \\
    \prob'^{\option}_{i,t}, \hid'^{\option}_{i,t}, e'_{i,t} 
    &= \option_i(\x_t, \hid^{\option}_{i,0}) \\
    \prob'^{\action}_t &= \sum_i p'^{\controller}_{t,i} \prob'^{\option}_{i,t}
\end{align}
where $\hid'^{\controller}_{t}$, $\hid'^{\controller}_{t}$, $\prob'^{\controller}_{t}$ and $p'^{\controller}_{t,i}$ are hidden states and distributions for the next subtask if the previous subtask is done.
Thus, we can formulate the output at time step $t$ as a weighted sum of the two situations:
\begin{align}
    \left[ \begin{matrix}
        \hid^{\controller}_{t} \\
        \prob^{\controller}_t \\
        \hid^{\option}_{t} \\
        \prob^{\action}_t
    \end{matrix} \right] 
    &= e_t \left[ \begin{matrix}
        \hid'^{\controller}_{t} \\
        \prob'^{\controller}_t \\
        \hid'^{\option}_{t} \\
        \prob'^{\action}_t
    \end{matrix} \right]  
    + (1 - e_t) \left[ \begin{matrix}
        \hid^{\controller}_{t-1} \\
        \prob^{\controller}_{t-1} \\
        \hat{\hid}^{\option}_{t} \\
        \hat{\prob}^{\action}_t
    \end{matrix} \right] 
    \label{eq:hierarchy}
\end{align}
The equation~\ref{eq:hierarchy} provides OCN an internal hierarchical inductive bias, that a higher-level component ($\controller$) only update its recurrent hidden state and output a new command ($\prob'^{\controller}_t$) when its current functioning subordinate ($\option_i$) reports ``done''.

\subsection{Inducing and Reusing Skills}
\textbf{Imitation Learning and Inducing Skills}
OCN imitates and induces skills from unstructured demonstrations.
In the rest of this paper, $\dem$ represents an unstructured demonstration $\{(x_t, a_t)\}_{t=1}^T$
$\dems$ represents a set of demonstrations of different tasks $[(\dem_1, \task_1), (\dem_2, \task_2), ...]$, where $\task_i$ are task ids, belongs to a shared task set $\tasks$.

Given a demonstration $\dem$, OCN can perform behavior cloning with a negative log-likelihood loss:
\begin{align}
    loss &= \mathrm{average}_t \left( \mathrm{NLLLoss} (\prob^{\action}_t, a_t) \right)
\end{align}
For different tasks $\task$, we can use two different methods to model their associated controllers.
The first method is to assign one controller $\controller_{\task}$ and a initial hidden state $\hid^{\controller}_{\task,0}$ to each $\task$.
The second method is to share the controller $\controller$, but assign a different initial hidden state $\hid^{\controller}_{\task,0}$ to each $\task$.
We choose the second method in this work because sharing $\controller$ could avoid the risk that different controllers choose to model the same subtask with options.
During the imitation learning, OCN allows gradient backpropagation through all probabilities $\prob$.
Thus, the gradient descent will try to induce an optimal set of options that can best increase the likelihood of the data.

\textbf{Reinforcement Learning and Reusing Skills}
Given the induced options from imitation learning, our model can learn to solve a new task by reusing these skills via reinforcement learning.
For example, after training on demonstrations of task 1 and task 2, OCN induce $N$ options $\{\option_1, ..., \option_N\}$.
Given a new task 3 without demonstrations, we can initialize a new controller $\controller_3$, that takes observations as input and outputs a probability distribution over $N$ induced options.
To learn $\controller_3$, we freeze all options and use PPO~\citep{schulman2017proximal} algorithm to learn $\controller_3$ from interactions with the environment. 

\begin{algorithm}[h]
    \centering
    
    \caption{PPO, Adapt OCN to a new task}
    \label{alg:ppo}
    \begin{algorithmic}[1]
    	\State Initialize controller $\controller$
        \State Freeze all options $\{\option_{1...N}\}$
        \For{iterations=1,2,...}
            \For{actor=1,2,...}
                \For{step t=1,2,...,T}
                    \State $p^\controller = \controller(x_t, h^{\controller}_{t-1})$
                    \State $i = \mathrm{sample}(p^\controller)$
                    \State Rollout $\option_i$\ until sample$(e_i)=1$
                \EndFor
                \State Compute advantage estimates $\hat{A}_1, ..., \hat{A}_T$
            \EndFor
        \State Optimize surrogate $L$ wrt $\controller$, with $K$ epochs and minibatch size $B$
        \EndFor
    \end{algorithmic}
\end{algorithm}

During the training, once the controller outputs an option distribution $p^{\controller}$, OCN samples from the distribution, the sampled option will rollout until it's done, then the process will repeat until the task is solved.  We outline the process in Algorithm~\ref{alg:ppo}. Thus, in the RL phase, our model only needs to explore at options space, which significantly reduces the number of interaction steps to solve the new tasks.

\section{Experiments}\label{sec:hirl_exp}

We perform experiments in Craft~\citep{andreas2017modular}, a grid-world environment focusing on navigation and collecting objects.
Our results show that with unstructured demonstrations, OCN can jointly learn to segment the trajectories into meaningful skills as well as model this rich set of skills with our pool of low-level options. 
During HRL finetuning, we show that OCN achieves better performance in more complex long-horizon tasks with either sparse or dense reward compared with existing baselines. We also provide further visualization to show the discovered options are reused during finetuning.

\paragraph{\textbf{Environment}}
Craft is adapted from previous works \citep{lu2021learning, andreas2017modular}. 
In this environment, an agent can move in a 2D grid map with actions (\textit{up}, \textit{down}, \textit{left}, \textit{right}) and interact with the objects with the action \textit{use}.
In our experiments, a subtask requires the agent to locate and collect a specific type of object.
For example, subtask A, \textit{get wood}, requires the agent to first navigate to the block that contains wood, then execute a \textit{use} action to collect one unit of wood.
Table \ref{tab:subtasks} provides a list of subtasks used in our experiments.
A task requires the agent to finish a sequence of subtasks in the given order.
The environment can provide either sparse reward or dense reward.
In the dense reward, the agent receives rewards after completing each subtask while in the sparse reward, the agent only receives rewards after completing all subtasks.

\begin{table}[h]
    \centering
    \begin{tabular}{r c c c c}
        \toprule
        Subtask & A & B & C & D \\
        \midrule
        Goal & \textit{get wood} & \textit{get gold} & \textit{get iron} & \textit{get grass} \\
        \bottomrule
    \end{tabular}
    \vspace{0.2cm}
    \caption{Subtasks and their goals}
    \vspace{-0.3cm}
    \label{tab:subtasks}
\end{table}

\paragraph{\textbf{Baselines}}
We compare OCN with a number of baselines including task decomposition methods and hierarchical methods. Our baselines are:
(1) compILE~\citep{kipf2019compile}, which leverages Variational Auto-Encoder to recover the subtask boundaries and models the subtasks with different options.
(2) OMPN ~\citep{lu2021learning} which studies inductive bias and discovers hierarchical structure from demonstrations.
(3) Mixture-of-Experts (MOE), which uses a similar architecture as OCN, but the options are only executed for one time step. This baseline is inspired by OptionGAN~\citep{henderson2018optiongan} which proposes the MoE framework without modeling termination functions. 

\paragraph{\textbf{Implementation Details}}
We train all imitation learning methods by utilizing behaviour cloning with a batch size of 512 and a learning rate of 0.001. For each task, we sample 6000 demonstrations and split 80\% for training and 20\% for validation. 
For reinforcement learning, we use PPO algorithm with a batch size of 1024 and a learning rate of 0.0003. 
We use Adam optimizer and a linear schedule to adjust the learning rate. The hidden state size $d_{\mathrm{hid}}$ of OCN and baselines is 128.
The depth $l_{\mathrm{MLP}}$ of cell network is 2.
The temperature $T$ for controller's softmax function is 10. 



\subsection{S1: Transferring from Single Model}

\begin{figure}[h]
    \centering
    \includegraphics[width=\linewidth]{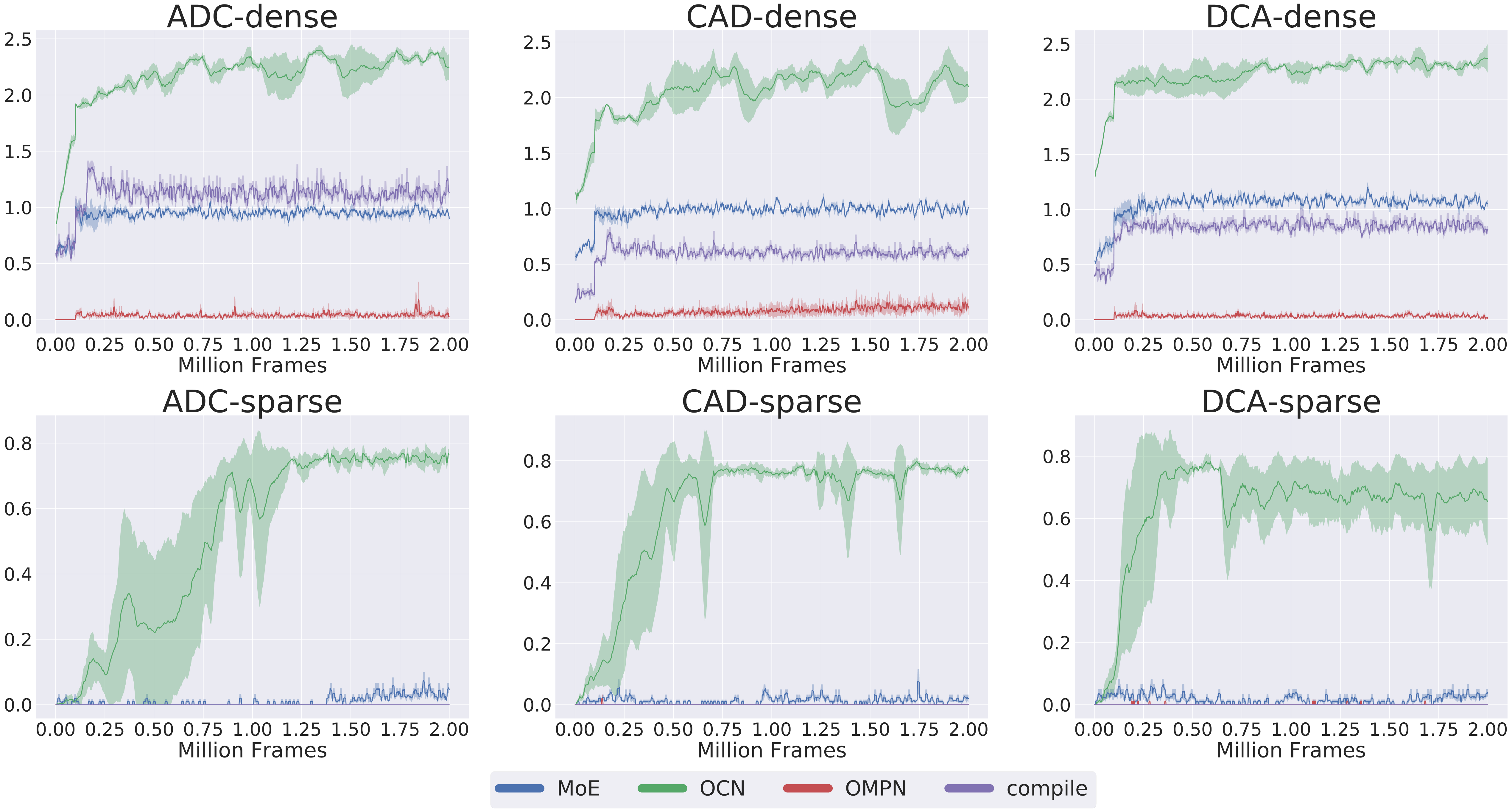}
    \caption{The learning curve of different methods on three finetuning tasks of \textbf{S1}.
    \texttt{dense} means dense reward setting. 
    \texttt{sparse} means sparse reward setting.}
    \label{fig:exp1}
\end{figure}

In this setting, the training task set is  $\{ \mathrm{AC, CD, DA} \}$.
During the imitation phase, we pretrain an OCN with one controller $\controller_1$ and three options $\{ \option_1, \option_2, \option_3 \}$ to imitate these demonstrations.
During the fine-tuning phase, the model needs to solve three new tasks: $\{\mathrm{ADC, CAD, DCA}\}$. We initialize a new controller for each while freezing the parameters of options. This is the classical setting where an agent is required to learn skills from short expert demonstrations and to transfer to long-horizon tasks.

As is shown in Figure~\ref{fig:exp1}, our method converges faster and achieves higher performance than baselines in both dense and sparse reward settings.
With dense rewards, our method achieves double the returns than the strongest baseline.
In the sparse reward setting, our method can get an average return of 0.7 with the maximum being 1, while other baselines struggle.  
We find that MoE fails to achieve similar performance even with a very similar architecture as OCN, since MoE does not model the termination function and the controller selects a new option every time step.
This result shows that exploring in option space is more efficient than other schemes, provided the new task is expressible as a combination of previous observed subtasks.

\subsection{S2: Transferring from Multiple Models}

\begin{figure}[h]
    \centering
    \includegraphics[width=\linewidth]{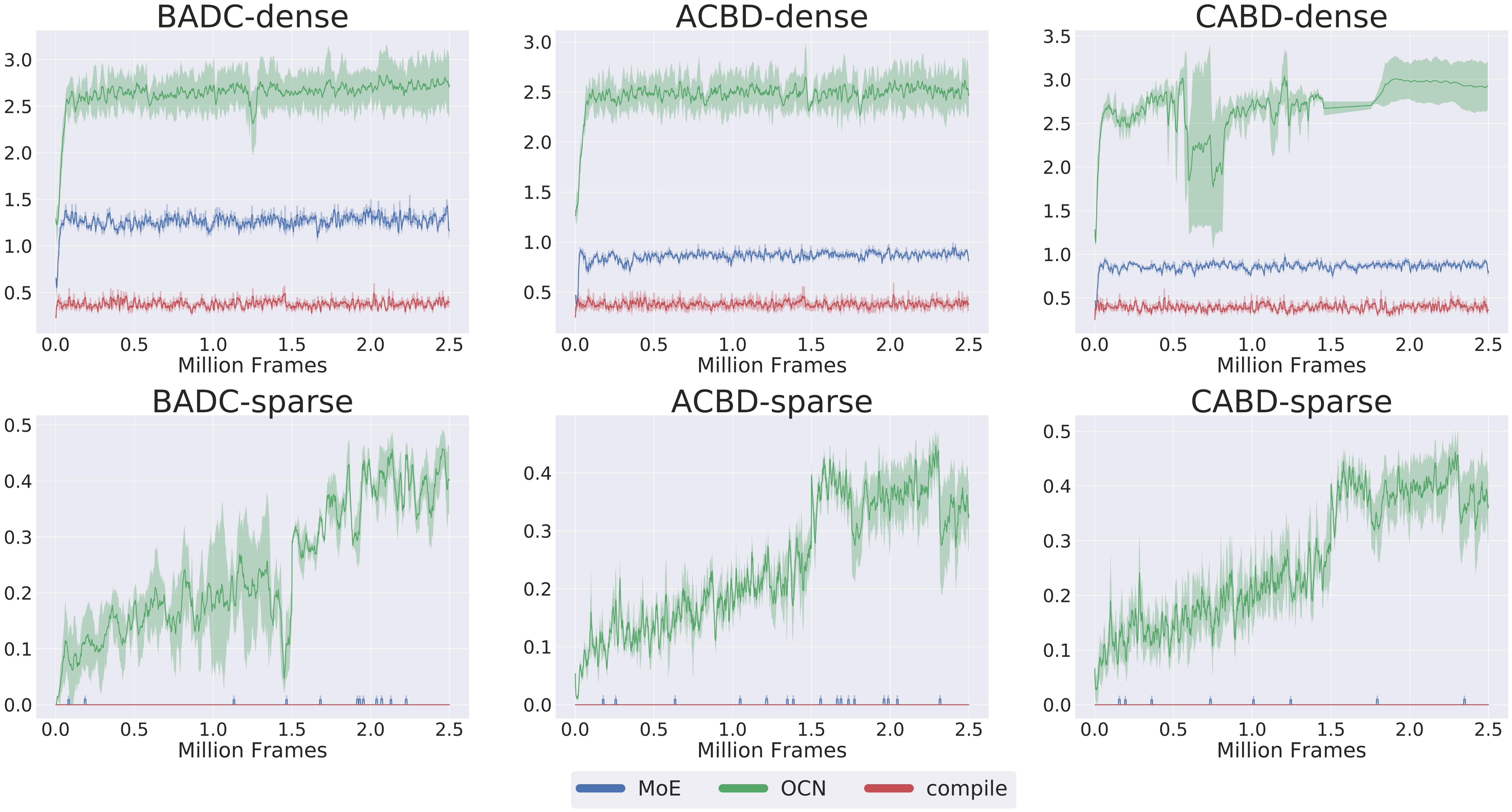}
    \caption{The learning curve of different methods on three finetuning tasks of \textbf{S2}. OMPN is not included because it does not learn an explicit set of options.}
    \label{fig:exp2}
\end{figure}

In this setting, we have two disjoint task set. The first set is $\{\mathrm{AB, BA}\}$ and the second task set is $\{\mathrm{CD, DC}\}$
We train two separate OCN models. Each model includes a controller and two options. Thus, at the end of imitation phase, we obtain four options $\{ \option_1, ..., \option_4 \}$. Then we initialize three new controllers to solve three new tasks: $\{\mathrm{BADC, ACBD, CABD}\}$.

This setting is related to the problem of data islands and federated learning~\citep{yang2019federated}, where two companies could each pretrain models on their separate datasets, merge the induced options, and share the controller finetuned on more challenging tasks. This is made possible because of the highly \emph{modularized design} of our architecture.

The results are shown in Figure~\ref{fig:exp2}.
We show that our model can still reuse merged option pools, while other baseline methods fail at this setting.
CompILE uses a continuous latent variable for communication between the controller and the action decoder, which causes compatibility issues while merging skills from different models. 
The MoE method still suffers from the long horizon problem.
Overall, this result highlights the flexibility of OCN and its promise in maintaining data privacy for collaborative machine learning applications. 

\subsection{Model Analysis}


\paragraph{\textbf{Visualization}}
\begin{figure}[h]
    \centering
    \includegraphics[width=1.0\linewidth]{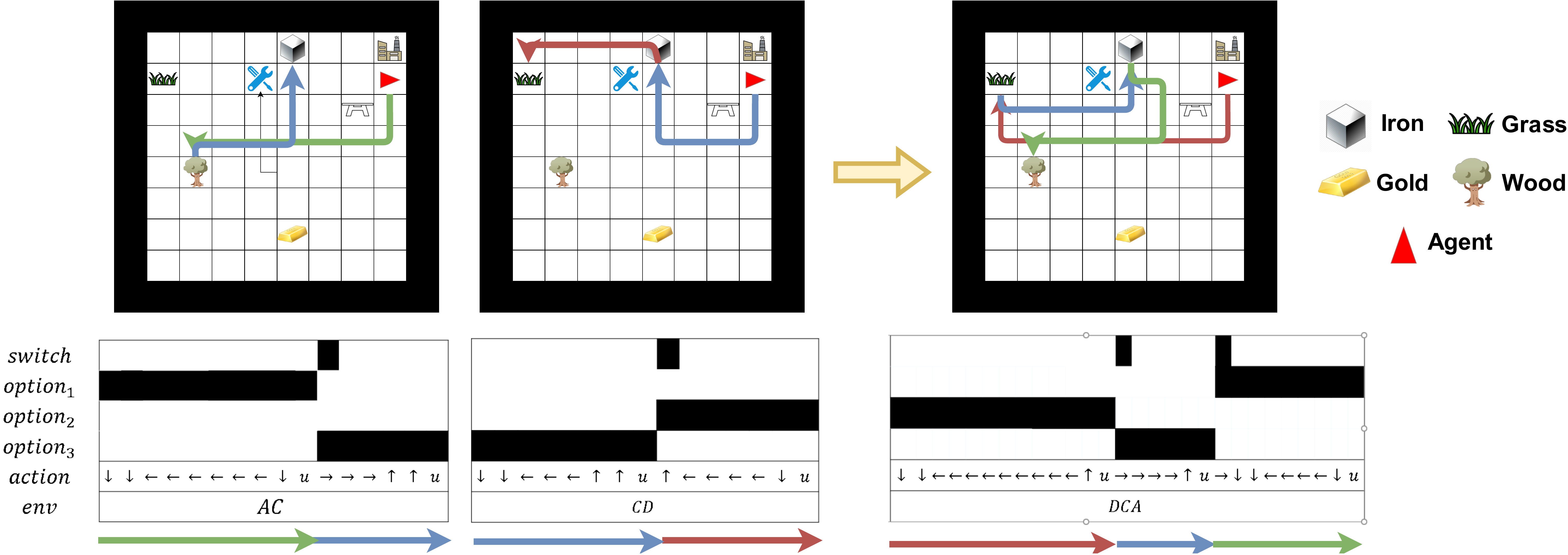}
    \caption{A trajectory of model finetuned on task DCA in \textbf{S1}.
    \texttt{switch} represents the value of $e_t$ at every time step. The option distribution is computed with $\prob^{\controller}_{t}$.}
    \label{fig:viz}
\end{figure}
Figure~\ref{fig:viz} shows a trajectory of the model, which includes options trained on tasks (AC, CD), and controller finetuned on task DCA.
As shown in the Figure~\ref{fig:viz}, the discovered options from the demonstrations with shorter horizon (AC, CD) are reused when we finetune the controller in a longer horizon task (DCA).
This observation confirms our hypothesis, that options can model subtasks and be reused for another task, from a qualitative perspective. 

\paragraph{\textbf{Quantitative Analysis}}
\begin{figure}[ht]
    \centering
    \includegraphics[width=\linewidth]{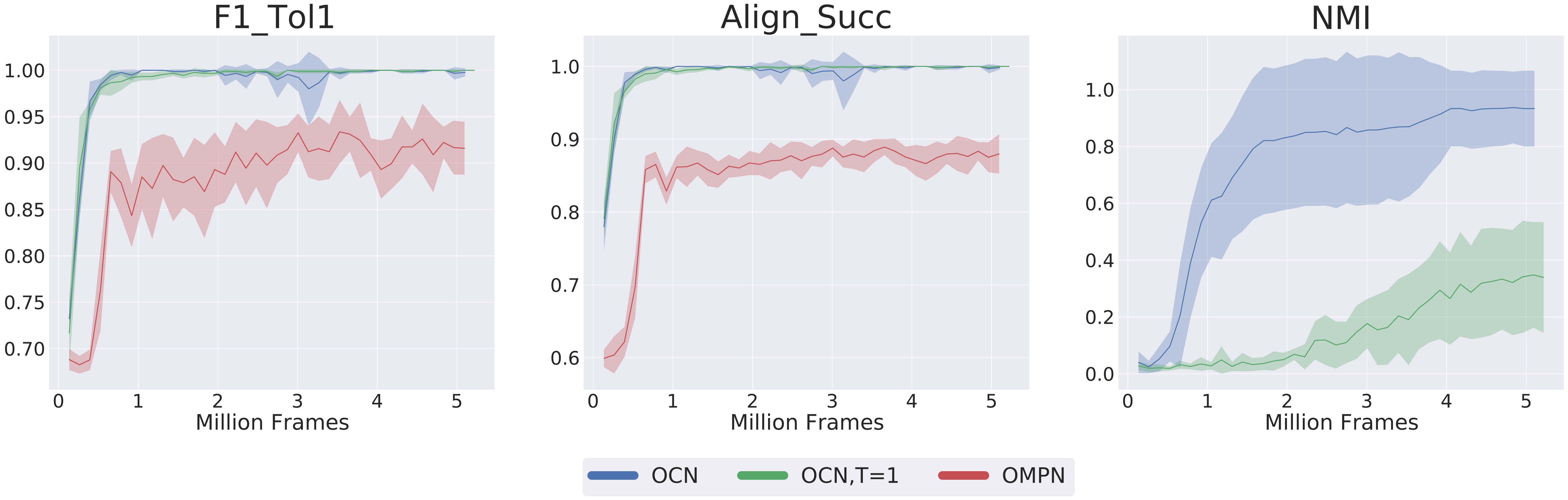}
\caption{Comparison of unsupervised trajectory parsing results during the imitation phase with OMPN~\citep{lu2021learning}. 
We use F1 scores with tolerance~\textbf{(Left)} and Task Alignment~(\textbf{Center}) to show the quality of learned task boundaries. 
We compute the normalized mutual information (\textbf{Right}) between the emerged option selection $\prob^{\controller}_t$ and the ground-truth to show that our model learns to associate each option to one subtask.
\texttt{T=1} means that the temperature term in the controller is removed.}
\label{fig:parsing}
\end{figure}

Figure~\ref{fig:parsing} shows the performances of parsing and option-subtask correlation during imitation phase.
We find that OCN can converge faster and achieve better parsing performance than the OMPN model. 
Figure~\ref{fig:parsing} right shows that, during the imitation phase, randomly initialized options slowly converged to model different subtasks.
At the end of imitation, OCN shows strong alignment between options and subtasks.
In 4 out of 5 runs, OCN actually achieves NMI=1, which means that the alignment between option and subtask is perfect.
On the other hand, if we remove the temperature term (i.e. set $T=1$) in controller, the NMI drops significantly. 
\begin{table}[h]
    \centering
    \begin{tabular}{c | c c c}
    \toprule
        \diagbox{Option}{subtask} & A & C & D \\
        \midrule
        1 & 0.96 & 0.07 & 0.03 \\
        2 & 0.00 & 0.02 & 0.95 \\
        3 & 0.01 & 0.98 & 0.01 \\ 
    \bottomrule
    \end{tabular}
    \vspace{0.2cm}
    \caption{The success rate of each option when testing on different subtasks.}
    \label{tab:subtask_succ}
    \vspace{-0.5cm}
\end{table}
This result suggest that the fast and slow learning schema is important for the model to learn the correct alignment between options and subtasks.
Furthermore, Table~\ref{tab:subtask_succ} shows the success rate of using each option to solve each subtask. We find that there is a one-to-one correspondence between subtasks and learnt options. 
Overall, these results confirmed our hypothesis that options can be used to model subtasks from a quantitative perspective.

\subsection{Hyperparameters Analysis}

\begin{figure}[h]
    \centering
    \includegraphics[width=\linewidth]{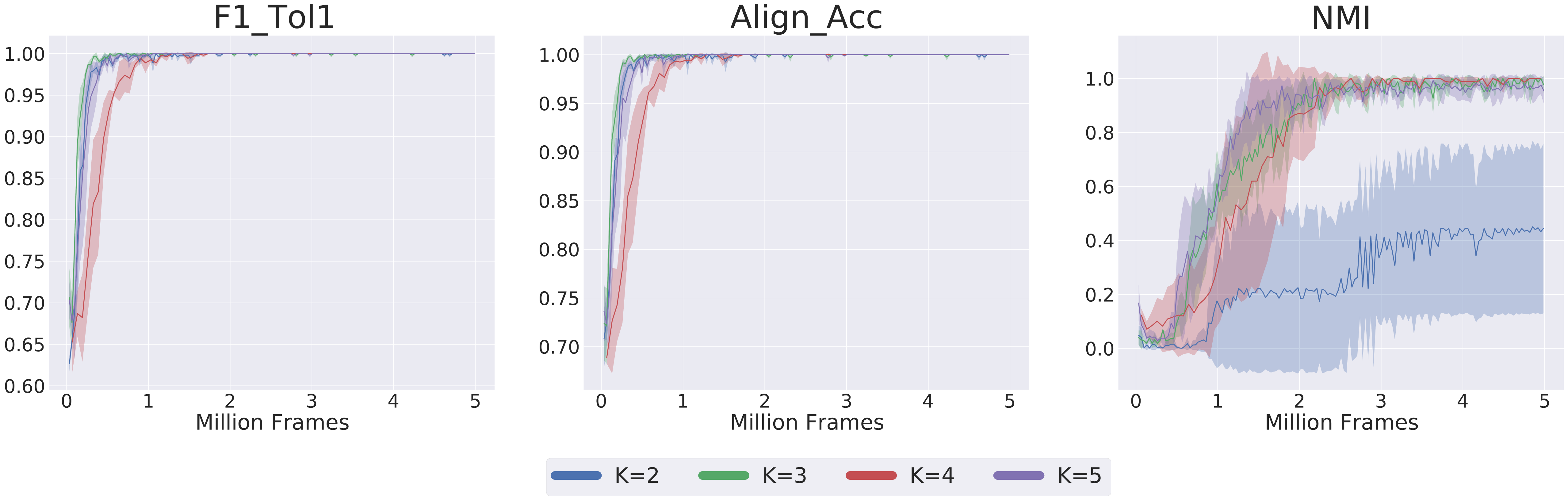}
    \caption{Comparison of parsing results during different $K$ at F1 scores with tolerance, task align accuracy and NMI.}
    \label{fig:ana}
\end{figure}

\begin{figure}[h]
    \centering
    \includegraphics[width=0.8\linewidth]{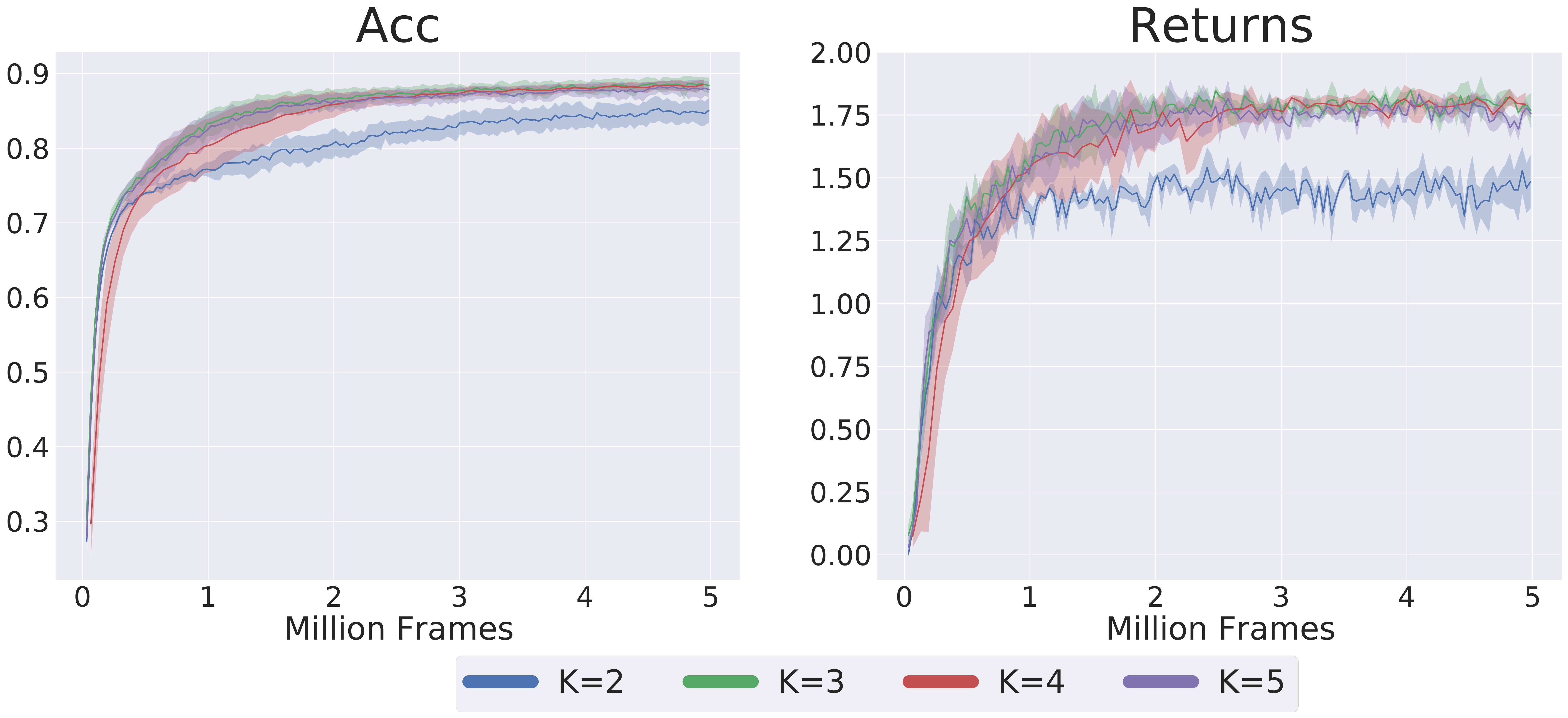}
    \caption{Comparison of prediction accuracy of actions and the returns during different $K$}
    \label{fig:ret}
\end{figure}

Our model does not require the assumption about the number of skills. We analyze the effect of the number of options $K$. As shown in Figure ~\ref{fig:ana}, when $K$ is larger than or equal to the number of skills, which is 3 in this experiment, our model basically remains similar results at three metrics: Align Acc, F1 Tol1, and NMI and achieve almost 1. When $K=2$, which means $K$ is smaller than the number of skills, one of the options must execute two different skills, which is contrary to our assumption and only achieves 0.4 at $NMI$. 
We also compare the prediction accuracy of the actions and the returns in Figure ~\ref{fig:ret}. Our performance isn't influenced by the number of skills when $K$ is larger than the number of skills.

    

    


\section{Rethinking Modularization}
Modern Neural Network research focuses on training end-to-end models on a large amount of data.
This data-driven paradigm has enjoyed great success.
However, it also has some drawbacks: 
1) a neural network model usually focuses on a single task, which means that we need to deploy a standalone model to the user's device for each functionality;
2) neural network models function like black boxes, when it fails we usually don't have a method to diagnose the source of failure;
3) fixing a fault requires retraining the entire model, which is too expensive for many cases and potentially brings in other problems.

We want to suggest that the combination of syntactic inductive bias and modularization could be a potential solution to these problems. 
For example, OCN uses the syntactic inductive bias to induce the latent structure of a given task, while maintaining the use of an end-to-end schema.
At the same time, OCN recognizes and models each repetitive component in the tree structure as a standalone neural network module (skill) with a human-understandable semantic meaning.
This process allows researchers and engineers to understand the internal mechanism of the trained model.
Furthermore, each module could be reused to solve other tasks, this could minimize the cost of solving a new task and allow us to deploy a new function to a device without sending an entirely new model.
Researchers and engineers can also quickly diagnose the faults in a given model, by identifying the malfunction components.
They can then use a small amount of specifically prepared data to retrain the component, such that the fault can be fixed with less cost and without touching the other components of the model.

Though the proposed method and experimental settings in this chapter are toyish, we do hope this could inspire other researchers who want to study the aforementioned problems.

\chapter{Conclusion}\label{cha:conclusion}

In this thesis, we introduced a group of syntactic inductive biases for neural network models, as well as their applications in Natural Language Processing and Reinforcement Learning.

In Chapter~\ref{cha:constituency}, we reviewed the history of constituency-augmented neural networks and unsupervised constituency parsing.
Traditionally, researchers tend to recognize the two as separated domains.
However, limited resources and lack of flexibility restricted the development of constituency-augmented models.
Grammar induction also suffered from a lack of practical use cases. 
We proposed to combine the two domains.
The basic idea is to add a constituency inductive bias to generic neural network architecture.
Thus, the new model is constituency augmented, but only requires raw corpus as training data and does not require extra parsed corpus.
On the other hand, the induced grammar can be directly used to improve the model's performance on downstream tasks.
We introduced \textit{Ordered Neurons} -- a constituency inductive bias for recurrent neural networks.
Based on the ordered neurons, we further introduce two neural network architectures: ON-LSTM and Ordered Memory (OM).
The ON-LSTM is a simple combination of Ordered Neurons and the LSTM model. 
It has the same generic interface as a standard LSTM model, making it a good alternative for many natural language tasks.
The OM model is designed from scratch based on the same inductive bias.
It includes a soft shift-reduce parsing mechanism and an explicit composition function to compose lower-level constituents to higher-level constituents.
These mechanisms make OM a good fit for formal language tasks.
We present experimental results on formal language tasks, language modeling, and unsupervised constituency parsing.
    
In Chapter~\ref{cha:dependency}, we reviewed the history of dependency-augmented neural networks and unsupervised dependency parsing.
Previous works also revealed that the self-attention distributions in transformers have a strong connection to the corresponding dependency relations.
Inspired by this observation, we proposed the \textit{dependency-constrained connection} -- an inductive bias for transformer or graph neural networks.
Based on this inductive bias, we introduced Unsupervised Dependency Graph Network (UDGN).
It combines a dependency parser and a graph neural network.
The dependency parser computes probability distribution for a dependency graph. 
The graph neural network uses the probability distribution to control the information propagation between nodes.
While training on a task, the gradient backpropagation updates both the information propagation mechanism (the graph neural network) and the parser to minimize the loss function. 
We then present experimental results on masked language modeling, unsupervised dependency parsing, and semantic textual similarity tasks.
Experiment results confirm our hypothesis that a dependency graph is efficient for information propagation.
The proposed model achieves competitive performance when compared to the transformer and other baselines.
    
In Chapter~\ref{cha:rl}, we reviewed the history of Hierarchical Imitation and Reinforcement Learning (HIRL).
We found that the temporal hierarchical structure used in HIRL is similar to the constituency tree structure.
Thus, we introduced the Option-Controller Network (OCN) -- a HIRL model with constituency inductive bias.
A typical OCN includes a controller and a set of options.
The controller focuses on higher-level planning, while the options provide lower-level skills.
During execution, the controller starts by selecting an option to execute, then it will wait until the selected option report finish to select a second option.
This mechanism is developed from Ordered Neurons inductive bias.
In experiments, we perform behavior cloning from unstructured demonstrations coming from different tasks, and during the RL finetuning, we freeze the learned options and only re-initialize the controller.
Experiment results show that the learned options can function as a plug-and-play module for other OCN models.
While facing a new task, we can simply gather previously learned options and initialize a new controller, then the RL algorithm will optimize the controller to solve the new task.

\begin{figure}[h]
    \centering
    \includegraphics[width=\linewidth]{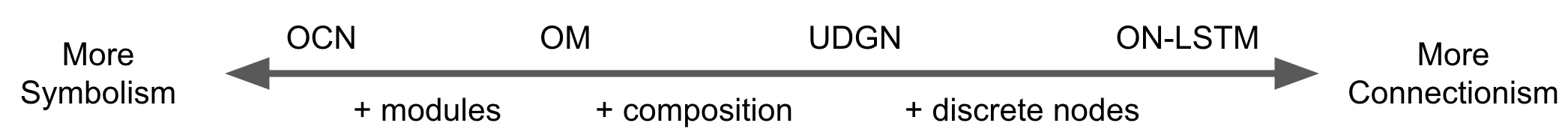}
    \caption{The spectrum from connectionism to symbolism of our proposed models.}
    \label{fig:spectrum}
\end{figure}

In Figure \ref{fig:spectrum}, we provide a spectrum of our models, arranged according to their level of discreteness.
In general, a model that has a discrete inner structure and a symbolic-like pattern is considered more symbolic. 
On the contrary, a model that has no inner structure (fully connected) and no symbolic patterns (distributed representations) are considered as more connectionist.
In our works, ON-LSTM is the most connectionist model, because it has the least discrete internal structures and no vector representations for non-terminal nodes in the tree structure. 
UDGN is less connectionist than ON-LSTM because it learns a more discrete internal structure and it has a vector representation for each node in the dependency graph.
OM is more symbolic than UDGN, because it can converge to a discrete internal structure, has a vector representation for each node, and also models the compositional function for these nodes.
OCN is the most symbolic model in our work, because it has all the previous features (except that it models decomposition instead of composition), and it also models each task and subtask as an individual module, that a new agent or user can reuse them as simple meaningful symbols.

Altogether, we are excited about the progress that has been made in this field for the past few years and have been glad to be able to contribute. 
At the same time, we also believe that current progress is just the initial steps for this newly emerged domain. 
As we relentlessly argued through this thesis, the goal of this field is not just to induce the latent structure of natural language, but also to encourage neural network models to use it as the innate structure of reasoning. 
Based on this reason, we want to encourage researchers to evaluate their model on both unsupervised parsing performance and how the model performance in terms of generalization and robustness.
In the rest of this chapter, we will discuss some important future milestones.

\section{Future Directions} \label{cha:future}

\subsection{Emerging Discrete and Useful Structure}
Syntactic inductive biases are first proposed as a guideline to develop unsupervised parsing models.
For most of the current methods, the induced structure is greedily sampled from a distribution of all possible structures.
Almost all methods focus on evaluating the coherence between the sampled structure and the gold structure given by expert annotators. 
The sharpness of the distribution is usually not taken into consideration.
However, as we argued in the introduction, the other important role of syntactic inductive bias is to regularize the internal connection of neural network models.
In this case, a smooth distribution could result in a weak or nonexistent regularization effect.
For example, in UDGN, if the dependency distribution is close to a uniform distribution, then the latent structure of the model reduces to fully connected.
In Ordered Memory, if the attention distribution at every time step is uniform, the model cannot compose information as a recursive neural network, thus failing to model the intrinsic composition function.
In this case, it's less likely that the model can still generalize to Out-Of-Distribution datapoints.
Similarly, in Option-Controller Network, learning reusable skills also require the structural decisions to be sharp.
Hence, we would like to argue that the sharpness of distribution is at least as important as the correctness of induced latent structure for tasks that require strong generalization ability.

Encouraging a sharp distribution is also an important step towards combining symbolism and connectionism.
The discrete structure given by a sharp distribution can be considered as the computation graph in a symbolic system.
A sharp distribution can guarantee that information can only be combined or communicated through the induced structure.
In a tree-like structure, this effect can create information bottlenecks.
So the model can learn the vector representation of a phrase or a sentence.
This is similar to the symbolic system, that we can find meaningful intermediate results in the computation graph.
These intermediate results contribute to the excellent interpretability of symbolic methods.
In the Ordered Memory model, we can find these intermediate results stored in memory slots.
For example, after processing a ListOps equation, each parenthesis in the equation can find its distributed representation stored in at least one memory slot.
For natural language, we proposed ON-LSTM and UDGN to learn different types of latent structures.
However, how we emerge sharp distributions and get interpretable intermediate results in a natural language task remains an open question.

On the other hand, we also want to argue that the quality of latent structure can be evaluated by multiple metrics.
Unsupervised parsing performance is an important metric.
It can provide us insights into what structure the model induced.
Also, gold trees are just human annotations. 
They are not necessarily the ground-truth structure or the best structure for certain tasks~\citep{dai2021does}.
For future works, we would like to explore other evaluation metrics that assess the quality of induced structure from a practical angle, for example, the systematic generalization ability or fine-tuning performance on downstream tasks.

\subsection{Inducing Reusable Operators}
Once a model could stably produce a discrete structure, a reasonable next step is to try to replace the single composition/communicate function with a set of operators.
For example, in UDGN, we use the competitive mechanism to select different communication channels for each pair of tokens.
In OCN, we have different options for different subtasks.
The idea is inspired by the operators in the formal language and the syntactic functions in theoretical linguistics.
In Logical Inference, we have operators like \textit{or} and \textit{and}.
In dependency grammar, we have syntactic functions like \textit{SUBJ} and \textit{PRED}.
One common feature of these operators is reusability.
It means that the same operator can be used in many different contexts and still hold the same functionality.
The only constraint is that inputs of the operator should share some common features.

Introducing a set of operators has two major advantages.
The first one is that simpler operators usually have better reusability because they require less input information and are less sensitive to the context.
For example, in OCN, an option that models only one subtask can be reused by any controller when the subtask is encountered.
If the option models two different subtasks, the controller will have to pass a command vector to the option to disambiguate the subtask.
This command vector could cause extra effort for the new controller to learn the protocol.
Secondly, it's easier to diagnose and fix a bug.
For example, in OCN, it's possible to associate some particular failures with a specific component (an option or a controller).
An easy fix would be simply reinitializing the failed component and fixing the rest of the model, then retraining the model on a specially designed dataset.

In our work, UDGN provides a potential solution to learn syntactic functions in a natural language setting. 
In an undirected dependency graph, UDGN uses a competitive mechanism to select the module for each edge to propagate information.
But, how to clearly define and learn reusable operators in natural language tasks remains an open question.

\subsection{Systematic Generalization}
For language, systematic generalization requires a model to be able to reason about all valid sequences of tokens despite being trained on a very small subset of them.
In a symbolic framework, systematic generalization is straightforward.
Once the operators and grammar are well-defined, any inputs that satisfy the pre-defined grammar can be processed by the system.
However, this setting also has great limitations in the real world.
Firstly, natural language has very complicated grammar, and even a state-of-the-art parser could fail in many cases.
Secondly, the operators used by natural language are complicated to be pre-defined.
In this thesis, we try to solve the two problems with a data-driven approach, that is, given a pre-designed inductive bias, we want the model to induce the syntax and operators from the supervised or unsupervised training losses.
We observe some successes on synthetic tasks, including ListOps, Logical Inference, and Craft. 
But reaching systematic generalization on natural language will still require a lot of future effort and creativity. 





\bibliographystyle{plainnat}     
\def\bibname{References} 
\bibliography{ref}     

\end{document}